\documentclass{article}
\usepackage{arydshln}
\usepackage[cmyk]{xcolor}
\usepackage{amsfonts}
\usepackage{amsmath}
\usepackage{hyperref}
\usepackage[title]{appendix}
\usepackage[a4paper, total={6.2in, 9.5in}]{geometry}
\usepackage{empheq}
\usepackage{graphicx}
\usepackage{amssymb}
\usepackage{color, colortbl}
\usepackage{caption}
\usepackage{multirow}
\usepackage{mathrsfs}
\usepackage{authblk}
\usepackage{ulem}

\let\originalLabelenumi\labelenumi

\newcounter{subEquation}  
\renewcommand{\thesubEquation}{\theequation.\alph{subEquation}}  

\usepackage[x11names]{xcolor}
\usepackage{tikz}
\usetikzlibrary{matrix, positioning}
\usepackage{nicematrix}

\makeatletter
\DeclareRobustCommand*{\Myhref}[1][]{
  \begingroup
  \setkeys{Hyp}{#1}
  \@ifnextchar\bgroup\Hy@href{\hyper@normalise\href@}%
}
\makeatother

\newcommand{\EnumCols}{red,DarkOrange1,Green3}
\newcommand{\EnumItems}[1]{\foreach \X[count=\Y] in \EnumCols
{\ifnum\Y=#1\relax
\xdef\EnumCol{\X}
\fi
}
\tikz[baseline=(EnumItems.base),remember
picture]{%
\node[fill=\EnumCol,inner sep=3pt,font = \bfseries, fill opacity=0.7] (EnumItems){#1};}
}
\newcommand{\EnumHighlight}{\tikz[overlay,remember picture]{%
\fill[\EnumCol,fill opacity=0.7] ([yshift=9pt,xshift=-\pgflinewidth]EnumItems.east) -- ++(9pt,-9pt)
-- ++(-9pt,-9pt) -- cycle;
}}

\usepackage{empheq}

\usepackage[most]{tcolorbox}

\newcommand*{\vertbar}{\rule[-1ex]{0.5pt}{2.5ex}}
\newcommand*{\horzbar}{\rule[.5ex]{2.5ex}{0.5pt}}

\newcommand{\centerwithin}[2]{
  {\mathmakebox[\widthof{\ensuremath{{}#2{}}}][c]{{#1}}}}

\usepackage{scalerel}

\let\bigopsize\bigoplus
\def\bigoplus{{\scalerel*{\boldsymbol\oplus}{\bigopsize}}}

\newcommand{\mathcolorbox}[2]{\colorbox{#1}{$\displaystyle #2$}}
\definecolor{shadecolor}{cmyk}{0,0,0.41,0}
\definecolor{light-blue}{cmyk}{0.25,0,0,0}
\definecolor{green-yellow}{rgb}{0.68, 1.0, 0.18}
\definecolor{Gray}{gray}{0.9}
\definecolor{brilliantlavender}{rgb}{0.96, 0.73, 1.0}

\newcolumntype{z}{>{\columncolor{shadecolor}}c}
\newcolumntype{b}{>{\columncolor{shadecolor}}c}
\newcolumntype{y}{>{\columncolor{green-yellow}}c}
\newcolumntype{x}{>{\columncolor{Gray}}c}
\newcolumntype{g}{>{\columncolor{Gray}}c}

\DeclareMathAlphabet\mathbfcal{OMS}{cmsy}{b}{n}

\newsavebox{\mysaveboxM} 
\newsavebox{\mysaveboxT}

\newcommand*\forwardBox[2][Example]{%
    \sbox{\mysaveboxM}{#2}%
    \sbox{\mysaveboxT}{\fcolorbox{black}{orange}{#1}}%
\sbox{\mysaveboxM}{%
      \parbox[b][\ht\mysaveboxM+.5\ht\mysaveboxT+.5\dp\mysaveboxT][b]{%
        \wd\mysaveboxM}{#2}%
    }%
\sbox{\mysaveboxM}{%
      \fcolorbox{black}{green-yellow}{%
        \makebox[0.95\linewidth-5em]{\usebox{\mysaveboxM}}%
      }%
}%
\usebox{\mysaveboxM}%
    \makebox[0pt][r]{%
      \makebox[\wd\mysaveboxM][c]{%
        \raisebox{\ht\mysaveboxM-0.5\ht\mysaveboxT
+0.5\dp\mysaveboxT-0.5\fboxrule}{\usebox{\mysaveboxT}}%
}%
}%
}

\newcommand*\backPropBox[2][Example]{%
    \sbox{\mysaveboxM}{#2}%
    \sbox{\mysaveboxT}{\fcolorbox{black}{light-blue}{#1}}%
\sbox{\mysaveboxM}{%
      \parbox[b][\ht\mysaveboxM+.5\ht\mysaveboxT+.5\dp\mysaveboxT][b]{%
        \wd\mysaveboxM}{#2}%
    }%
\sbox{\mysaveboxM}{%
      \fcolorbox{black}{shadecolor}{%
        \makebox[0.98\linewidth-5em]{\usebox{\mysaveboxM}}%
      }%
}%
\usebox{\mysaveboxM}%
    \makebox[0pt][r]{%
      \makebox[\wd\mysaveboxM][c]{%
        \raisebox{\ht\mysaveboxM-0.5\ht\mysaveboxT
+0.5\dp\mysaveboxT-0.5\fboxrule}{\usebox{\mysaveboxT}}%
}%
}%
}

\newtcbox{\mymath}[1][]{%
    nobeforeafter, math upper, tcbox raise base,
    enhanced, colframe=blue!30!black,
    colback=blue!30, boxrule=1pt,
    #1}

\usetikzlibrary{backgrounds}
\usepackage{changepage}

\begin{document}

\title{Deep learning for pedestrians: backpropagation in Transformers}
\author{Laurent Bou\'e \\ \Myhref[hidelinks]{mailto:ranlot75@gmail.com}{\protect\includegraphics[width=1cm,height=0.6cm]{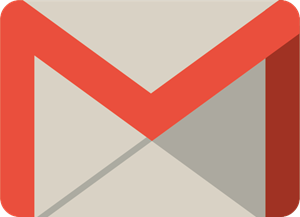}} \,\,\,\, \Myhref[hidelinks]{https://www.linkedin.com/in/laurent-bou\%C3\%A9-b7923853/}{\protect\includegraphics[width=0.8cm,height=0.8cm]{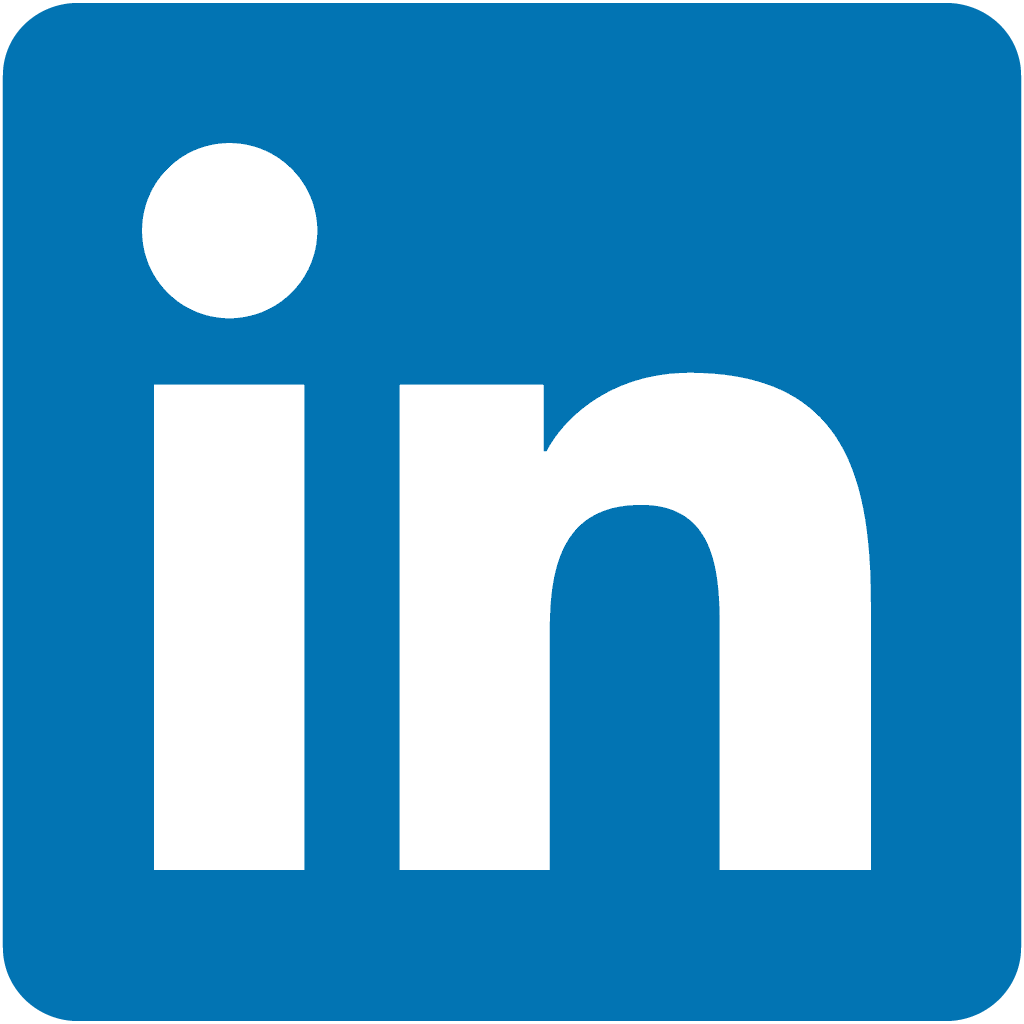}} \,\,\,\, \Myhref[hidelinks]{https://github.com/Ranlot}{\protect\includegraphics[width=1cm,height=1cm]{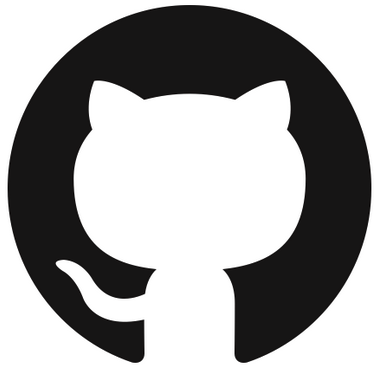}} \,\,\,\, \Myhref[hidelinks]{https://twitter.com/ranlot75}{\protect\includegraphics[width=0.8cm,height=0.8cm]{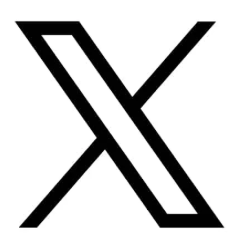}} }
\affil{Oracle~\thanks{(work initiated while at Microsoft)}}
\date{}
\maketitle

\begin{abstract}
This document is a follow-up to our previous paper dedicated to a vectorized derivation of backpropagation in CNNs~\cite{deepPedestrians}.  Following the same principles and notations already put in place there, we now focus on transformer-based next-token-prediction architectures.  To this end, we apply our lightweight index-free methodology to new types of layers such as embedding, multi-headed self-attention and layer normalization.  In addition, we also provide gradient expressions for LoRA layers to illustrate parameter-efficient fine-tuning.  Why bother doing manual backpropagation when there are so many tools that do this automatically?  Any gap in understanding of how values propagate forward will become evident when attempting to differentiate the loss function.  By working through the backward pass manually, we gain a deeper intuition for how each operation influences the final output.  A complete PyTorch implementation of a minimalistic~GPT-like network is also \href{https://github.com/Ranlot/backpropagation-Transformers}{provided} along with analytical expressions for of all of its gradient updates.
\end{abstract}

\section{Sequence modeling from~20,000 feet...}

\paragraph{Data representation.}  In the following, a token is understood to be any atomic unit of information such as words in natural language, pixels in an image,  amino acids in proteins, time stamps in time series forecasting...  A ``sample'' of data is understood to be a \underline{sequence} of~$n_\mathcal{T}$ tokens where the relative arrangement of tokens with respect to each other encodes a meaningful higher-level structure.  For instance, in natural language, the meaning of a sentence emerges from the way sequences of words are combined with each other to convey higher-level ideas.  Similarly, in computer vision, while each pixel in an image holds a value (such as color or intensity), higher-level concepts like objects emerge when coherent sequences of pixels are considered together.  \\

\noindent Each token~$t$ is identified~---~via a specialized ``tokenizer''~---~as an integer~$t  \sim \mathbb{N}  \in [1, \cdots , n_\text{vocab}]$ where~$n_\text{vocab}$ corresponds to the maximum number of tokens in our ``vocabulary''.  We refer the reader to~\cite{tokenizerReview} for a review of tokenizers for different data modalities.  Therefore, one single data input is denoted by a vector of integers~$[t_1 \sim \mathbb{N}, \cdots , t_{n_\mathcal{T}} \sim \mathbb{N}] \sim \mathbb{N}^{n_\mathcal{T}}$ of size~$n_\mathcal{T}$ corresponding to the number of tokens in the sequence.

\paragraph{Next-token prediction.}  Let us denote the model as a parametrized function~$\mathcal{N}_\mathcal{P}$ that takes as input a sequence of tokens~${\bf a}_0 \sim \mathbb{N}^{n_\mathcal{T}}$ and returns a new sequence
\begin{equation*}
{\bf y}_\text{pred} = \mathcal{N}_\mathcal{P}({\bf a}_0) \sim \mathbb{R}^{n_\mathcal{T}\times n_\text{vocab}}
\end{equation*}
where each token~$\sim \mathbb{N}$ is transformed into a normalized probability density vector~$\sim \mathbb{R}^{n_\text{vocab}}$ over the vocabulary of tokens.  Alongside this probabilistic prediction, each token is also associated with a ground-truth target token which, ideally, should match as best as possible the prediction vector produced by~$\mathcal{N}_\mathcal{P}$.  Graphically, this can be represented as
\begin{equation*}
{\bf a}_0 = \left(
\begin{array}{ccc}
t_1 \sim \mathbb{N} \\
 \vdots  \\
t_{n_\mathcal{T}} \sim \mathbb{N}
\end{array}
\right) \,\, \longrightarrow \,\, {\bf y}_\text{pred} = \left(
\begin{array}{ccc}
{\bf y}_\text{pred}(t = 1) \sim \mathbb{R}^{n_\text{vocab}} \\
 \vdots  \\
{\bf y}_\text{pred}(t = n_\mathcal{T}) \sim \mathbb{R}^{n_\text{vocab}}
\end{array}
\right) \quad \text{vs.} \quad {\bf y}_\text{gt} = \left(
\begin{array}{ccc}
y_\text{gt}(t = 1) \sim \mathbb{N} \\
 \vdots  \\
y_\text{gt}(t = n_\mathcal{T}) \sim \mathbb{N}
\end{array}
\right)
\end{equation*}
As usual in classification settings, the mismatch between the prediction and the ground-truth for token~$t$ is quantified via the cross-entropy loss function
\begin{equation*} 
{\bf \ell}_\mathbfcal{P} \left( {\bf y}_\text{gt} \, , {\bf y}_\text{pred} \right) = \left(
\begin{matrix}
- \, {\bf y}_\text{gt}(t=1)_\text{OHE} \cdot \log \, {\bf y}_\text{pred}(t=1) \sim \mathbb{R} \\
\vdots \\
- \, {\bf y}_\text{gt}(t=n_\mathcal{T})_\text{OHE} \cdot \log \, {\bf y}_\text{pred}(t=n_\mathcal{T}) \sim \mathbb{R}
\end{matrix}
\right) = - {\bf y}_\text{gt} \ominus \log  {\bf y}_\text{pred} \sim \mathbb{R}^{n_\mathcal{T}}
\end{equation*}
applied independently to all~$n_\mathcal{T}$ tokens in the sequence and where the ground-truth tokens have been lifted from integers into their equivalent ``One Hot Encoded'' (OHE) representations. \\

\noindent For next-token prediction tasks (relevant for model pre-training and supervised fine-tuning~---~SFT) the ground-truth~${\bf y}_\text{gt}$ is chosen as a ``shifted-by-one'' version of the input~${\bf a}_0$.  That is to say that, considering a token~$t^\star$ from the input~${\bf a}_0$, its target should be~$y_\text{gt}(t = t^\star) = t^\star + 1$ taken from the same~${\bf a}_0$.  This may be understood graphically as
\begin{equation*} 
{\bf y}_\text{gt} = \text{shiftByOne} ({\bf a}_0) \quad \Longleftrightarrow \quad
\begin{tikzpicture}[every node/.style={anchor=center}, baseline=(input.center)]

\node (label) at (0, 0) {${\bf y}_\text{gt} =$}; 

\node (label2) at (7, 0) {${\bf a}_0 = $}; 

\matrix (input) [matrix of math nodes, left delimiter={(}, right delimiter={)}, row sep=6pt, right=of label, xshift=-0.4cm] {
y_\text{gt}(t=1) = t_2 \sim \mathbb{N} \\
\vdots \\
y_\text{gt}(t=n_\mathcal{T}-1) = t_{n_\mathcal{T}} \sim \mathbb{N} \\
 \\
};

\matrix (target) [matrix of math nodes, right of=input, node distance=5.5cm, left delimiter={(}, right delimiter={)}, row sep=6pt] {
t_1 \sim \mathbb{N} \\
t_2 \sim \mathbb{N} \\
\vdots \\
t_{n_\mathcal{T}} \sim \mathbb{N} \\
};

  \foreach \i in {1,3} {
    \pgfmathtruncatemacro{\j}{\i+1}
    \draw[<->, thick, dashed] (input-\i-1.east) -- (target-\j-1.west);
  }

\end{tikzpicture}
\end{equation*}

\noindent Due to this shift-by-one property between the tokens of~${\bf y}_\text{gt}$ and~${\bf a}_0$ in standard next-token prediction tasks, the very first token~$t_1$ in~${\bf a}_0$ is not used in the loss function and the number of elements contributing to the loss for a sequence of~$n_\mathcal{T}$ tokens is reduced to~$n_\mathcal{T}-1$.

\section{Embedding layer}
\label{sec:embedding}

\noindent The purpose of embedding layers is to transform categorical variables from their discrete representations --- where they exist as static structureless elements of a set --- into continuous and ``meaningful''~$d$-dimensional vector-based representations~$\sim \mathbb{R}^d$ known as ``embeddings''. Here, ``meaningful'' implies that the learned embedding vectors are expected to capture useful and objective-dependent information (as enforced by the loss function). \\

\noindent While tokens are the primary categorical variables for which ``token embeddings'' are constructed, their relative positions within a sequence also constitute an abstract categorical variable giving rise to ``positional embeddings.''  In the case of token embeddings, some desirable properties would, for example, be that tokens with similar semantic meanings would be associated with vector representations that are close to each other.  Conversely, one would expect positional embeddings to reflect order-sensitive information.  Let us consider a specific token~$t^\star$ from an input sequence~${\bf a}_{i-1}$ of~$n_\mathcal{T}$ tokens~\footnote{\hypertarget{embedSource}{Typically}~${\bf a}_{i-1}$ would be the very first data source~$i=1$ so that~${\bf a}_{i-1}\equiv {\bf a}_0$.  Nonetheless, we stick to the~${\bf a}_{i-1}$ notation for input data into a layer and~${\bf a}_i$ for its output to stay with a consistent  terminology for all types of layers.}.  As a discrete unit of information, this token can always be represented as an integer~${\bf a}_{i-1}(t=t^\star) \sim \mathbb{N}$ but the range of possible values~${\bf a}_{i-1}(t=t^\star)\in[1, \cdots , n_\mathcal{V}]$ depends on whether we look at it from its vocabulary-membership perspective or from its position in the sequence.
\begin{itemize}
\item For the purposes of token embeddings, the tokenizer assigns~$t^\star$ with~${\bf a}_{i-1}^\text{token}(t=t^\star) \in [1, \cdots , n_\text{vocab}]$ where~$n_\text{vocab}$ denotes the total number of tokens in the vocabulary.  The exact integer value serves only to distinguish~$t^\star$ from all the other tokens in the vocabulary but carries no inherent meaning.
\item On the other hand, for positional embeddings, the integer representation of~$t^\star$ would correspond to the index of its position in the sequence of length~$n_\mathcal{T}$.  Therefore~${\bf a}_{i-1}^\text{position}(t=t^\star) \in [1,\cdots , n_\text{context}]$ where~$n_\text{context}$ is the context length of the model, i.e. the maximum number of tokens that the model can handle as a single sequence.  Unless the sequence~${\bf a}_{i-1}$ has been padded to exactly match the context length, we would have~$n_\text{context} \geq n_\mathcal{T}$.
\end{itemize}

\noindent Without loss of generality, let us denote by~$n_\mathcal{V}$ the number of possible integers that tokens may be associated with and introduce a database-like structure~${\bf w}_\text{emb} \sim \mathbb{R}^{n_\mathcal{V} \times d}$ where each row of~${\bf w}_\text{emb}$ contains the ``embedding'' representation for each one of the possible token values
\begin{equation}
{\bf w}_\text{emb} = \left(
\begin{array}{ccc}
\horzbar & {\bf w}_\text{emb}(t=1) \sim \mathbb{R}^{d} & \horzbar \\
& \vdots & \\
\horzbar & {\bf w}_\text{emb}(t=n_\mathcal{V}) \sim \mathbb{R}^{d} & \horzbar
\end{array}
\right) \sim \mathbb{R}^{n_\mathcal{V}\times d} \quad \text{where} \quad 
n_\mathcal{V} \equiv \begin{cases} 
 n_\text{vocab} & \text{``token''} \\
 n_\text{context} & \text{``positional''}
\end{cases}
\label{eq:embedStruct}
\end{equation}
Although embeddings may initially be assigned random values, it is important to realize that they must remain trainable so that, once optimized, tokens acquire representations that reflect the structure and requirements of the task. We will see in the backward pass how the embeddings receive error signals from the loss function allowing them to converge to representations that effectively encode relevant information.  In transformer-based deep learning architectures, it is the responsibility of mechanisms such as self-attention to learn inter-token dependencies as we will see in Section~\ref{sec:singleHead} and transform them into the error signals that reach the embedding layer. \\

\noindent Although token embeddings are usually learned via backpropagation, there does exist non-adjustable versions of positional embeddings that are designed to manually impose some structural constraints.  This is the case, for example, with RoPE (Rotary Positional Embeddings) which aims to break the \hyperlink{permutationAtt}{permutation equivariance} property of non-causal attention by injecting information about the relative position of the tokens.

\paragraph{Forward pass.}  During the forward pass, the job of the embedding layer is simply to pull out the relevant embeddings associated with the~$n_\mathcal{T}$ tokens present in the input sequence~${\bf a}_{i-1} \sim \mathbb{N}^{n_\mathcal{T}}$ from the embedding store~${\bf w}_\text{emb}$ defined in~eq.\eqref{eq:embedStruct}.  In order to formulate this database lookup as a differentiable vectorized operation, let us first transform~${\bf a}_{i-1}$ into its ``One Hot Encoded'' (OHE) representation. \\

\noindent As an example, let us go back to token~$t^\star$ and expand its integer representation~${\bf a}_{i-1}(t=t^\star)\sim \mathbb{N}$ into a~$n_\mathcal{V}$-dimensional binary sparse vector
\begin{equation*}
{\bf a}_{i-1}(t=t^\star) \sim \mathbb{N} \quad  \longrightarrow  \quad {\bf a}_{i-1}(t=t^\star)_\text{OHE} = \big[ a_{t^\star 1} , \cdots , a_{t^\star n_\mathcal{V}} \big]  \sim \mathbb{R}^{n_\mathcal{V}} \,\, \,\, \text{with} \,\,\,\, a_{t^\star t^\prime} = \delta_{t^\star t^\prime}
\end{equation*}
This way, the vector~${\bf a}_{i-1}(t=t^\star)_\text{OHE} \sim \mathbb{R}^{n_\mathcal{V}} $ only has a single non-null component~$a_{t^\star t^\prime} = 1$ when~$t^\prime = t^\star$ and all other components are~$a_{t^\star t^\prime} = 0$ for~$t^\prime \neq t^\star$ across the whole range~$t^\prime \in [1, \cdots , n_\mathcal{V}]$. Because of this~OHE representation, the product
\begin{equation*}
{\bf a}_{i-1}(t=t^\star)_\text{OHE} \, {\bf w}_\text{emb} = {\bf w}_\text{emb}(t=t^\star) \sim \mathbb{R}^{d}
\end{equation*}
immediately picks up the correct embedding vector for token~$t^\star$.  Applying this OHE representation to all~$n_\mathcal{T}$ tokens, the input data~${\bf a}_{i-1} \sim \mathbb{N}^{n_\mathcal{T}}$ can therefore be represented as a sparse array
\begin{equation*}
\text{ohe}({\bf a}_{i-1}) \equiv \left(
\begin{array}{ccc}
\horzbar & {\bf a}_0(t=1)_\text{OHE} \sim \mathbb{R}^{n_\mathcal{V}}  & \horzbar \\
& \vdots & \\
\horzbar & {\bf a}_0({t={n_\mathcal{T}}})_\text{OHE} \sim \mathbb{R}^{n_\mathcal{V}} & \horzbar
\end{array}
\right) \sim \mathbb{R}^{n_\mathcal{T}\times n_\mathcal{V}}
\end{equation*}
which can be multiplied by~${\bf w}_\text{emb} \sim \mathbb{R}^{n_\mathcal{V}\times d}$ to simultaneously pull out the relevant~$n_\mathcal{T}$ embedding vectors associated with the specific~$n_\mathcal{T}$ tokens present in the input sequence~${\bf a}_{i-1} \sim \mathbb{N}^{n_\mathcal{T}}$ and collect them into the output~${\bf a}_i \sim \mathbb{R}^{n_\mathcal{T}\times d}$ of the embedding layer
\begin{equation*}
{\bf a}_i = \text{ohe}({\bf a}_{i-1}) \, {\bf w}_\text{emb} = \left(
\begin{array}{ccc}
\horzbar & {\bf w}_\text{emb}(t=1) \sim \mathbb{R}^{d}  & \horzbar \\
& \vdots & \\
\horzbar & {\bf w}_\text{emb}(t=t_{n_\mathcal{T}}) \sim \mathbb{R}^{d} & \horzbar
\end{array}
\right) \sim \mathbb{R}^{n_\mathcal{T}\times d}
\end{equation*}

\noindent In summary, the forward pass of an embedding layer results in~${\bf a}_i \sim \mathbb{R}^{n_\mathcal{T}\times d}$ where the rows of~${\bf a}_i$ contain the embeddings of the~$n_\mathcal{T}$ tokens present in~${\bf a}_{i-1}$ by extracting them from the complete embedding store~${\bf w}_\text{emb}$ defined in~eq.\eqref{eq:embedStruct} via~OHE matrix multiplication
\begin{empheq}[box={\forwardBox[{\bf Embedding layer}: forward pass]}]{alignat=2}
{\bf a}_i &= \text{ohe}({\bf a}_{i-1}) \, {\bf w}_\text{emb} 
\label{embeddingLayer:forward}
\end{empheq}

\paragraph{Backward pass.}  As usual, we need to evaluate the recursive backward error flow described in~eq.(11) of the reference paper~\cite{deepPedestrians} with~$\boldsymbol{\Delta}_{i} \sim {\bf a}_i \sim \mathbb{R}^{n_\mathcal{T}\times d}$ so that
\begin{align*}
\boldsymbol{\Delta}_{i} \cdot \text{d}{\bf a}_i  &= \boldsymbol{\Delta}_{i} \cdot \text{d} \left( \text{ohe}({\bf a}_{i-1}) \, {\bf w}_\text{emb} \right) \\
&= \underbrace{ \mathcolorbox{shadecolor}{ \text{ohe}({\bf a}_{i-1})^t \, \boldsymbol{\Delta}_{i} }}_{\textstyle
    \begin{gathered}
      \frac{\partial \mathcal{L}_{\text{seq}}}{\partial {\bf w}_\text{emb}}
    \end{gathered} } \cdot \, \text{d} {\bf w}_\text{emb} + \underbrace{ \mathcolorbox{shadecolor}{ \boldsymbol{\Delta}_{i} \, {\bf w}_\text{emb}^t } }_{\textstyle
    \begin{gathered}
      \boldsymbol{\Delta}_{i-1}
    \end{gathered} } \cdot \, \text{d} \left(\text{ohe}({\bf a}_{i-1})\right)
\end{align*}

\noindent  Normally, we consider the input data sequence~${\bf a}_{i-1} \equiv {\bf a}_0$ (see \hyperlink{embedSource}{footnote}) so that the backward error flow stops here with~$\text{d}\left( \text{ohe}({\bf a}_{i-1})\right) \sim \text{d} {\bf a}_{i-1} = \mathbf{0}$ and therefore there is no need to consider~$\boldsymbol{\Delta}_{i-1}$.  Some scenarios where one may be interested in keeping this term involve adversarial attacks, feature attribution/interpretability...  In summary, the backward pass through an embedding layer is given by
\begin{empheq}[box={\backPropBox[{\bf Embedding layer}: backward pass]}]{alignat=2}
\frac{\partial \mathcal{L}_{\text{seq}}}{\partial {\bf w}_\text{emb}} &= \text{ohe}({\bf a}_{i-1})^t \, \boldsymbol{\Delta}_{i} &\quad  &\sim \mathbb{R}^{n_\mathcal{V}\times d} 
\end{empheq}

\section{Self-attention layer: Single head}
\label{sec:singleHead}

\noindent Let us start by looking at a single head of self-attention.  We will see in~Section~\ref{sec:multiHead} how a complete self-attention layer combines together multiple heads and in Section~\ref{sec:TransformerBlock} how self-attention layers themselves are composed with each other into transformer blocks.

\subsection{Self-attention layer: Single head --- Forward pass}

\begin{table}
\hspace*{-2.58cm}
\captionsetup{singlelinecheck=off}
\begin{tabular}{|| x | y | x | z ||}
\hline
 \multicolumn{4}{|c|}{\rule{0pt}{1.05\normalbaselineskip} {\bf One head~$h$ of self-attention}} \\[0.2em] \hline \hline
{\bf Layer} & \cellcolor{orange}{\bf Forward pass} & {\bf Shape} & \cellcolor{light-blue}{\bf Backward pass} \\
\hline
\hline
\rule{0pt}{1.1\normalbaselineskip}
Input data & ${\bf a}_{i-1} $ & $\mathbb{R}^{ n_\mathcal{T}\times d}$ &  $\boldsymbol{\Delta}^h_{i-1} = \boldsymbol{\Delta}_{v_h} + \boldsymbol{\Delta}_{q_h} + \boldsymbol{\Delta}_{k_h}$ \\[0.6em] \hline
\multicolumn{2}{|c|}{\rule{0pt}{1.05\normalbaselineskip} {\cellcolor{orange!40} \it $\downarrow$ (input feature maps for the~$n_\mathcal{T}$ tokens) $\downarrow$ }} & \multicolumn{1}{|c|}{\rule{0pt}{1.05\normalbaselineskip} {$\downarrow$\quad\quad$\uparrow$}}  &  \multicolumn{1}{|c|}{\rule{0pt}{1.05\normalbaselineskip} { \cellcolor{light-blue!90} \it $\uparrow$ (downstream error signal for head~$h$) $\uparrow$ }} \\[0.2em] \hline
 \multicolumn{4}{|c|}{\rule{0pt}{0.5\normalbaselineskip} } \\ \hline
\rule{0pt}{1.05\normalbaselineskip} Fully connected --- ``queries'' & ${\bf q}_h = {\bf a}_{i-1} \, {\bf w}_{q_h} + \widetilde{{\bf b}_{q_{h}}} $ & $\mathbb{R}^{n_\mathcal{T}  \times d_\rho}$ & $\boldsymbol{\Delta}_{q_h} = \boldsymbol{\Delta}^h_\text{raw} \, {\bf k}_h {\bf w}_{q_h}^t  $  \quad \quad \quad \quad \, \colorbox{light-blue}{\it (terminal)} \\[0.2em]
\hline   \rule{0pt}{1.1\normalbaselineskip}
Fully connected --- ``keys'' & ${\bf k}_h = {\bf a}_{i-1} \, {\bf w}_{k_h} + \widetilde{{\bf b}_{k_{h}}} $ & $\mathbb{R}^{n_\mathcal{T}  \times d_\rho}$ & $\boldsymbol{\Delta}_{k_h} = (\boldsymbol{\Delta}^h_\text{raw})^t \, {\bf q}_h \, {\bf w}_{k_h}^t $ \quad \quad \quad  \colorbox{light-blue}{\it (terminal)} \\[0.2em] \hline
\multicolumn{4}{|l|}{\rule{0pt}{0.5\normalbaselineskip} } \\ 
\hline  \rule{0pt}{1.1\normalbaselineskip}
\rule{0pt}{10pt} Raw attention weights & $ \boldsymbol{\rho}^\text{raw}_{({\bf a}_{i-1}, \, h) } = {\bf q}_h \, {\bf k}_h^t $   & $\mathbb{R}^{n_\mathcal{T} \times n_\mathcal{T}}$ &   \multicolumn{1}{|c|}{\cellcolor{Gray} \it ($\boldsymbol{\Delta}^h_{\text{raw}}$ splits into queries and keys branches)} \\[0.6em]
\hline   \rule{0pt}{1.1\normalbaselineskip}
\rule{0pt}{10pt} Scaling & $\boldsymbol{\rho}^\text{scaled}_{({\bf a}_{i-1}, \, h) } = \boldsymbol{\rho}^\text{raw}_{({\bf a}_{i-1}, \, h) } \, / \sqrt{d_\rho} $   & $\mathbb{R}^{n_\mathcal{T} \times n_\mathcal{T}}$ & $\boldsymbol{\Delta}^h_{\text{raw}} = \boldsymbol{\Delta}^h_{\text{scaled}} / \sqrt{d_\rho}$  \\[0.6em]
\hline  \rule{0pt}{1.1\normalbaselineskip}
\rule{0pt}{10pt} Causal mask & $\boldsymbol{\rho}^\text{causal}_{({\bf a}_{i-1}, \, h) } = {\bf m}  \circ \boldsymbol{\rho}^\text{scaled}_{({\bf a}_{i-1}, \, h) }   $   & $ \mathbb{T}_L( \mathbb{R}^{n_\mathcal{T} \times n_\mathcal{T}}) $ & $\boldsymbol{\Delta}^h_{\text{scaled}} = \boldsymbol{\Delta}^h_{\text{causal}}$  \\[0.6em] \hline 
\multicolumn{4}{|c|}{\rule{0pt}{0.5\normalbaselineskip} } \\  \hline
\rule{0pt}{18pt} Attention weights & $ \boldsymbol{\rho}_{({\bf a}_{i-1}, \, h) } = \text{softmax} \, \boldsymbol{\rho}^\text{causal}_{({\bf a}_{i-1}, \, h) } $   & $\mathbb{T}_L( \mathbb{R}^{n_\mathcal{T}\times n_\mathcal{T}} )$ & $ \boldsymbol{\Delta}^h_{\text{causal}} = \Big[  \boldsymbol{\Delta}^h_i {\bf v}_h^t   -  \widetilde{ \big( \boldsymbol{\Delta}^h_i {\bf v}_h^t \big) \ominus \boldsymbol{\rho}_{({\bf a}_{i-1}, \, h) } }  \Big] \circ  \boldsymbol{\rho}_{({\bf a}_{i-1}, \, h) } $  \\[0.6em] 
\hline  \rule{0pt}{1.1\normalbaselineskip}
Fully connected --- ``values'' & ${\bf v}_h = {\bf a}_{i-1} {\bf w}_{v_h} + \widetilde{{\bf b}_{v_{h}}} $ & $\mathbb{R}^{n_\mathcal{T}  \times d_h}$ & $\boldsymbol{\Delta}_{v_h} = \boldsymbol{\rho}_{({\bf a}_{i-1}, \, h) }^t \boldsymbol{\Delta}^h_i {\bf w}^t_{v_h} $ \quad \quad \quad \colorbox{light-blue}{\it (terminal)} \\[0.2em] \hline
\multicolumn{4}{|c|}{\rule{0pt}{0.5\normalbaselineskip} } \\ \hline
\multicolumn{2}{|c|}{\rule{0pt}{1.05\normalbaselineskip} {$\downarrow$}} & \multicolumn{1}{|c|}{\rule{0pt}{1.05\normalbaselineskip} {$\downarrow$\quad\quad$\uparrow$}} & \multicolumn{1}{|r|}{\rule{0pt}{1.05\normalbaselineskip} {\cellcolor{Gray} \it ($\boldsymbol{\Delta}^h_i$ splits into attention and values branches) }}   \\[0.2em] \hline
\multicolumn{2}{|c|}{\rule{0pt}{1.05\normalbaselineskip} { \cellcolor{orange!40} \it $\downarrow$ (transformed feature maps for the~$n_\mathcal{T}$ tokens) $\downarrow$ }} & \multicolumn{1}{|c|}{\rule{0pt}{1.05\normalbaselineskip} {$\downarrow$\quad\quad$\uparrow$}} & \multicolumn{1}{|c|}{\rule{0pt}{1.05\normalbaselineskip} {\cellcolor{light-blue!90} \it $\uparrow$ (upstream error signal for head~$h$) $\uparrow$  }}   \\[0.2em] \hline
\rule{0pt}{12pt} Output data & ${\bf a}^h_i = \boldsymbol{\rho}_{({\bf a}_{i-1}, \, h) } \, {\bf v}_h $ & $\mathbb{R}^{n_\mathcal{T}\times  d_h}$ & $\boldsymbol{\Delta}^h_i $ \\[0.3em]
\hline
\end{tabular}
\caption{Illustration of the data flow through an attention head.  In practice it is common to choose the dimensionality~$d_\rho$ of the queries/keys feature maps to match the dimensionality of the values output feature maps with~$d_\rho \equiv d_h$.  As we discuss in the main part of the text, this is not a requirement and we keep it different here for the sake of generality.  In the backward pass, we denote by ``terminal'' the branches where the upstream error signal~$\boldsymbol{\Delta}^h_i$ assigned to attention head~$h$ reaches the input feature maps~${\bf a}_{i-1}$ of the~$n_\mathcal{T}$ tokens.  Those terminal error signals converge with each other at the input data layer and accumulate to produce the downstream error signal~$\boldsymbol{\Delta}^h_{i-1}$.}
\label{table:singleHead}
\end{table}

\noindent Let us denote the input data by~${\bf a}_{i-1} \sim \mathbb{R}^{n_\mathcal{T}\times d}$ consisting of a sequence of~$n_\mathcal{T}$ tokens which each, individually, have a~$d$-dimensional feature map representation
\begin{equation*}
{\bf a}_{i-1} = \big[ \, {\bf a}_{i-1}(t=1) \sim \mathbb{R}^d, \cdots , {\bf a}_{i-1}(t=n_\mathcal{T}) \sim \mathbb{R}^d \, \big] \sim \mathbb{R}^{n_\mathcal{T}\times d} 
\end{equation*}
These feature maps may be their initial embedding representations or the output of previous transformer blocks. The purpose an attention head~$h$ is to learn a new~$d_h$-dimensional representation for each one of the~$n_\mathcal{T}$ tokens  
\begin{equation*}
{\bf a}_{i-1} \quad \longrightarrow \quad {\bf a}_i^h = \big[ \, {\bf a}_i^h(t=1) \sim \mathbb{R}^{d_h}, \cdots , {\bf a}_i^h(t=n_\mathcal{T}) \sim \mathbb{R}^{d_h} \, \big] \sim \mathbb{R}^{n_\mathcal{T}\times d_h}
\end{equation*}
The~$h$ superscript in~${\bf a}_i^h$ is here to denote that each attention head produces different output feature maps.  \\

\noindent It is expected that all the tokens in~${\bf a}_{i-1}$ should be treated collectively as a coherent sequence rather than in isolation from each other.  In other words, we would like to think of~${\bf a}_{i-1}$ as a one ``sample'' even though it is composed of multiple elements.  This way, the output feature maps should be such that tokens get information from other tokens in the sequence. Stacking the sequence of tokens together, the self-attention head~$h$ is summarized as:
\begin{empheq}[box={\mymath[colback=Gray,drop lifted shadow]}]{align*}
{\bf a}_{i-1} \sim \mathbb{R}^{n_\mathcal{T}\times d} &\equiv
\left(
\begin{array}{ccc}
\horzbar & {\bf a}_{i-1}(t=1) \sim \mathbb{R}^d  & \horzbar \\
& \vdots & \\
\horzbar &  {\bf a}_{i-1}(t=n_\mathcal{T}) \sim \mathbb{R}^d & \horzbar
\end{array}
\right) \\ &\centerwithin{\downarrow}{\equiv} \\
&\hspace{1.5cm} \centerwithin{\hbox{\tcbox[colback=green-yellow, colframe=orange]{Sequence modeling: one head $h$ of self-attention}}}{\equiv} \\
&\centerwithin{\downarrow}{\equiv} \\
{\bf a}_i^h \sim \mathbb{R}^{n_\mathcal{T}\times d_h} &\equiv
\left(
\begin{array}{ccc}
\horzbar & {\bf a}_i^h(t=1) \sim \mathbb{R}^{d_h}  & \horzbar \\
& \vdots & \\
\horzbar &  {\bf a}_i^h(t=n_\mathcal{T}) \sim \mathbb{R}^{d_h} & \horzbar
\end{array}
\right)
\end{empheq}

\noindent The complete data flow through an attention head (with causal mask) is detailed in Table~\ref{table:singleHead}.  In order to understand the logic and design choices, it is instructive to start from the desired output representation of the tokens and work backwards from there.

\paragraph{Starting from the end: What should a reasonable~${\bf a}_i^h$ look like?} Fully connected layers are the ``bread and butter'' of deep learning architectures.  To change the dimensionality of the token feature maps from the input~$d$ to the output~$d_h$, it is tempting to start by passing the input data~${\bf a}_{i-1}$ through a fully connected layer parametrized by~$\{ {\bf w}_{v_h} \sim \mathbb{R}^{d \times d_h}, {\bf b}_{v_h} \sim \mathbb{R}^{d_h} \}$.  The index~$h$ indicates that those parameters are specific to one attention head~$h$.  Therefore, we have
\begin{equation*}
\begin{NiceMatrix}
{\bf a}_{i-1}  = 
\left( \,
\begin{matrix}
{\bf a}_{i-1}(t=1) \sim \mathbb{R}^{d} \\
\vdots \\
{\bf a}_{i-1}(t=n_{\mathcal{T}})  \sim \mathbb{R}^{d}
\end{matrix} \,
\right)
\sim \mathbb{R}^{ n_\mathcal{T} \times d} \,\,\, & \,\,  \{ {\bf w}_{v_h} , {\bf b}_{v_h} \} \,\,\, & \,\, {\bf v}_h  = 
\left( \,
\begin{matrix}
{\bf v}_h(t=1) \sim \mathbb{R}^{d_h} \\
\vdots \\
{\bf v}_h(t=n_{\mathcal{T}}) \sim \mathbb{R}^{d_h}
\end{matrix} \,
\right)
\sim \mathbb{R}^{ n_\mathcal{T} \times d_h}
\CodeAfter
  \begin{tikzpicture}[->]
  \draw[double distance=1pt] (1-1.east) -- (1-2.west) ;
  \draw[double distance=1pt] (1-2.east) -- (1-3.west) ;
  \end{tikzpicture}
\end{NiceMatrix}
\end{equation*}

\begin{empheq}[box={\forwardBox[{\bf Values}: forward pass]}]{alignat=2}
{\bf v}_h &= {\bf a}_{i-1} {\bf w}_{v_h} + \widetilde{{\bf b}_{v_h}}
\label{values:forward}
\end{empheq}

\noindent Unfortunately, the ``values''~${\bf v}_h \sim \mathbb{R}^{n_\mathcal{T}\times d_h}$ produced by this operation are all independent from each other.  In other words,~${\bf v}_h(t = t^\star)$ depends solely on input token~${\bf a}_{i-1}(t = t^\star)$ without mixing in any information from any of the other tokens.  This is exactly the same as considering the~$n_\mathcal{T}$ tokens as independent samples which contradicts our goal of treating the entire sequence~${\bf a}_{i-1}$ itself as a single coherent sample.  \\

\noindent Let us now design a simple way to remedy this shortfall and start treating the tokens as context-aware sequential elements rather than a disorganized group of independent elements. \\

\noindent For clarity, we assume a \hyperlink{causality}{causal} relationship where a token can only be ``aware'' of the tokens that occur before it in the sequence and not those that follow it. In this case
\begin{itemize}
\item the first token does not have any context and therefore it is reasonable to define its output feature map~${\bf a}_i^h(t=1) \sim \mathbb{R}^{d_h}$ directly equal to its value feature map
\begin{flalign}
{\bf a}_i^h(t=1) \equiv \rho_{11} \, {\bf v}_h(t=1)  \quad \text{with} \quad  \rho_{11}  = 1 &&
\refstepcounter{equation}
\setcounter{subEquation}{0}
\refstepcounter{subEquation}
\tag{\thesubEquation} \label{eq:weighted1}
\end{flalign}
\item the second token may take information from both the first token as well as from itself.  Therefore, it is reasonable to assign its output representation~${\bf a}_i^h(t=2) \sim \mathbb{R}^{d_h}$ as a weighted average of the two available value feature maps
\begin{flalign}
{\bf a}_i^h(t=2) \equiv \rho_{21} \, {\bf v}_h(t=1) + \rho_{22} \, {\bf v}_h(t=2)  \quad \text{with} \quad  \rho_{21} +  \rho_{22} = 1 &&
\refstepcounter{subEquation}
\tag{\thesubEquation} \label{eq:weighted2}
\end{flalign}
\item similarly, the third token is expressed as a linear combination of the value feature maps from the first three tokens so that~${\bf a}_i^h(t=3) \sim \mathbb{R}^{d_h}$ is given by
\begin{flalign}
{\bf a}_i^h(t=3) \equiv \rho_{31} \, {\bf v}_h(t=1) + \rho_{32} \, {\bf v}_h(t=2) + \rho_{33} \, {\bf v}_h(t=3) \quad \text{with} \quad  \rho_{31} + \rho_{32} + \rho_{33} = 1 &&
\refstepcounter{subEquation}
\tag{\thesubEquation} \label{eq:weighted3}
\end{flalign}
\item finally, we reach the last token which has access to all of the tokens in the sequence.  Therefore, the representation of the final token~${\bf a}_i^h(t=n_\mathcal{T})\sim \mathbb{R}^{d_h}$ is written as 
\begin{flalign} 
& {\bf a}_i^h(t=n_\mathcal{T}) \equiv \rho_{n_{\mathcal{T}}1} \, {\bf v}_h(t=1) + \rho_{n_{\mathcal{T}}2} \, {\bf v}_h(t=2) + \rho_{n_{\mathcal{T}}3} \, {\bf v}_h(t=3) + \cdots +  \rho_{n_{\mathcal{T}}n_{\mathcal{T}}} \, {\bf v}_h(t=n_{\mathcal{T}}) && \nonumber \\
& \text{with} \quad \rho_{n_{\mathcal{T}}1} + \rho_{n_{\mathcal{T}}2} + \rho_{n_{\mathcal{T}}3} + \cdots + \rho_{n_{\mathcal{T}}n_{\mathcal{T}}} = 1 &&
\refstepcounter{subEquation}
\tag{\thesubEquation} \label{eq:weighted4}
\end{flalign}
\end{itemize}

\noindent Notice how we have introduced normalized ``attention weights''~$\rho_{\alpha \beta}$ as coefficients that define the weighted averages.  These coefficients link together different pairs of tokens according to their relative relevance to each other.  We will discuss \hyperlink{tok2tok}{later} how one may go beyond these second-order (pairwise token-to-token) interactions and consider higher-level token interactions that span the entire sequence. \\

\noindent Crucially, we have not yet specified how these weights are determined: This will be the topic of the next paragraph.  Nonetheless, before providing their exact expressions, we should already mention that we do not intend to freeze these coefficients to any permanent values. Instead, they should depend --- in a learnable way --- upon the specific tokens present in~${\bf a}_{i-1}$ and their pairwise relationships with one another.  In other words, we expect the attention weights to be a function parametrized by a set of head~$h$ specific parameters
\begin{equation*}
\rho_{\alpha \beta} \equiv \text{att}_{h} \, \big\langle \, {\bf a}_{i-1}(t=t_\alpha), \, {\bf a}_{i-1}(t=t_\beta) \, \big\rangle    
\end{equation*}
This means that the values of~$\rho_{\alpha \beta}$ will change, not only during training, as parameters associated with the head-specific attention function~$\text{att}_h$ are learned, but also during inference by dynamically assuming new values depending on the specific tokens~$t=t_\alpha$ and~$t=t_\beta$ in the input sample. \\

\noindent In other words, the output feature maps~${\bf a}_i^h$ are weighted averages of the values feature maps~${\bf v}_h$ where the attention coefficients~$\rho_{\alpha \beta}$ are data-dependent on~${\bf a}_{i-1}$.  Note that, at this point, we are considering only a single head~$h$ of self-attention. \\

\noindent Ultimately, we will consider multiple heads so that the self-attention function~$\text{att}_h$ will have different parameters for each head (although the overall form of the functional dependence of~$\rho_{\alpha \beta}$ on~${\bf a}_{i-1}$ will remain the same).  \\

\noindent Before moving on to specifying a functional form for the attention weights, let us organize them into a matrix representation~$\boldsymbol{\rho}_{({\bf a}_{i-1}, \, h)}$ where the subscript makes it clear that those coefficients should depend on the input sample~${\bf a}_{i-1}$ and on the specific self-attention head~$h$ . Since we have considered only pairwise~$\rho_{\alpha \beta}$ connections between the tokens, the full attention weight matrix is square~$\sim \mathbb{R}^{n_\mathcal{T}\times n_\mathcal{T}}$.  In addition, we have also enforced causality so the coefficients must respect~$\rho_{\alpha \beta} = 0$ if~$\beta > \alpha$.  In this case,~$\boldsymbol{\rho}_{({\bf a}_{i-1}, \, h)} \sim \mathbb{T}_L(\mathbb{R}^{n_\mathcal{T}\times n_\mathcal{T}}) $ is a square lower triangular matrix~\footnote{\hypertarget{causality}{This is a common situation for generative language models where a strict left-to-right order needs to be enforced such as, for example, in autoregressive text generation.  In contrast, architectures using encoders designed for language translation generally do not use causal masks so they are free to capture the overall relationships between words.  In this case, the attention weight matrix may also have a non-zero upper triangular part~$\boldsymbol{\rho}_{({\bf a}_{i-1}, \, h)} \sim \mathbb{R}^{n{_\mathcal{T}} \times n{_\mathcal{T}}}$. We will see this when we provide a \hyperlink{attnWeights}{general definition for attention weights}.}}. Using the matrix representation of~$\boldsymbol{\rho}_{({\bf a}_{i-1}, \, h)}$, the weighted averages defining the output sequence~${\bf a}_i^h$ of a self-attention layer can be expressed as

\begin{equation}
\tcboxmath[colframe=black, colback=gray!20, drop lifted shadow, boxsep=0pt, left=6pt, right=6pt]{
 {\bf a}_i^h \equiv \boldsymbol{\rho}_{({\bf a}_{i-1}, \, h) } \, {\bf v}_h } 
 = \left(
\begin{matrix}
\rho_{11}  & 0  & 0 & \cdots & 0 \\
\rho_{21}  & \rho_{22} & 0 & \cdots & 0 \\
\rho_{31}  & \rho_{32} & \rho_{33} & \cdots & 0 \\
\vdots & \vdots & \vdots & \ddots & \vdots \\
\rho_{n{_\mathcal{T}}1}  & \rho_{n{_\mathcal{T}}2} & \rho_{n{_\mathcal{T}}3} &  \cdots & \rho_{n{_\mathcal{T}} n{_\mathcal{T}}}
\end{matrix} \right)
\left(
\begin{array}{ccc}
\horzbar & {\bf v}_h(t=1)  & \horzbar \\
\horzbar & {\bf v}_h(t=2)  & \horzbar \\
\horzbar & {\bf v}_h(t=3)  & \horzbar \\
 & \vdots & \\
\horzbar & {\bf v}_h({t={n_\mathcal{T}}}) & \horzbar
\end{array}
\right)
\label{eq:finalDesiredAout}
\end{equation}

\noindent One can verify via straightforward expansion of~eq.\eqref{eq:finalDesiredAout}, that we exactly recover the expected expressions~${\bf a}_i^h(t=1), \cdots , {\bf a}_i^h(t=n_\mathcal{T})$ for the causal weighted averages given by~eqs.\eqref{eq:weighted1}–\eqref{eq:weighted4}, see~eq.\eqref{eq:rowLinearMix} for an explicit derivation.

\paragraph{Defining the attention weights.}  Having worked our way backwards starting with the desired output representation~${\bf a}_i^h$, we are now in a position to define the attention function~$\text{att}_h$ at the heart of the attention weight matrix~$\boldsymbol{\rho}_{({\bf a}_{i-1}, h)}$. \\

\noindent Let us consider two tokens~$t_\alpha$ and~$t_\beta$ and their feature maps~${\bf a}_{i-1}(t=t_\alpha) \sim \mathbb{R}^d$ and~${\bf a}_{i-1}(t=t_\beta) \sim \mathbb{R}^d$ from the input sequence~${\bf a}_{i-1}$.  Ideally, we would like their pairwise attention weight~$\rho_{\alpha \beta} \sim \mathbb{R}$ to quantify the level of ``relevance'' these tokens have for each other.  How to define a good expression for~$\rho_{\alpha \beta}$?
\begin{equation*}
\rho_{\alpha \beta} = \text{att}_h \, \langle \, {\bf a}_{i-1}(t=t_\alpha), \, {\bf a}_{i-1}(t=t_\beta) \, \rangle = \,\, ?
\end{equation*}
\noindent Rather than diving directly into the complete formulation, let us explore various possible implementations, examine their limitations and gradually build a deeper understanding for the definition of proper attention weights.

\renewcommand{\labelenumi}{\EnumItems{\arabic{enumi}}}

\begin{enumerate}
\item \uwave{As a first attempt}, one might be tempted to define the attention weights as a direct vector dot-product~$\rho_{\alpha \beta} \stackrel{?}{\equiv} {\bf a}_{i-1}(t=t_\alpha) \cdot {\bf a}_{i-1}(t=t_\beta)$.  Although this definition ensures that~$\rho_{\alpha \beta} \sim 1$ when the input feature maps are aligned with each other (and~$\rho_{\alpha \beta} \sim 0$ when they are orthogonal to each other), one issue with this choice is that the resulting attention weights would be fixed by the initial feature maps without the possibility of learning from data.  (Those initial feature maps would most likely even be random or determined by pretrained embeddings.)
\item \uwave{In our second attempt}, we solve this problem of static attention weights by introducing a fully-connected layer with adjustable parameters~$\{ {\bf w}_{q_h} \sim \mathbb{R}^{d \times d_\rho}, {\bf b}_{q_h} \sim \mathbb{R}^{d_\rho} \}$.  The input token feature maps are transformed into so-called ``query'' feature maps in~$d_\rho$ dimensions
\begin{equation*}
\begin{NiceMatrix}
\big( {\bf a}_{i-1}(t=t_\alpha) \sim \mathbb{R}^d \, , \, {\bf a}_{i-1}(t=t_\beta) \sim \mathbb{R}^d \, \big) \,\,\,\, & \,\,\,\,  \{ {\bf w}_{q_h} , {\bf b}_{q_h} \} \,\,\,\, & \,\,\,\, \big( {\bf q}_h(t=t_\alpha) \sim \mathbb{R}^{d_\rho} \, , \,{\bf q}_h(t=t_\beta) \sim \mathbb{R}^{d_\rho} \big)
\CodeAfter
  \begin{tikzpicture}[->]
  \draw[double distance=1pt] (1-1.east) -- (1-2.west) ;
  \draw[double distance=1pt] (1-2.east) -- (1-3.west) ;
  \end{tikzpicture}
\end{NiceMatrix}
\end{equation*}
We can now try to define the attention weights as the dot product~$\rho_{\alpha \beta} \stackrel{?}{\equiv} {\bf q}_h(t=t_\alpha) \cdot {\bf q}_h(t=t_\beta)$.  Using this revised tentative dot-product, the attention weights would be free to evolve as the parameters~${\bf w}_{q_h}$ and~${\bf b}_{q_h}$ are updated during training.  However, this definition of attention weights would still be rather restrictive as it enforces an undesired symmetric relationship between the tokens since~$\rho_{\alpha \beta} = \rho_{\beta \alpha}$.  Ideally, we would like to define attention weights in a way that takes into account the potentially asymmetric nature of token relationships~\cite{motivatingSelf}.
\item \EnumHighlight \uwave{For our third attempt}, we look for a definition of token-to-token attention that goes beyond simple symmetric similarity and that, instead, allows~$\rho_{\alpha \beta} \neq \rho_{\beta \alpha}$. This can be achieved by introducing an additional fully-connected layer parametrized by~$\{ {\bf w}_{k_h} \sim \mathbb{R}^{d \times d_\rho}, {\bf b}_{k_h} \sim \mathbb{R}^{d_\rho}\}$ of matching dimensionality to~$\{{\bf w}_{q_h} \sim \mathbb{R}^{d \times d_\rho}, {\bf b}_{q_h} \sim \mathbb{R}^{d_\rho}\}$ which was introduced in the previous attempt.  Each token would now be associated with two different and independent representations, so-called ``queries'' and ``keys'':
\begin{equation*}
\begin{NiceMatrix}
 & \,\, \{ {\bf w}_{q_h} , {\bf b}_{q_h} \} \,\, & \,\, \big( {\bf q}_h(t=t_\alpha) \sim \mathbb{R}^{d_\rho} \, , {\bf q}_h(t=t_\beta) \sim \mathbb{R}^{d_\rho} \big)  \\
\big( {\bf a}_{i-1}(t=t_\alpha) \sim \mathbb{R}^d \, , \, {\bf a}_{i-1}(t=t_\beta) \sim \mathbb{R}^d \, \big) \,\,  &    &   \\
 & \,\, \{ {\bf w}_{k_h} , {\bf b}_{k_h} \} \,\, & \,\, \big( {\bf k}_h(t=t_\alpha) \sim \mathbb{R}^{d_\rho} \, , {\bf k}_h(t=t_\beta) \sim \mathbb{R}^{d_\rho} \big)
\CodeAfter
  \begin{tikzpicture}[->]
  \draw[double distance=1pt] (2-1.east) -- (1-2.west) ;
  \draw[double distance=1pt] (2-1.east) -- (3-2.west) ;
  \draw[double distance=1pt] (1-2.east) -- (1-3.west) ;
  \draw[double distance=1pt] (3-2.east) -- (3-3.west) ;
  \end{tikzpicture}
\end{NiceMatrix}
\end{equation*}
This allows us to define the attention weights between two tokens~$t_\alpha$ and~$t_\beta$ as the dot-product 
\begin{equation}
\rho_{\alpha \beta} = {\bf q}_h(t=t_\alpha) \cdot {\bf k}_h(t=t_\beta) \sim \mathbb{R}
\label{eq:pairAttentionWeights}
\end{equation}
Since the parameters of the queries and keys are different from each other~$\{ {\bf w}_{q_h} , {\bf b}_{q_h} \} \neq \{ {\bf w}_{k_h} , {\bf b}_{k_h} \}$, we now have broken the symmetry and attention weights are such that~$\rho_{\alpha \beta} \neq \rho_{\beta \alpha}$. \\ 

\noindent It is this expression for quantifying the degree of pairwise attention~$\rho_{\alpha \beta}$ between two tokens that has emerged as the preferred candidate for self-attention.  Other, even more general, formulations may also be proposed (such as multiplicative bilinear attention, see~\cite{makingSenseSelf} for an easy informal review) but the dot-product presented in~eq.\eqref{eq:pairAttentionWeights} remains a favorite among practitioners. 
\end{enumerate}

\let\labelenumi\originalLabelenumi\textbf{}

\noindent Now that we have agreed upon a formulation using~eq.\eqref{eq:pairAttentionWeights} for attention weights, we need to evaluate~$\rho_{\alpha \beta}$ for all possible pairwise token-to-token combinations in~${\bf a}_{i-1}$.  Therefore, going beyond just two tokens, the complete set of~$n_\mathcal{T}^2$ attention weights requires first the creation of two independent linear transformations of the input tokens~${\bf a}_{i-1}$ into~queries~${\bf q}_h$ and keys~${\bf k}_h$ using two fully-connected layers:

\par\noindent\begin{minipage}[c][][c]{0.9\textwidth}
\begin{equation*}
\begin{NiceMatrix}
& \,\, \{ {\bf w}_{q_h} , {\bf b}_{q_h}  \} \,\, & \,\, {\bf q}_h = 
\left( \,
\begin{matrix}
{\bf q}_h(t=1) \sim \mathbb{R}^{d_\rho} \\
\vdots \\
{\bf q}_h(t=n_{\mathcal{T}}) \sim \mathbb{R}^{d_\rho}
\end{matrix} \,
\right) \sim \mathbb{R}^{n_{\mathcal{T}} \times d_\rho} & \text{(queries)} \\ 
 {\bf a}_{i-1}  = 
\left( \,
\begin{matrix}
{\bf a}_{i-1}(t=1) \sim \mathbb{R}^{d} \\
\vdots \\
{\bf a}_{i-1}(t=n_{\mathcal{T}})  \sim \mathbb{R}^{d}
\end{matrix} \,
\right) \sim \mathbb{R}^{n_{\mathcal{T}} \times d}
& \multicolumn{2}{c}{\text{where} \,\, {\bf w}_{q_h} \sim {\bf w}_{k_h} \sim \mathbb{R}^{d\times d_\rho} \,\, \text{and} \,\, {\bf b}_{q_h} \sim {\bf b}_{k_h} \sim \mathbb{R}^{d_\rho} } \\
& \,\, \{ {\bf w}_{k_h} , {\bf b}_{k_h}  \} \,\, & \,\, {\bf k}_h  = 
\left( \,
\begin{matrix}
{\bf k}_h(t=1) \sim \mathbb{R}^{d_\rho} \\
\vdots \\
{\bf k}_h(t=n_{\mathcal{T}}) \sim \mathbb{R}^{d_\rho}
\end{matrix} \,
\right) \sim \mathbb{R}^{n_{\mathcal{T}} \times d_\rho} & \text{(keys)} \\
\CodeAfter
  \begin{tikzpicture}[->]
  \draw[double distance=1pt] ([xshift=-40pt, yshift=-5pt]2-1.north east) -- (1-2.west) ;
  \draw[double distance=1pt] ([xshift=-40pt, yshift=5pt]2-1.south east) -- (3-2.west) ;
  \draw[double distance=1pt] (1-2.east) -- (1-3.west) ;
  \draw[double distance=1pt] (3-2.east) -- (3-3.west) ;
  \end{tikzpicture}
\end{NiceMatrix}
\end{equation*}
\end{minipage}
\begin{minipage}[c][][c]{0.1\textwidth}
\begin{align}
\refstepcounter{equation}
\tag{\theequation}\label{eq:queriesKeys}  
\end{align}
\end{minipage}

\begin{empheq}[box={\forwardBox[{\bf Queries and Keys}: forward pass]}]{alignat=2}
{\bf q}_h &= {\bf a}_{i-1} {\bf w}_{q_h} + \widetilde{{\bf b}_{q_h}} \refstepcounter{equation}
\setcounter{subEquation}{0}
\refstepcounter{subEquation}
\tag{\thesubEquation} \label{queries:forward} \\
{\bf k}_h &= {\bf a}_{i-1} {\bf w}_{k_h} + \widetilde{{\bf b}_{k_h}}
\refstepcounter{subEquation}
\tag{\thesubEquation}
\label{keys:forward}
\end{empheq}

\noindent Note that, until now, each token in the input sequence has been processed independently by the fully connected layers.  This means that the feature maps in~${\bf q}_h$ and~${\bf k}_h$ still do not communicate or share information with each other, even within the queries or keys themselves, as the processing remains independent across tokens.  It is only once we perform the pairwise token-to-token vector dot-products~${\bf q}_h(t=t_\alpha)\cdot {\bf k}_h(t=t_\beta)$ between the tokens' query/key representations that their affinity/relationship with each other is materialized via their attention weights.  \\

\noindent For the sake of clarity, we are now adding a ``raw'' label~$\rho_{\alpha \beta}^\text{raw} \longleftarrow \rho_{\alpha \beta}$ to the attention weights defined in~eq.\eqref{eq:pairAttentionWeights} to indicate that these weights are still left as straightforward dot-products;  Normalization will be the topic of the next paragraph.  The complete set of raw attention weights are then collected together into a \hypertarget{attnWeights}{square} attention weight matrix~$\boldsymbol{\rho}_{({\bf a}_{i-1}, \, h) }^\text{raw} \sim \mathbb{R}^{ {n_\mathcal{T}} \times {n_\mathcal{T}}  }$ defined as
\begin{equation*}
\hspace{0.7cm}
\makebox[\displaywidth]{$\displaystyle
\boldsymbol{\rho}_{({\bf a}_{i-1}, \, h) }^\text{raw} \equiv 
\begingroup 
\setlength\arraycolsep{1.5pt}
\begin{pmatrix}
{\bf q}_h(t=1) \cdot {\bf k}_h(t=1) & {\bf q}_h(t=1) \cdot {\bf k}_h(t=2)   &  {\bf q}_h(t=1) \cdot {\bf k}_h(t=3)  & \cdots &  {\bf q}_h(t=1) \cdot {\bf k}_h(t=n_\mathcal{T})  \\
 {\bf q}_h(t=2) \cdot {\bf k}_h(t=1)  &  {\bf q}_h(t=2) \cdot {\bf k}_h(t=2)  &  {\bf q}_h(t=2) \cdot {\bf k}_h(t=3) & \cdots & {\bf q}_h(t=2) \cdot {\bf k}_h(t={n_\mathcal{T}}) \\
 {\bf q}_h(t=3) \cdot {\bf k}_h(t=1) & {\bf q}_h(t=3) \cdot {\bf k}_h(t=2) & {\bf q}_h(t=3) \cdot {\bf k}_h(t=3)  & \cdots &  {\bf q}_h(t=3) \cdot {\bf k}_h(t={n_\mathcal{T}}) \\
\vdots & \vdots & \vdots & \ddots & \vdots \\
 {\bf q}_h(t={n_\mathcal{T}}) \cdot {\bf k}_h(t=1)  & {\bf q}_h(t={n_\mathcal{T}}) \cdot {\bf k}_h(t=2) &  {\bf q}_h(t={n_\mathcal{T}}) \cdot {\bf k}_h(t=3) &  \cdots & {\bf q}_h(t={n_\mathcal{T}}) \cdot {\bf k}_h(t={n_\mathcal{T}})
\end{pmatrix}
\endgroup
    $}
\end{equation*}

\noindent To conclude, the raw (unnormalized) attention weights~$\boldsymbol{\rho}_{({\bf a}_{i-1}, \, h) }^\text{raw}$ are expressed as a matrix product between the queries and the keys
\begin{empheq}[box={\mymath[colback=Gray,drop lifted shadow]}]{equation*} 
\makebox[0.97\displaywidth]{$\displaystyle
\begin{array}{c@{\hspace{0.1cm}}c}
\phantom{zzzzzzzzzzzzzzzzzzzzzzzz} {\bf k}_h^t \sim \mathbb{R}^{d_\rho \times n_\mathcal{T} } \equiv   & 
 \left(
\begin{array}{ccccc}
\vertbar & \vertbar &  \vertbar &  & \vertbar \\
{\bf k}_h(t=1)  & {\bf k}_h(t=2)  & {\bf k}_h(t=3) & \ldots & {\bf k}_h(t={n_\mathcal{T}})   \\
\vertbar & \vertbar &   \vertbar   &    & \vertbar 
\end{array}
\right) \\[0.1cm]
{\bf q}_h \sim \mathbb{R}^{n_\mathcal{T}\times d_\rho} \equiv
\left(
\begin{array}{ccc}
\horzbar & {\bf q}_h(t=1)  & \horzbar \\
\horzbar & {\bf q}_h(t=2)  & \horzbar \\
\horzbar & {\bf q}_h(t=3)  & \horzbar \\
 & \vdots & \\
\horzbar & {\bf q}_h({t={n_\mathcal{T}}}) & \horzbar
\end{array}
\right)
& \Longrightarrow \quad \quad \boldsymbol{\rho}^\text{raw}_{({\bf a}_{i-1}, \, h) } = {\bf q}_h \, {\bf k}_h^t \sim  \mathbb{R}^{n_\mathcal{T} \times n_\mathcal{T}}
\end{array}
   $}
\end{empheq}

\begin{empheq}[box={\forwardBox[{\bf Raw attention weights}: forward pass]}]{alignat=2}
\boldsymbol{\rho}^\text{raw}_{({\bf a}_{i-1}, \, h) } &= {\bf q}_h \, {\bf k}_h^t
\label{eq:rawAttnDef}
\end{empheq}

\noindent A few observations before we move on the final steps
\begin{itemize}
\item  The construction of the self-attention matrix~$\boldsymbol{\rho}^\text{raw}_{({\bf a}_{i-1}, \, h) }$ is the only place (with the exception of a simpler softmax normalization that will be described the next paragraph) where the inter-token relationships is actually exploited via the pairwise attention weights.  All other computations, such as queries, keys and values in the self-attention layer operate on tokens as independent entities.  By definition, each \hypertarget{attenVec}{row} of~$\boldsymbol{\rho}^\text{raw}_{({\bf a}_{i-1}, \, h) } \sim \mathbb{R}^{n_\mathcal{T} \times n_\mathcal{T}}$ consists of a vector
\begin{equation}
\boldsymbol{\rho}_{({\bf a}_{i-1}, \, h) }^\text{raw} (t={t^\star}) = \big[ \rho_{t{^\star}1}^\text{raw} \, , \cdots , \rho_{t{^\star} n_\mathcal{T} }^\text{raw} \big] \sim \mathbb{R}^{n_\mathcal{T}} 
\label{eq:attnVecDef}
\end{equation}
that contains the~$n_\mathcal{T}$ attention weights of a specific token~$t^\star$ with all the other tokens in the sequence.  Choosing rows for this definition of attention weight vectors allows us to connect back with the expressions of the desired causal output in~eq.\eqref{eq:finalDesiredAout} and this choice is standard convention in the literature.
\item Although this definition of attention weights with~eq.\eqref{eq:pairAttentionWeights} does not show an explicit dependence of~$\boldsymbol{\rho}^\text{raw}_{({\bf a}_{i-1}, \, h) }$ on the input data~${\bf a}_{i-1}$ and specific head~$h$, these dependencies are implicit through the way that queries~${\bf q}_h$ are keys~${\bf k}_h$ are built via fully-connected layers applied to~${\bf a}_{i-1}$ using head-specific parameters~$\{ {\bf w}_{q_h} , {\bf b}_{q_h} \}$ and~$\{ {\bf w}_{k_h} , {\bf b}_{k_h} \}$.
\item These attention weights based on token-to-token dot-product means that a \hypertarget{tok2tok}{single} layer of self-attention can only model up to two-token relationships.  We will discuss at the \hyperlink{compositionLayers}{end} of this Section how composing together multiple layers of self-attention allows one to model higher-level relationships between the tokens.
\item A common choice, for practical convenience and optimization benefits, is to choose~$d_\rho = d_h$ so that the dimensionality of the feature maps for the queries~${\bf q}_h$, keys~${\bf k}_h$ and values~${\bf v}_h$ are all the same.  This decision is widely applied in general-purpose deep learning libraries.  Nonetheless, it is important to realize that this alignment of dimensions is incidental rather than a fundamental restriction.  In fact, it is even possible to have different dimensionality for different attention heads, i.e.~$d_\rho = d_\rho(h)$ since all heads are independent of each other.  The only requirement is that queries and keys within the same head have the same dimensionality so that we can carry out their vector dot-product to define attention weights.  Regardless of the choice for~$d_\rho(h)$, in the end, the complete attention matrix will always be square~$\boldsymbol{\rho}^\text{raw}_{({\bf a}_{i-1}, \, h) } \sim \mathbb{R}^{{n_\mathcal{T}}\times {n_\mathcal{T}}}$ with scalar components~$\rho_{\alpha \beta}^\text{raw} \sim \mathbb{R}$.
\end{itemize}

\paragraph{Final enhancements.}  We have now gone over the heart of self-attention and how to define the attention matrix~$\boldsymbol{\rho}^\text{raw}_{({\bf a}_{i-1}, \, h) }$.  The only few steps left are some small enhancements to turn these raw attention weights into a more appropriate version~$\boldsymbol{\rho}_{({\bf a}_{i-1}, \, h) }$ which we will use as our final attention weight matrix to produce the weighted averages of the value feature maps (where we started from in the first paragraph). \\

\noindent First, the raw attention weights~$\rho_{\alpha \beta}^\text{raw}$ are defined as a dot-product between two~$d_\rho$-dimensional vectors.  If we assume that the components of these keys and queries feature map values are distributed according to a normal distribution (i.e. ${\bf q}_h(t=t_\alpha) \sim \left[ \mathcal{N}(0,1) , \cdots, \mathcal{N}(0,1) \right]$ and~$ {\bf k}_h(t=t_\beta))  \sim \left[ \mathcal{N}(0,1) , \cdots, \mathcal{N}(0,1) \right]$) then, the expected mean of their product~$\rho_{\alpha \beta}^\text{raw}$ should be~$ \langle \mu ( \rho_{\alpha \beta}^\text{raw}) \rangle = 0$ and their expected variance is~$\langle \sigma^2(\rho_{\alpha \beta}^\text{raw}) \rangle = d_\rho$.  To remove the dependence of the statistics of attention weights on the dimensionality~$d_\rho$ of the queries/keys feature maps (since~$d_\rho$ is an internal detail of the self-attention mechanism which does not appear in either~${\bf a}_{i-1}$ or~${\bf a}_i^h$), we rescale the attention weights~$\rho_{\alpha \beta}^\text{scaled} = \rho_{\alpha \beta}^\text{raw} /\sqrt{d_\rho}$ so that we now have~$\langle \sigma^2(\rho_{\alpha \beta}^\text{scaled}) \rangle = 1$.  In summary, we introduce a scaled version of the attention weights
\begin{equation}
\boldsymbol{\rho}^\text{scaled}_{({\bf a}_{i-1}, \, h) } = \boldsymbol{\rho}^\text{raw}_{({\bf a}_{i-1}, \, h) } \, / \sqrt{d_\rho}
\label{eq:sqrtScaling}
\end{equation}

\noindent Second, for the sake of this paper, we decided to focus on causal models where a strict left-to-right order must be enforced.  This is implemented by introducing a masking matrix~${\bf m}\sim  \mathbb{R}^{n_\mathcal{T} \times n_\mathcal{T} }$ of the same dimensionality as the attention  matrix and populated by binary 1/0 components in the lower triangular part so that~${\bf m} \sim \mathbb{T}_L ( \mathbb{R}^{{n_\mathcal{T}}\times {n_\mathcal{T}}})$ and~$\rho_{\alpha \beta}^\text{scaled} = 0$ if~$\beta>\alpha$.  The causal attention weights are therefore given by
\begin{equation}
\boldsymbol{\rho}^\text{causal}_{({\bf a}_{i-1}, \, h) } = {\bf m} \circ \boldsymbol{\rho}^\text{scaled}_{({\bf a}_{i-1}, \, h) }
\label{eq:mask}    
\end{equation}
which is expressed more explicitly as
\begin{equation*}
\left(
\begin{matrix}
\rho_{11}  & 0  & 0 & \cdots & 0 \\
\rho_{21}  & \rho_{22} & 0 & \cdots & 0 \\
\rho_{31}  & \rho_{32} & \rho_{33} & \cdots & 0 \\
\vdots & \vdots & \vdots & \ddots & \vdots \\
\rho_{n{_\mathcal{T}}1}  & \rho_{n{_\mathcal{T}}2} & \rho_{n{_\mathcal{T}}3} &  \cdots & \rho_{n{_\mathcal{T}} n{_\mathcal{T}}}
\end{matrix} \right)_\text{causal} =
\left(
\begin{matrix}
1  & 0  & 0 & \cdots & 0 \\
1  & 1 & 0 & \cdots & 0 \\
1  & 1 & 1 & \cdots & 0 \\
\vdots & \vdots & \vdots & \ddots & \vdots \\
1  & 1 & 1 &  \cdots & 1
\end{matrix} \right) \circ
\left(
\begin{matrix}
\rho_{11}  & \rho_{12}  & \rho_{13} & \cdots & \rho_{1 n_\mathcal{T}} \\
\rho_{21}  & \rho_{22} & \rho_{23} & \cdots & \rho_{2 n_\mathcal{T}} \\
\rho_{31}  & \rho_{32} & \rho_{33} & \cdots & \rho_{3 n_\mathcal{T}} \\
\vdots & \vdots & \vdots & \ddots & \vdots \\
\rho_{n{_\mathcal{T}}1}  & \rho_{n{_\mathcal{T}}2} & \rho_{n{_\mathcal{T}}3} &  \cdots & \rho_{n{_\mathcal{T}} n{_\mathcal{T}}}
\end{matrix} \right)_\text{scaled} 
\end{equation*}

\noindent where~$\circ$ stands for the Hadamard product and we see that we recover an attention matrix with the same causal shape the one as described in~eq.\eqref{eq:finalDesiredAout}. \\

\noindent Finally, we ensure a proper normalization of the attention weights to close the loop and completely recover the desired output weighted averages initially discussed in~eqs.\eqref{eq:weighted1}–\eqref{eq:weighted4} (i.e., not just the same causal shape but also the normalization constraint discussed there).  As discussed \hyperlink{attenVec}{previously}, each row of the attention matrix~$\sim \mathcal{R}^{n_\mathcal{T}\times n_\mathcal{T}}$ contains the~$n_\mathcal{T}$ attention vectors~$\sim \mathcal{R}^{n_\mathcal{T}}$ for all the tokens in the sequence. By analogy with~eq.\eqref{eq:attnVecDef}, let us consider a specific token~$t^\star$ and its causal attention weight vector
\begin{equation*}
\boldsymbol{\rho}_{({\bf a}_{i-1}, \, h) }^\text{causal} (t={t^\star}) = \big[ \rho_{t{^\star}1}^\text{causal} \, , \cdots , \rho_{t{^\star} n_\mathcal{T}}^\text{causal}  \big] \sim \mathbb{R}^{n_\mathcal{T}}  
\end{equation*}
whose components quantify the  attention scores between token~$t^\star$ and all the other~$n_\mathcal{T}$ tokens in the sequence. Causality means that, depending on the position of~$t^\star \in [ 1, \cdots , n_\mathcal{T} ]$ in the sequence, some of its attention scores would be identically null as prescribed by the mask~${\bf m}$ above.  Applying a softmax function to this causal attention weight vector produces a probability distribution
\begin{equation*}
\boldsymbol{\rho}_{({\bf a}_{i-1}, \, h) } (t={t^\star}) = \big[ \rho_{t{^\star}1} \, , \cdots , \rho_{t{^\star} n_\mathcal{T}} \big] \equiv \text{softmax} \, \boldsymbol{\rho}_{({\bf a}_{i-1}, \, h) }^\text{causal} (t={t^\star}) \sim \mathbb{R}^{n_\mathcal{T}}
\end{equation*}
where the components are normalized such that
\begin{equation}
\sum \boldsymbol{\rho}_{({\bf a}_{i-1}, \, h) } (t={t^\star}) = \sum_{t^\prime=1}^{n_\mathcal{T}} \rho_{t^\star t^\prime} = 1 
\label{eq:normalization}
\end{equation}

\noindent Repeating the same softmax normalization to all the causal attention weight vectors, i.e. the rows of~$\boldsymbol{\rho}_{({\bf a}_{i-1}, \, h) }^\text{causal}$, leads to the final expression for the self-attention matrix~$\boldsymbol{\rho}_{({\bf a}_{i-1}, \, h) } \sim \mathbb{R}^{n_\mathcal{T} \times n_\mathcal{T}}$
\begin{equation} \hspace{-0.5cm}
\left( \,
\begin{matrix}
\horzbar \,\,\boldsymbol{\rho}_{({\bf a}_{i-1}, \, h) }(t=1) \,\, \horzbar \\
    \vdots \\
\horzbar \,\, \boldsymbol{\rho}_{({\bf a}_{i-1}, \, h) }(t=n_\mathcal{T})  \,\, \horzbar
\end{matrix} \,
\right) \equiv \left(
\begin{matrix} 
\text{softmax} \, \boldsymbol{\rho}_{({\bf a}_{i-1}, \, h) }^\text{causal} (t=1)  \\
\vdots   \\
\text{softmax} \, \boldsymbol{\rho}_{({\bf a}_{i-1}, \, h) }^\text{causal}  (t=n_\mathcal{T}) 
\end{matrix} \right)
\Longrightarrow \, \tcboxmath[colframe=black, colback=gray!20, drop lifted shadow, boxsep=0pt, left=4pt, right=6pt]{
\boldsymbol{\rho}_{({\bf a}_{i-1}, \, h) } = \text{softmax} \,\boldsymbol{\rho}^\text{causal}_{({\bf a}_{i-1}, \, h) } 
}
\label{eq:normalizedAttnScores}
\end{equation}

\noindent In addition to producing normalized probability distributions (offering direct interpretable attention allocation), softmax normalization is known to also be associated with beneficial implicit regularization mechanisms~\cite{softmaxRegulator}.  

\paragraph{Summary and some remarks.}  We can now present the output representation
\begin{equation}
{\bf a}_i^h = \boldsymbol{\rho}_{({\bf a}_{i-1}, \, h) } \, {\bf v}_h \sim \mathbb{R}^{n_\mathcal{T}\times d_h}
\label{eq:finalAout}
\end{equation}
initially proposed in~eq.\eqref{eq:finalDesiredAout} where each step in the construction has been explained. (As a special case, if we restrict the attention weight matrix to be the identity matrix, then the output of the self-attention head reduces to the values feature maps with~${\bf a}_i^h \equiv {\bf v}^h$; This makes sense as tokens only attend to themselves with such attention weights.)  \\

\noindent In summary, an attention head~$h$ can be seen as a function parametrized by~$\mathcal{P}_h$ that takes in as input arguments the~$d$-dimensional feature maps of the~$n_\mathcal{T}$ tokens~${\bf a}_{i-1}\sim \mathbb{R}^{n_\mathcal{T}\times d}$ and returns their transformed $d_h$-dimensional representations~${\bf a}_i^h \sim \mathbb{R}^{n_\mathcal{T}\times d_h}$ with the following signature
\begin{equation}
\text{Att}_{\mathcal{P}_h} \, : \,   {\bf a}_{i-1} \sim \mathbb{R}^{n_\mathcal{T}\times d}  \,\, \longrightarrow \,\, {\bf a}_i^h \sim \mathbb{R}^{n_\mathcal{T}\times d_h} \,\,\,\, \text{with }\, \mathcal{P}_h \equiv \begin{cases}
{\bf w}_{q_h} \sim \mathcal{R}^{d\times d_\rho} \,\,\,\, ; \,\,\, {\bf b}_{q_h} \sim \mathcal{R}^{d_\rho} \\
{\bf w}_{k_h} \sim \mathcal{R}^{d\times d_\rho} \\
{\bf w}_{v_h} \sim \mathcal{R}^{d_h} \,\,\,\,\,\,\,\,\,\, ; \,\,\, {\bf b}_{v_h} \sim \mathcal{R}^{d_h}
\end{cases}
\label{eq:attHfunction}
\end{equation}
where we intentionally ignored the biases~${\bf b}_{k_h}$ of the keys (we will see below that self-attention does not depend on those) and where the exact expression for~${\bf a}_i^h = \text{Att}_{\mathcal{P}_h} ({\bf a}_{i-1})$ is given by
\begin{empheq}[box={\forwardBox[{\bf Self-attention}: forward pass]}]{alignat=2}
{\bf a}_i^h &= \text{softmax} \Bigg( {\bf m} \circ \dfrac{{\bf q}_h \, {\bf k}_h^t}{ \sqrt{d_\rho}} \Bigg) \, {\bf v}_h
\label{selfAttention:forward}
\end{empheq}
where the~queries and keys~${\bf q}_h \sim {\bf k}_h \sim \mathbb{R}^{n_\mathcal{T} \times d_\rho}$ depend on~${\bf a}_{i-1}$ via the fully-connected layers expressed in~eqs.\eqref{queries:forward}-\eqref{keys:forward} and the values~${\bf v}_h \sim \mathbb{R}^{n_\mathcal{T} \times d_h}$ also depend on~${\bf a}_{i-1}$ via another fully-connected layer given by~eq.\eqref{values:forward}. \\

\noindent Before moving on to the backward pass, let us make a few general observations about self-attention:

\paragraph{Computational complexity.}  By virtue of its own definition as pairwise dot-products between tokens, the attention matrix~$\boldsymbol{\rho}_{({\bf a}_{i-1}, \, h) } \sim {\bf q}_h \, {\bf k}_h^t \sim \mathbb{R}^{n_\mathcal{T} \times n_\mathcal{T}}$ requires quadratic complexity with respect to the number~$n_\mathcal{T}$ of input tokens. This can be seen by looking at the computational complexity of the matrix product~$\mathcal{O}({\bf q}_h \, {\bf k}_h^t) \sim d_\rho n_\mathcal{T}^2$.  This quadratic scaling with the number of tokens appears also when evaluating the linear mixing of values feature maps with~$\mathcal{O}(\boldsymbol{\rho}_{({\bf a}_{i-1}, \, h) } \, {\bf v}_h) \sim d_h n_\mathcal{T}^2$.  Overall, this confirms that self-attention has a \hypertarget{selfAttQuad}{computational complexity} with grows quadratically with the number of tokens~$\mathcal{O}({\bf a}_i^h) \sim n_\mathcal{T}^2$.  This potential bottleneck has been discussed extensively in the literature and we refer the readers to external references and their citations for reviews of computational complexity~\cite{complexitySelf}, approximation methods~\cite{reformer, performer} and low-level optimization~\cite{flashAtt}.  We also discuss in a small \hyperlink{kvcacheNote}{side-note} here the technique of~KV~cache optimization.

\paragraph{Adjustable bias parameters.}  Practitioners rarely include bias terms such as~${\bf b}_{q_h} \sim {\bf b}_{k_h} \sim \mathbb{R}^{d_\rho}$ and~${\bf b}_{v_h} \sim \mathbb{R}^{d_h} $ in self-attention layers: The original paper ignored them altogether~\cite{originalSelf}.  In fact, one can show that, because of the shift-invariance property of the softmax normalization, the self-attention layer as defined in~eq.\eqref{selfAttention:forward} does not even depend on the bias term~${\bf b}_{k_h}$ of the keys~\footnote{
This can be seen by explicitly expanding out the expressions of the queries~${\bf q}_h$ and keys~${\bf k}_h$ which define the normalized attention weights~$\boldsymbol{\rho}_{({\bf a}_{i-1}, \, h) }$ in~eq.\eqref{eq:normalizedAttnScores}
\begin{align}
\boldsymbol{\rho}_{({\bf a}_{i-1}, \, h) } \sim \text{softmax} \left( {\bf q}_h \, {\bf k}_h^t \right) &= \text{softmax} \left( \left( {\bf a}_{i-1} {\bf w}_{q_h} + \widetilde{\bf b}_{q_h} \right) \left( {\bf a}_{i-1} {\bf w}_{k_h} + \widetilde{\bf b}_{k_h} \right)^t \right) \nonumber \\
&= \text{softmax} \left( \left( {\bf a}_{i-1} {\bf w}_{q_h} + \widetilde{\bf b}_{q_h} \right) \left( {\bf a}_{i-1} {\bf w}_{k_h} \right)^t +  \left( {\bf a}_{i-1} {\bf w}_{q_h} + {\bf b}_{q_h} \right) \widetilde{{\bf b}_{k_h}}^t \right) \nonumber \\
&= \text{softmax} \left( \left( {\bf a}_{i-1} {\bf w}_{q_h} + \widetilde{\bf b}_{q_h} \right) \left( {\bf a}_{i-1} {\bf w}_{k_h} \right)^t  \right)
\label{eq:softMaxInvariance}
\end{align}
where the last equality shows that the dependence of~$\boldsymbol{\rho}_{({\bf a}_{i-1}, \, h) }$ on~${\bf b}_{k_h}$ completely disappears.  Essentially, it stems from the fact that any matrix~${\bf H}$ multiplied by the transpose of a {\bf broadcast} vector~${\bf b}_{k_h}$ produces a matrix~${\bf H} \widetilde{\bf b}_{k_h}^t$ where the rows are all constant.  Since the softmax normalization is shift-invariant, this constant shift of the rows cancels out with~$\text{softmax} ( {\bf G} + {\bf H} \widetilde{\bf b}_{k_h}^t ) = \text{softmax} \, {\bf G}$ so that the dependence on~${\bf b}_{k_h}$ drops out. This identity is proven in detail in~eq.\eqref{eq:softMaxId} of the appendix section.  Note that the apparent asymmetry between queries and keys is incidental.  In order to produce a square self-attention matrix, one of either keys or queries needs to be transposed and it is for this one that the dependence of~$\boldsymbol{\rho}_{({\bf a}_{i-1}, \, h) }$ on their bias term will drop out.  If one were to \hypertarget{QKreverse}{swap} the notation but continue to define attention weight vectors as the rows of~$\boldsymbol{\rho}_{({\bf a}_{i-1}, \, h) } \sim \text{softmax} ({\bf k}_{h} \, {\bf q}_{h}^t)$, in this case it would be the~${\bf b}_{q_h}$ that would become redundant.  Other than this, the self-attention layer would behave in exactly the same manner confirming that queries and keys are internal parameters used to define attention weights and remain completely interchangeable.}.  This means that the~$d_\rho$ parameters of~${\bf b}_{k_h}$ are ``impotent'', in the sense that have no influence over the output of self-attention and therefore over the loss function itself (we will even \hyperlink{gradBiasesNull2}{confirm} in the backward pass section that the derivative of the loss with respect to~${\bf b}_{k_h}$ is indeed identically \hypertarget{gradBiasesNull}{null} showing that those parameters cannot ever learn anything).  Therefore, the bias term~${\bf b}_{k_h}$ can be removed without any loss of generality.  Even though, this is not the case for the other bias terms~${\bf b}_{q_h}$ and~${\bf b}_{v_h}$ which do have valid contributions to self-attention, they are still frequently ignored by practitioners.

\paragraph{Composition of multiple layers of self-attention.}  \hypertarget{compositionLayers}{Finally,} let us discuss how one may go beyond pairwise token interactions by composing together multiple layers of self-attention.  For the sake of simplicity, let us take~$d_h = d$ so that the input~${\bf a}_{i-1}$ and output~${\bf a}_i^h$ of a self-attention head have the same dimensionality~${\bf a}_i^h \equiv {\bf a}_i \sim {\bf a}_{i-1} \sim \mathbb{R}^{n_\mathcal{T}\times d}$ (dropping the superscript~$h$ from~${\bf a}_{i}^h$ and from~$\mathcal{P}_h \equiv \mathcal{P}$ to lighten the notation).  Thanks to this dimensionality matching, one may compose together multiple attention heads, i.e. use the output representation as a new input.  We will see in~Section~\ref{sec:multiHead} how the same dimensionality matching may be achieved even when~$d_h \neq d$ by combining together multiple attention heads into a multi-headed attention layer.  In any case, the observation we make here remains conceptually identical for multi-headed attention albeit at the cost of tedious but superficial modifications (for example the superscript~$h$ would need to be kept to distinguish between different heads).   Starting with~${\bf a}_{i-1}$ and applying the attention head twice takes us through a series of token feature maps going from~${\bf a}_{i-1}$ on to~${\bf a}_i = \text{Att}_\mathcal{P}({\bf a}_{i-1})$ finishing with~${\bf a}_{i+1} = \text{Att}_\mathcal{P}({\bf a}_i)$ summarized as
\begin{equation*}
{\bf a}_{i+1} \sim \mathbb{R}^{n_\mathcal{T}\times d} = \text{Att}_{\mathcal{P}} \left( \text{Att}_\mathcal{P} \left( {\bf a}_{i-1} \sim \mathbb{R}^{n_\mathcal{T}\times d} \right) \sim \mathbb{R}^{n_\mathcal{T}\times d} \right) \sim \mathbb{R}^{n_\mathcal{T}\times d}
\end{equation*}

\noindent  Let us consider the final~${\bf a}_{i+1} \sim \mathbb{R}^{n_\mathcal{T}\times d}$ and focus on the feature vector~${\bf a}_{i+1}(t=t^\star) \sim \mathbb{R}^d$ of a specific token~$t^\star$.  Its representation as a weighted sum over all of the~$n_\mathcal{T}$ values' feature vectors in the sequence is defined by~eq.\eqref{eq:finalAout}, which is specialized to~$t^\star$ following~eq.\eqref{eq:weighedSumIndividual}, yielding 
\begin{equation*}
{\bf a}_{i+1}(t=t^\star) = \sum_{t_\alpha} \rho_{t^\star t_\alpha}^{(i)} {\bf v}_{(i)} (t=t_\alpha) = \sum_{t_\alpha} \rho_{t^\star t_\alpha}^{(i)} {\bf a}_{i} (t=t_\alpha) {\bf w}_v 
\end{equation*}
where we ignored the biases in~eq.\eqref{values:forward} (this does not affect the conclusions and just helps simplify the notation) and introduced an index~$(i)$ in the attention weights and values vectors to indicate that those are specific to~${\bf a}_i$.  Moving on, we now need an expression for~${\bf a}_{i} (t=t_\alpha)$, which following the same logic is expressed as a similar weighted sum
\begin{equation*}
{\bf a}_{i}(t=t_\alpha) = \sum_{t_\beta} \rho_{t_\alpha t_\beta}^{(i-1)} {\bf v}_{(i-1)} (t=t_\beta) = \sum_{t_\beta} \rho_{t_\alpha t_\beta}^{(i-1)} {\bf a}_{i-1} (t=t_\beta) {\bf w}_v
\end{equation*}

\noindent Putting all the pieces back together, we get
\begin{equation}
{\bf a}_{i+1}(t=t^\star) = \sum_{t_\alpha} \sum_{t_\beta} \rho_{t^\star t_\alpha}^{(i)}  \rho_{t_\alpha t_\beta}^{(i-1)} {\bf a}_{i-1} (t=t_\beta) {\bf w}_v^2
\label{eq:thirdOrder}
\end{equation}
thereby showing explicitly how the weighted sum that defines~${\bf a}_{i+1}$ after two layers of self-attention now involves third-order~$(t^\star, t_\alpha, t_\beta)$ token interactions instead of just pairwise relationships when we considered only a single layer of self-attention.   Composing even more layers together generates higher-order interactions eventually spanning the entire sequence.~\footnote{While on the topic of composition of deep learning layers, this is an occasion to observe, in practice, the importance of normalization (such as the softmax and scaling methods discussed above).  Indeed, as a high-level approximation, one can model the matrix products that appear in~eq.\eqref{eq:thirdOrder} as a product of random matrices.  Under this assumption, it may be shown that the variance of~${\bf a}_{i+1}$ is unbounded and grows exponentially with the number of products~\cite{RMT}.  Therefore, normalization layers are required to rescale the data representations and stabilize the training of the model.}

\subsection{Self-attention layer: Single head --- Backward pass}

\noindent Just like any other layer, the backward pass through a self-attention layer starts by evaluating the recursive backward error flow~$\boldsymbol{\Delta}^h_i \cdot \text{d}{\bf a}_i^h$ and gradient extraction described in~eq.(11) of the reference paper~\cite{deepPedestrians}.  Here~$\boldsymbol{\Delta}^h_i \sim \mathbb{R}^{n_\mathcal{T} \times d_h}$ represents the upstream error flow going into attention head~$h$ that was produced by layers closer to the loss function and~${\bf a}_i^h\sim \mathbb{R}^{n_\mathcal{T}\times d_h} $ represents the~$n_\mathcal{T}$ feature maps, each of dimensionality~$\sim \mathbb{R}^{d_h}$, produced by the same attention head~$h$. We defer the discussion of how the complete error signal~$\boldsymbol{\Delta}_i \sim \mathbb{R}^{n_\mathcal{T}\times d}$ is split for each attention head~$h$ to~Section~\ref{sec:multiHead}. \\

\noindent Given an attention head~$h$, its output data representation~${\bf a}_i^h = \boldsymbol{\rho}_{({\bf a}_{i-1}, \, h) } \, {\bf v}_h $ is given by a weighted average of the value feature maps~${\bf v}_h$ with the attention matrix~$\boldsymbol{\rho}_{({\bf a}_{i-1}, \, h) }$, see~eq.\eqref{eq:finalAout}.  Writing out the recursive backward error flow explicitly, we have
\begin{align*}
\boldsymbol{\Delta}^h_i \cdot \text{d}{\bf a}_i^h &= \boldsymbol{\Delta}^h_i \cdot  \big[ \big( \text{d} \boldsymbol{\rho}_{({\bf a}_{i-1}, \, h) } \big)  {\bf v}_h + \boldsymbol{\rho}_{({\bf a}_{i-1}, \, h) } \text{d} {\bf v}_h  \big] \\
&= \boldsymbol{\Delta}^h_i \cdot  \big( \text{d} \boldsymbol{\rho}_{({\bf a}_{i-1}, \, h) } \big)  {\bf v}_h +  \boldsymbol{\Delta}^h_i \cdot  \big( \boldsymbol{\rho}_{({\bf a}_{i-1}, \, h) } \text{d} {\bf v}_h \big) \\
&= \mathcolorbox{Gray}{\big( \boldsymbol{\Delta}^h_i {\bf v}_h^t \big) \cdot \text{d} \boldsymbol{\rho}_{({\bf a}_{i-1}, \, h) } } + \mathcolorbox{Gray}{\big(  \boldsymbol{\rho}_{({\bf a}_{i-1}, \, h) }^t \boldsymbol{\Delta}^h_i \big) \cdot \text{d} {\bf v}_h }
\end{align*}

\noindent At this point, the error flow \hypertarget{split1}{splits} into \colorbox{Gray}{\bf two different branches} due to the definition of~${\bf a}_i^h$ as a product between the self-attention weights~$\boldsymbol{\rho}_{({\bf a}_{i-1}, \, h) }$ and the feature maps of the values~${\bf v}_{h}$. \\

\noindent Since~${\bf v}_h$ is produced by a fully-connected layer with the input sequence~${\bf a}_{i-1}$, \colorbox{Gray}{this branch is terminal} as it already comes back to the source data of the self-attention layer.   Applying directly the formulas already established in Section~5 of the reference paper~\cite{deepPedestrians} for error signal propagation and gradient extraction through fully-connected layers, we immediately get
\begin{align*}
\mathcolorbox{Gray}{\big(  \boldsymbol{\rho}_{({\bf a}_{i-1}, \, h) }^t \boldsymbol{\Delta}^h_i \big) \cdot \text{d} {\bf v}_h} &= \big(  \boldsymbol{\rho}_{({\bf a}_{i-1}, \, h) }^t \boldsymbol{\Delta}^h_i \big) \cdot \text{d} \big( {\bf a}_{i-1} {\bf w}_{v_h} + \widetilde{{\bf b}_{v_h}} \big) \nonumber  \\
&= \underbrace{ \mathcolorbox{shadecolor}{{\bf a}_{i-1}^t \, \boldsymbol{\rho}_{({\bf a}_{i-1}, \, h) }^t \boldsymbol{\Delta}^h_i}}_{\textstyle
    \begin{gathered}
      \frac{\partial \mathcal{L}_{\text{seq}}}{\partial {\bf w}_{v_h}}
    \end{gathered} } \, \cdot \, \text{d} {\bf w}_{v_h} + \underbrace{ \mathcolorbox{shadecolor}{ \sum_\text{tokens} \boldsymbol{\rho}_{({\bf a}_{i-1}, \, h) }^t \boldsymbol{\Delta}^h_i } }_{\textstyle
    \begin{gathered}
      \frac{\partial \mathcal{L}_{\text{seq}}}{\partial {\bf b}_{v_h}}
    \end{gathered} } \cdot \, \text{d} {\bf b}_{v_h} + \underbrace{ \mathcolorbox{shadecolor}{\boldsymbol{\rho}_{({\bf a}_{i-1}, \, h) }^t \boldsymbol{\Delta}^h_i {\bf w}^t_{v_h} } }_{\textstyle
    \begin{gathered}
      \boldsymbol{\Delta}^h_{v_h}
    \end{gathered} } \cdot \, \text{d} {\bf a}_{i-1} 
\end{align*}

\noindent Because of the softmax normalization of~$\boldsymbol{\rho}_{({\bf a}_{i-1}, \, h) }$, the expression~${\partial \mathcal{L}_{\text{seq}}} / {\partial {\bf b}_{v_h}}$ above for the gradient for the biases can be simplified further.  Writing it out explicitly using the graphical representation of the transpose of the attention weight matrix, see.\eqref{eq:normalizedAttnScores}, we have
\begin{align}
\frac{\partial \mathcal{L}_{\text{seq}}}{\partial {\bf b}_{v_h}} &= \sum_\text{tokens} \left( \,
\begin{matrix}
\vertbar &  & \vertbar \\[0.4em]
\boldsymbol{\rho}_{({\bf a}_{i-1}, \, h) }(t=1) & \cdots &  \boldsymbol{\rho}_{({\bf a}_{i-1}, \, h) }(t=n_\mathcal{T}) \\
\vertbar & & \vertbar \\
\end{matrix} \,
\right) \boldsymbol{\Delta}^h_i \nonumber \\
&\centerwithin{\downarrow}{\,\,=} \colorbox{light-blue}{using eq.\eqref{eq:sumRows}} \nonumber  \\
&= \left( \sum_{t^\prime=1}^{n_\mathcal{T}} \rho_{1 t^\prime} \, , \cdots , \sum_{t^\prime=1}^{n_\mathcal{T}} \rho_{n_\mathcal{T} t^\prime} \right) \boldsymbol{\Delta}^h_i \nonumber \\
&\centerwithin{\downarrow}{\,\,=} \colorbox{light-blue}{because of the softmax normalization of attention weight vectors, see~eq.\eqref{eq:normalization}}. \nonumber  \\
&= ( 1 , \cdots , 1 ) \, \boldsymbol{\Delta}^h_i \nonumber \\
\frac{\partial \mathcal{L}_{\text{seq}}}{\partial {\bf b}_{v_h}} &= \sum_\text{tokens} \boldsymbol{\Delta}^h_i \sim \mathbb{R}^{d_h}
\label{eq:biasesValuesSimplified}
\end{align}
In other words, the gradients of the biases of the fully-connected layer that determines~${\bf v}_{h}$ do not depend on the attention weight matrix~$\boldsymbol{\rho}_{({\bf a}_{i-1}, \, h) }$ of attention head~$h$ but only on its (head-specific) allocated upstream error signal~$\boldsymbol{\Delta}^h_i$.  This is a direct consequence of the softmax normalization of the attention weight matrix.  We will see \hyperlink{gradientConstraint}{later} how the same constraint also impacts other bias gradients. \\

\noindent In summary, the backward pass through the value feature maps~${\bf v}_h$ branch of self-attention leads to error propagation and gradients given by the following expressions
\begin{empheq}[box={\backPropBox[{\bf Values}: backward pass]}]{alignat=2}
\boldsymbol{\Delta}^h_{v_h} &= \boldsymbol{\rho}_{({\bf a}_{i-1}, \, h) }^t \boldsymbol{\Delta}^h_i {\bf w}^t_{v_h} &\quad &\sim \mathbb{R}^{n_\mathcal{T} \times d} \label{grad:deltaVh} \\
\frac{\partial \mathcal{L}_{\text{seq}}}{\partial {\bf w}_{v_h}} &= \big( \boldsymbol{\rho}_{({\bf a}_{i-1}, \, h) } \, {\bf a}_{i-1} \big)^t \boldsymbol{\Delta}^h_i  &\quad  &\sim \mathbb{R}^{d \times d_h} \label{grad:Vw} \\
\frac{\partial \mathcal{L}_{\text{seq}}}{\partial {\bf b}_{v_h}} &= \sum_\text{tokens} \boldsymbol{\Delta}^h_i &\quad  &\sim \mathbb{R}^{d_h} \label{grad:Bw}
\end{empheq}

\noindent \\

\noindent We can now move on to the \colorbox{Gray}{second branch} related to backpropagation through the self-attention weight matrix~$\boldsymbol{\rho}_{({\bf a}_{i-1}, \, h) }$.  Repeating the expression already derived above and replacing~$\boldsymbol{\rho}_{({\bf a}_{i-1}, \, h) }$ by its definition as the softmax normalized version of~$\boldsymbol{\rho}^\text{causal}_{({\bf a}_{i-1}, \, h) }$, see~eq.\eqref{eq:normalizedAttnScores}, we have
{\allowdisplaybreaks
\begin{align}
\mathcolorbox{Gray}{\big( \boldsymbol{\Delta}^h_i {\bf v}_h^t \big) \cdot \text{d} \boldsymbol{\rho}_{({\bf a}_{i-1}, \, h) } }  &=  \big( \boldsymbol{\Delta}^h_i {\bf v}_h^t \big) \cdot \text{d} \big( \text{softmax} \,\boldsymbol{\rho}^\text{causal}_{({\bf a}_{i-1}, \, h) } \big) \nonumber \\
&\centerwithin{\downarrow}{\,\,=} \colorbox{light-blue}{using eq.(13) of the reference paper \cite{deepPedestrians}} \nonumber \\
&= \big( \boldsymbol{\Delta}^h_i {\bf v}_h^t \big) \cdot \Big[ \boldsymbol{\rho}_{({\bf a}_{i-1}, \, h) } \circ  \big( \text{d} \boldsymbol{\rho}^\text{causal}_{({\bf a}_{i-1}, \, h) } - \widetilde{ \boldsymbol{\rho}_{({\bf a}_{i-1}, \, h) } \ominus \text{d} \boldsymbol{\rho}^\text{causal}_{({\bf a}_{i-1}, \, h) } }   \big)  \Big] \nonumber \\
&= \big( \boldsymbol{\Delta}^h_i {\bf v}_h^t \big) \cdot \Big[ \boldsymbol{\rho}_{({\bf a}_{i-1}, \, h) } \circ \text{d} \boldsymbol{\rho}^\text{causal}_{({\bf a}_{i-1}, \, h) } - \boldsymbol{\rho}_{({\bf a}_{i-1}, \, h) } \circ \big( \widetilde{ \boldsymbol{\rho}_{({\bf a}_{i-1}, \, h) } \ominus \text{d} \boldsymbol{\rho}^\text{causal}_{({\bf a}_{i-1}, \, h) } \big) }  \Big] \nonumber \\
&= \big( \boldsymbol{\Delta}^h_i {\bf v}_h^t \big) \cdot \Big[ \boldsymbol{\rho}_{({\bf a}_{i-1}, \, h) } \circ \text{d} \boldsymbol{\rho}^\text{causal}_{({\bf a}_{i-1}, \, h) } \Big] - \big( \boldsymbol{\Delta}^h_i {\bf v}_h^t \big) \cdot\Big[\boldsymbol{\rho}_{({\bf a}_{i-1}, \, h) } \circ \big( \widetilde{ \boldsymbol{\rho}_{({\bf a}_{i-1}, \, h) } \ominus \text{d} \boldsymbol{\rho}^\text{causal}_{({\bf a}_{i-1}, \, h) } \big) }  \Big] \nonumber \\
&\centerwithin{\downarrow}{\,\,=} \colorbox{light-blue}{using eq.(53) of the reference paper \cite{deepPedestrians} on the left-most term} \nonumber \\
&= \big( \boldsymbol{\Delta}^h_i {\bf v}_h^t \big) \cdot \Big[ \boldsymbol{\rho}_{({\bf a}_{i-1}, \, h) } \circ \text{d} \boldsymbol{\rho}^\text{causal}_{({\bf a}_{i-1}, \, h) } \Big] - \Big[ \big( \boldsymbol{\Delta}^h_i {\bf v}_h^t \big) \circ  \boldsymbol{\rho}_{({\bf a}_{i-1}, \, h) } \Big] \cdot \big( \widetilde{ \boldsymbol{\rho}_{({\bf a}_{i-1}, \, h) } \ominus \text{d} \boldsymbol{\rho}^\text{causal}_{({\bf a}_{i-1}, \, h) } \big) } \nonumber \\
&\centerwithin{\downarrow}{\,\,=} \colorbox{light-blue}{using eqs.\eqref{eq:BroadcastFtDot}-\eqref{eq:potPourri} on the left-most term} \nonumber \\ 
&= \Big[ \big( \boldsymbol{\Delta}^h_i {\bf v}_h^t \big) \circ  \boldsymbol{\rho}_{({\bf a}_{i-1}, \, h) } \Big] \cdot\text{d} \boldsymbol{\rho}^\text{causal}_{({\bf a}_{i-1}, \, h) }  - \Big( \Big[ \widetilde{ \big( \boldsymbol{\Delta}^h_i {\bf v}_h^t \big) \ominus \boldsymbol{\rho}_{({\bf a}_{i-1}, \, h) } } \Big] \circ   \boldsymbol{\rho}_{({\bf a}_{i-1}, \, h) } \Big) \cdot \text{d} \boldsymbol{\rho}^\text{causal}_{({\bf a}_{i-1}, \, h)  }  \nonumber \\
&= \underbrace{ \mathcolorbox{shadecolor}{\Big[  \boldsymbol{\Delta}^h_i {\bf v}_h^t   -  \widetilde{ \big( \boldsymbol{\Delta}^h_i {\bf v}_h^t \big) \ominus \boldsymbol{\rho}_{({\bf a}_{i-1}, \, h) } }  \Big] \circ  \boldsymbol{\rho}_{({\bf a}_{i-1}, \, h) }}}_{\textstyle
    \begin{gathered}
      \boldsymbol{\Delta}^h_{\text{causal}} \sim \mathbb{R}^{n_\mathcal{T} \times n_\mathcal{T}}
    \end{gathered} } \cdot \, \text{d} \boldsymbol{\rho}^\text{causal}_{({\bf a}_{i-1}, \, h)  } \label{eq:DeltaCausal}
\end{align}
}

\noindent At this point, it is interesting to pause and take a deeper look at the structure of~$\boldsymbol{\Delta}^h_{\text{causal}}$.  We saw in the forward pass that the rows of~$\mathcolorbox{Gray}{\boldsymbol{\rho}_{({\bf a}_{i-1}, \, h) } }$ are normalized into a probability distribution.  Now that we have propagated the error flow back through the softmax function responsible for this normalization, we should expect~$\boldsymbol{\Delta}^h_{\text{causal}}$ to also reflect this constraint on the self-attention weights.  Generally, $\boldsymbol{\Delta}^h_{\text{causal}}$ can be written as a row stack of error flow vectors~$\sim \mathbb{R}^{n_\mathcal{T}}$ associated with the~$n_\mathcal{T}$ tokens in the sequence just as we did for the rows of~$\boldsymbol{\rho}_{({\bf a}_{i-1}, \, h) }$, see~eq.\eqref{eq:normalizedAttnScores}.  Therefore, we have
\begin{equation*}
\boldsymbol{\Delta}^h_{\text{causal}} = \left( \,
\begin{matrix}
\horzbar \,\, \boldsymbol{\delta}^h_\text{causal}(t=1) \sim \mathbb{R}^{n_\mathcal{T}} \,\, \horzbar \\
    \vdots \\
\horzbar \,\, \boldsymbol{\delta}^h_\text{causal}(t=n_\mathcal{T}) \sim \mathbb{R}^{n_\mathcal{T}} \,\, \horzbar
\end{matrix} \,
\right) = \left( \,
\begin{matrix}
\delta^\text{causal}_{11} & \cdots & \delta^\text{causal}_{1n_\mathcal{T}}  \\
\vdots  &  \ddots & \vdots \\
\delta^\text{causal}_{n_\mathcal{T}1} & \cdots & \delta^\text{causal}_{n_\mathcal{T} n_\mathcal{T}}
\end{matrix} \,
\right)_h \sim \mathbb{R}^{n_\mathcal{T}\times n_\mathcal{T}}
\end{equation*}

\noindent Now, in complete analogy with~eq.~\eqref{eq:normalization} where we considered one row~$\boldsymbol{\rho}_{({\bf a}_{i-1}, \, h) } (t={t^\star}) \sim \mathbb{R}^{n_\mathcal{T}}$ of the self-attention matrix and verified the normalization~$\sum \boldsymbol{\rho}_{({\bf a}_{i-1}, \, h) } (t={t^\star}) = 1$ of the pairwise attention weights~$\rho_{t^\star t^\prime}$ between token~$t^\star$ and all the other tokens~$t^\prime \in [ 1, \cdots , n_\mathcal{T} ]$, let us now consider one row~$\boldsymbol{\delta}_{\text{causal}} (t=t^\star) \sim \mathbb{R}^{n_\mathcal{T}}$ of ~$\boldsymbol{\Delta}^h_{\text{causal}}$ and evaluate the sum of its~$n_\mathcal{T}$ components
{\allowdisplaybreaks
\begin{align*}
\sum_{t^\prime = 1}^{n_\mathcal{T}}  \boldsymbol{\delta}_{\text{causal}} (t=t^\star) &= \sum_{t^\prime = 1}^{n_\mathcal{T}} \Big[ \left( \boldsymbol{\Delta}^h_i {\bf v}_h^t \right)_{t^\star}   - \widetilde{  \big( \boldsymbol{\Delta}^h_i {\bf v}_h^t \big)_{t^\star} \cdot \left( \boldsymbol{\rho}_{({\bf a}_{i-1}, \, h) } \right)_{t^\star} } \Big] \circ \left( \boldsymbol{\rho}_{({\bf a}_{i-1}, \, h) } \right)_{t^\star} \\
&\centerwithin{}{\,\,=} \colorbox{light-blue}{since we consider one row of $\boldsymbol{\Delta}^h_{\text{causal}}$, the operator~$\ominus$ reduces to a dot-product} \\
&\centerwithin{}{\,\,=} \colorbox{light-blue}{we use the notation $(\cdots)_{t^\star} \sim \mathbb{R}^{n_\mathcal{T}}$ as shorthand for $(\cdots)(t=t^\star)$, i.e. the~$t^\star$ row of~$(\cdots)$} \\
&= \sum_{t^\prime = 1}^{n_\mathcal{T}} \Big[ \left( \boldsymbol{\Delta}^h_i {\bf v}_h^t \right)_{t^\star} \circ \left( \boldsymbol{\rho}_{({\bf a}_{i-1}, \, h) } \right)_{t^\star} \big]  - \sum_{t^\prime = 1}^{n_\mathcal{T}} \Big[ \widetilde{ \big[ \big( \boldsymbol{\Delta}^h_i {\bf v}_h^t \big)_{t^\star} \cdot \left( \boldsymbol{\rho}_{({\bf a}_{i-1}, \, h) } \right)_{t^\star} \big] } \circ \left( \boldsymbol{\rho}_{({\bf a}_{i-1}, \, h) } \right)_{t^\star} \Big] \\
&\centerwithin{\downarrow}{\,\,=} \colorbox{light-blue}{because $\sum {\bf a} \circ {\bf b} = {\bf a} \cdot {\bf b} \sim \mathbb{R}$ and the broadcast is $\widetilde{{\bf a}\cdot {\bf b}} = ({\bf a} \cdot {\bf b}) [ 1 , \cdots , 1] \sim \mathbb{R}^{n_\mathcal{T}}  $} \\
&\centerwithin{}{\,\,=} \colorbox{light-blue}{and $({\bf a} \cdot {\bf b})$ can be taken out of the sum in the left-most expression} \\
&= \left( \boldsymbol{\Delta}^h_i {\bf v}_h^t \right)_{t^\star} \cdot \left( \boldsymbol{\rho}_{({\bf a}_{i-1}, \, h) } \right)_{t^\star}  - \big[ \big( \boldsymbol{\Delta}^h_i {\bf v}_h^t \big)_{t^\star} \cdot \left( \boldsymbol{\rho}_{({\bf a}_{i-1}, \, h) } \right)_{t^\star} \big] \sum_{t^\prime=1}^{n_\mathcal{T}} \boldsymbol{\rho}_{({\bf a}_{i-1}, \, h) } (t={t^\star}) \\
&\centerwithin{\downarrow}{\,\,=} \colorbox{light-blue}{because of the softmax normalization defined in eq.\eqref{eq:normalization} with $\sum_{t^\prime=1}^{n_\mathcal{T}} \boldsymbol{\rho}_{({\bf a}_{i-1}, \, h) } (t={t^\star}) = 1$} \\
&= \left( \boldsymbol{\Delta}^h_i {\bf v}_h^t \right)_{t^\star} \cdot \left( \boldsymbol{\rho}_{({\bf a}_{i-1}, \, h) } \right)_{t^\star}  -  \big( \boldsymbol{\Delta}^h_i {\bf v}_h^t \big)_{t^\star} \cdot \left( \boldsymbol{\rho}_{({\bf a}_{i-1}, \, h) } \right)_{t^\star} \\
&= 0
\end{align*}
}

\noindent This confirms that, just like the rows of~$\boldsymbol{\rho}_{({\bf a}_{i-1}, \, h) }$ which are constrained by~eq.\eqref{eq:normalization}, also the rows of~$\boldsymbol{\Delta}^h_{\text{causal}}$ reflect this conservation law on probability mass with another constraint on the gradients
\begin{equation}
\sum_{t^\prime = 1}^{n_\mathcal{T}}  \boldsymbol{\delta}^h_{\text{causal}} (t=t^\star) = \sum_{t^\prime = 1}^{n_\mathcal{T}}  \left( \delta^{\text{causal}}_{t^\star t^\prime} \right)_h = 0 \quad \text{; for each} \,\,t^\star \in \big[ 1, \cdots , n_\mathcal{T} \big]
\label{eq:gradConstraintMirror}
\end{equation}
This \hypertarget{gradientConstraint2}{observation} is a general property due to the softmax normalization and will manifest itself \hyperlink{gradientConstraint}{later} when we inspect the gradients with respect to the biases of the keys. \\

\noindent Now that we have finished this pause on analyzing the structure of~$\boldsymbol{\Delta}^h_{\text{causal}}$, we can move back to propagating the error signal one more step back through the causal mask prescribed by~eq.\eqref{eq:mask}.  Since this operation is parameter-free, there are no gradients to extract and
\begin{align}
\boldsymbol{\Delta}^h_\text{causal} \cdot \, \text{d} \boldsymbol{\rho}^\text{causal}_{({\bf a}_{i-1}, \, h)  } &=  \boldsymbol{\Delta}^h_\text{causal} \cdot \text{d} \left( {\bf m} \circ \boldsymbol{\rho}^\text{scaled}_{({\bf a}_{i-1}, \, h) } \right) \nonumber  \\ 
&= \boldsymbol{\Delta}^h_\text{causal} \cdot \left(  {\bf m} \circ \text{d} \boldsymbol{\rho}^\text{scaled}_{({\bf a}_{i-1}, \, h) } \right) \nonumber \\ 
&\centerwithin{\downarrow}{\,\,=} \colorbox{light-blue}{eq.(53) in the reference paper~\cite{deepPedestrians} } \nonumber \\
&= {\bf m} \circ \boldsymbol{\Delta}^h_\text{causal}\cdot \, \text{d} \boldsymbol{\rho}^\text{scaled}_{({\bf a}_{i-1}, \, h) } \nonumber \\
&\centerwithin{\downarrow}{\,\,=} \colorbox{light-blue}{because $\boldsymbol{\Delta}^h_\text{causal}$ is proportional to $\boldsymbol{\rho}_{({\bf a}_{i-1}, \, h) }$ which already contains the mask ${\bf m}$} \nonumber \\
&= \underbrace{ \mathcolorbox{shadecolor}{\boldsymbol{\Delta}^h_\text{causal}}}_{\textstyle
    \begin{gathered}
      \boldsymbol{\Delta}^h_{\text{scaled}}
    \end{gathered} } \cdot \, \text{d} \boldsymbol{\rho}^\text{scaled}_{({\bf a}_{i-1}, \, h) } \label{eq:DeltaCausalScaled}
\end{align}

\noindent Since~$\boldsymbol{\Delta}^h_\text{causal}$ is proportional to $\boldsymbol{\rho}_{({\bf a}_{i-1}, \, h) }$ which already contains the mask ${\bf m}$, the error flow stays unchanged with~$\boldsymbol{\Delta}^h_\text{scaled} = \boldsymbol{\Delta}^h_\text{causal}$. \\

\noindent The next step is another parameter-free scaling given by~eq.\eqref{eq:sqrtScaling}.  Applying the usual recursive backward error propagation, we have
\begin{align}
\boldsymbol{\Delta}^h_\text{scaled} \cdot \, \text{d} \boldsymbol{\rho}^\text{scaled}_{({\bf a}_{i-1}, \, h) } &= \boldsymbol{\Delta}^h_\text{scaled} \cdot \, \text{d} \left( \boldsymbol{\rho}^\text{raw}_{({\bf a}_{i-1}, \, h) } \, / \sqrt{d_\rho} \right) \nonumber \\
&= \underbrace{ \mathcolorbox{shadecolor}{\boldsymbol{\Delta}^h_\text{scaled} / \sqrt{d_\rho}}}_{\textstyle
    \begin{gathered}
      \boldsymbol{\Delta}^h_{\text{raw}}
    \end{gathered} } \cdot \, \text{d} \boldsymbol{\rho}^\text{raw}_{({\bf a}_{i-1}, \, h) } \label{eq:DeltaScaledRaw}
\end{align}

\noindent We have now reached the point where the raw self-attention weights~$\boldsymbol{\rho}^\text{raw}_{({\bf a}_{i-1}, \, h) } \sim \mathbb{R}^{n_\mathcal{T}\times n_\mathcal{T}}$ are determined by~eq.\eqref{eq:rawAttnDef} as the dot-product between the queries~${\bf q}_h \sim \mathbb{R}^{n_\mathcal{T}\times d_\rho}$ and the keys~${\bf k}_h \sim \mathbb{R}^{n_\mathcal{T}\times d_\rho}$ feature maps of the attention head~$h$.  Propagating~$\boldsymbol{\Delta}^h_{\text{raw}} \sim \mathbb{R}^{n_\mathcal{T}\times n_\mathcal{T}}$ back through this product, we have
\begin{align*}
\boldsymbol{\Delta}^h_\text{raw} \cdot \, \text{d} \boldsymbol{\rho}^\text{raw}_{({\bf a}_{i-1}, \, h) } &= \boldsymbol{\Delta}^h_\text{raw} \cdot \, \text{d} \left( {\bf q}_h \, {\bf k}_h^t \right) \\
&= \boldsymbol{\Delta}^h_\text{raw} \cdot \big[ \left( \text{d}  {\bf q}_h \right) \, {\bf k}_h^t +   {\bf q}_h \left( \text{d} {\bf k}_h^t \right) \big] \\
&\centerwithin{\downarrow}{\,\,=} \colorbox{light-blue}{eq.(52) in the reference paper~\cite{deepPedestrians} and eq.\eqref{eq:FrobeniusTranspose} of this paper} \\
&= \mathcolorbox{Gray}{\left( \boldsymbol{\Delta}^h_\text{raw} \, {\bf k}_h \right) \cdot \text{d} {\bf q}_h } + \mathcolorbox{Gray}{ \left( (\boldsymbol{\Delta}^h_\text{raw})^t \, {\bf q}_h  \right)  \cdot \text{d} {\bf k}_h }
\end{align*}
which \hypertarget{split2}{splits} into two different branches due to the queries/keys product.
\begin{itemize}
\item Let us first focus on the queries~${\bf q}_h$ and use their definition from eq.\eqref{queries:forward} to evaluate the first branch
\begin{align*}
\mathcolorbox{Gray}{\left( \boldsymbol{\Delta}^h_\text{raw} \, {\bf k}_h \right) \cdot \text{d} {\bf q}_h} &= \left( \boldsymbol{\Delta}^h_\text{raw} \, {\bf k}_h \right) \cdot \text{d} \left( {\bf a}_{i-1} {\bf w}_{q_h} + \widetilde{{\bf b}_{q_h}} \right) \\
&=  \underbrace{ \mathcolorbox{shadecolor}{{\bf a}_{i-1}^t \, \boldsymbol{\Delta}^h_\text{raw} \, {\bf k}_h} }_{\textstyle
    \begin{gathered}
      \frac{\partial \mathcal{L}_{\text{seq}}}{\partial {\bf w}_{q_h}}
    \end{gathered} } \, \cdot \, \text{d} {\bf w}_{q_h} + \underbrace{ \mathcolorbox{shadecolor}{ \sum_\text{tokens} \boldsymbol{\Delta}^h_\text{raw} \, {\bf k}_h } }_{\textstyle
    \begin{gathered}
      \frac{\partial \mathcal{L}_{\text{seq}}}{\partial {\bf b}_{q_h}}
    \end{gathered} } \cdot \, \text{d} {\bf b}_{q_h} + \underbrace{ \mathcolorbox{shadecolor}{\boldsymbol{\Delta}^h_\text{raw} \, {\bf k}_h {\bf w}_{q_h}^t  } }_{\textstyle
    \begin{gathered}
      \boldsymbol{\Delta}_{q_h}
    \end{gathered} } \cdot \, \text{d} {\bf a}_{i-1} 
\end{align*}
\noindent As a simple fully-connected layer, those expressions for the gradients and error signal propagation through the queries can be directly adapted from~Section~5 of the reference paper~\cite{deepPedestrians}.
\item Next, we look at the keys~${\bf k}_h$ and use their definition from eq.\eqref{keys:forward} to evaluate the second branch
\begin{align*}
\mathcolorbox{Gray}{\left( (\boldsymbol{\Delta}^h_\text{raw})^t \, {\bf q}_h  \right)  \cdot \text{d} {\bf k}_h} &=  \left( (\boldsymbol{\Delta}^h_\text{raw})^t \, {\bf q}_h  \right)  \cdot \text{d} \left( {\bf a}_{i-1} {\bf w}_{k_h} + \widetilde{{\bf b}_{k_h}} \right) \\
&= \underbrace{ \mathcolorbox{shadecolor}{ {\bf a}_{i-1}^t \, (\boldsymbol{\Delta}^h_\text{raw})^t \, {\bf q}_h }}_{\textstyle
    \begin{gathered}
      \frac{\partial \mathcal{L}_{\text{seq}}}{\partial {\bf w}_{k_h}}
    \end{gathered} } \, \cdot \, \text{d} {\bf w}_{k_h} + \underbrace{ \mathcolorbox{shadecolor}{ \sum_\text{tokens} (\boldsymbol{\Delta}^h_\text{raw})^t \, {\bf q}_h } }_{\textstyle
    \begin{gathered}
      \frac{\partial \mathcal{L}_{\text{seq}}}{\partial {\bf b}_{k_h}} = \mathbf{0}
    \end{gathered} } \cdot \, \text{d} {\bf b}_{k_h} + \underbrace{ \mathcolorbox{shadecolor}{ (\boldsymbol{\Delta}^h_\text{raw})^t \, {\bf q}_h \, {\bf w}_{k_h}^t } }_{\textstyle
    \begin{gathered}
      \boldsymbol{\Delta}_{k_h}
    \end{gathered} } \cdot \, \text{d} {\bf a}_{i-1} 
\end{align*}
\noindent Since this is another fully-connected layer, we get similar expressions as those we saw for the queries.  The only difference is that, now, the gradient of the loss with respect to the biases is identically null.  Let us see why this is the case.  We have seen previously that the softmax normalization imposes a mirror \hyperlink{gradientConstraint2}{constraint} on the rows of~$\boldsymbol{\Delta}^h_\text{raw} = \boldsymbol{\Delta}^h_\text{causal} / \sqrt{d_\rho} \sim \mathbb{R}^{n_\mathcal{T}\times n_\mathcal{T}}$, see~eq.\eqref{eq:gradConstraintMirror}.  Since we are interested in its transposed version~$(\boldsymbol{\Delta}^h_\text{raw})^t$, rows become columns and we represent this graphically as
\begin{equation*}
\boldsymbol{\Delta}^h_\text{raw} = \left( \,
\begin{matrix}
\horzbar \,\, \boldsymbol{\delta}^h_\text{raw}(t=1) \sim \mathbb{R}^{n_\mathcal{T}} \,\, \horzbar \\
    \vdots \\
\horzbar \,\, \boldsymbol{\delta}^h_\text{raw}(t=n_\mathcal{T}) \sim \mathbb{R}^{n_\mathcal{T}} \,\, \horzbar
\end{matrix} \,
\right) \quad \Longrightarrow \quad  (\boldsymbol{\Delta}^h_\text{raw})^t = \left( \,
\begin{matrix}
\vertbar &  & \vertbar \\[0.4em]
\boldsymbol{\delta}^h_\text{raw}(t=1) & \cdots &  \boldsymbol{\delta}^h_\text{raw}(t=n_\mathcal{T}) \\
\vertbar & & \vertbar \\
\end{matrix} \,
\right) 
\end{equation*}
where~$\boldsymbol{\delta}^h_\text{raw}(t=t^\star) = \big[ \delta^\text{raw}_{t^\star 1} \, , \cdots , \delta^\text{raw}_{t^\star n}  \big]_h $ for each~$t^\star \in \big[ 1, \cdots , n_\mathcal{T} \big]$.  Let us now evaluate the expression derived above for the gradient of the loss with respect to~${\bf b}_{k_h}$ as
\begin{align}
\frac{\partial \mathcal{L}_{\text{seq}}}{\partial {\bf b}_{k_h}} =  \sum_\text{tokens} (\boldsymbol{\Delta}^h_\text{raw})^t \, {\bf q}_h &= \dfrac{1}{\sqrt{d_\rho}}\sum_\text{tokens} \left( \begin{array}{c|c|c}
\delta_{11}^\text{causal} & \cdots & \delta_{n_\mathcal{T}1}^\text{causal} \\
\vertbar    & \cdots & \vertbar  \\[0.4em]
\delta_{1n_\mathcal{T}}^\text{causal} & \cdots &\delta_{n_\mathcal{T}n_\mathcal{T}}^\text{causal}
\end{array} \right)_h
{\bf q}_h \nonumber \\
&\centerwithin{\downarrow}{\,\,=} \colorbox{light-blue}{using eq.\eqref{eq:sumRows}} \nonumber  \\
&= \dfrac{1}{\sqrt{d_\rho}}  \left( \sum_{t^\prime=1}^{n_\mathcal{T}}( \delta_{1 t^\prime}^\text{causal})_h \, , \cdots , \sum_{t^\prime=1}^{n_\mathcal{T}} (\delta_{n_\mathcal{T} t^\prime}^\text{causal})_h \right) 
{\bf q}_h \nonumber \\
&\centerwithin{\downarrow}{\,\,=} \colorbox{light-blue}{because of the constraint $\sum_{t^\prime = 1}^{n_\mathcal{T}} ( \delta^{\text{causal}}_{t^\star t^\prime})_h = 0$} \nonumber  \\
&\centerwithin{}{\,\,=} \colorbox{light-blue}{due to softmax normalization, see eq.\eqref{eq:gradConstraintMirror}} \nonumber  \\
\frac{\partial \mathcal{L}_{\text{seq}}}{\partial {\bf b}_{k_h}} &= \mathbf{0} \sim \mathbb{R}^{d_\rho}
\label{eq:biasZeroGrad}
\end{align}

\noindent This shows that the gradients with respect to the biases of the keys are identically \hypertarget{gradBiasesNull2}{null}.  This is a consequence of our \hyperlink{gradientConstraint2}{previous observation}  on the \hypertarget{gradientConstraint}{gradient constraint}.  It is also \hyperlink{gradBiasesNull}{consistent} with our discussion about the independence of the self-attention weights~$\boldsymbol{\rho}_{({\bf a}_{i-1}, \, h) }$ on the biases~${\bf b}_{k_h}$ of the keys during the forward pass.
\end{itemize}

\stepcounter{equation}
\noindent In summary, we have
\begin{empheq}[box={\backPropBox[{\bf Queries and Keys}: backward pass]}]{alignat=10}
\boldsymbol{\Delta}_{q_h} &= \boldsymbol{\Delta}^h_\text{raw} \, {\bf k}_h {\bf w}_{q_h}^t &\quad ; \quad && \boldsymbol{\Delta}_{k_h} &{}={}&& (\boldsymbol{\Delta}^h_\text{raw})^t \, {\bf q}_h \, {\bf w}_{k_h}^t  &\quad  & \sim \mathbb{R}^{n_\mathcal{T} \times d}  \tag{\theequation.a-b} \label{eq:DeltaKQ} \\ \stepcounter{equation}
\frac{\partial \mathcal{L}_{\text{seq}}}{\partial {\bf w}_{q_h}}  &=  {\bf a}_{i-1}^t \, \boldsymbol{\Delta}^h_\text{raw} \, {\bf k}_h &\quad ; \quad && \frac{\partial \mathcal{L}_{\text{seq}}}{\partial {\bf w}_{k_h}} &{}={}&& ( \boldsymbol{\Delta}^h_\text{raw} \, {\bf a}_{i-1})^t \, {\bf q}_h  &\quad  & \sim \mathbb{R}^{d \times d_\rho} \tag{\theequation.a-b} \label{grad:QVw} \\ \stepcounter{equation}
\frac{\partial \mathcal{L}_{\text{seq}}}{\partial {\bf b}_{q_h}}  &= \sum_\text{tokens} \boldsymbol{\Delta}^h_\text{raw} \, {\bf k}_h &\quad ; \quad && \frac{\partial \mathcal{L}_{\text{seq}}}{\partial {\bf b}_{k_h}} &{}={}&& \mathbf{0}  &\quad  & \sim \mathbb{R}^{d_\rho} \tag{\theequation.a-b} \label{grad:QVb}
\end{empheq}

\noindent At this point, we have extracted the gradients of the loss function for a single sequence of~$n_\mathcal{T}$ tokens with respect to all of the parameters of the self-attention head~$h$ that transformed the~$d$-dimensional token feature maps of the input sequence~${\bf a}_{i-1} \sim \mathbb{R}^{n_\mathcal{T}\times d}$ into new~$d_h$-dimensional feature maps of the output sequence~${\bf a}_i \sim \mathbb{R}^{n_\mathcal{T}\times d_h}$.  Those gradients comprise of three fully-connected layers associated with the values~$\partial \mathcal{L}_\text{seq} / \partial \{ {\bf w}_{v_h}, {\bf b}_{v_h} \}$, see~eqs.\eqref{grad:Vw}-\eqref{grad:Bw}, queries~$\partial \mathcal{L}_\text{seq} / \partial \{ {\bf w}_{q_h}, {\bf b}_{q_h} \}$ and keys~$\partial \mathcal{L}_\text{seq} / \partial \{ {\bf w}_{k_h}, {\bf b}_{k_h} \}$, see~eqs.\eqref{grad:QVw}-\eqref{grad:QVb}. \\

\noindent Along the way, we have also propagated the upstream error signal~$\boldsymbol{\Delta}^h_i \sim \mathbb{R}^{n_\mathcal{T}\times d_h}$ allocated to attention head~$h$ from the~${\bf a}_i^h \sim \mathbb{R}^{n_\mathcal{T}\times d_h}$ output sequence level downstream back to the level of the input sequence~${\bf a}_{i-1} \sim \mathbb{R}^{n_\mathcal{T}\times d}$. As per usual terminology, let us denote by~$\boldsymbol{\Delta}^h_{i-1}\sim \mathbb{R}^{n_\mathcal{T}\times d}$ this downstream error signal.  Since there are three trainable layers (values, queries and keys) in the attention head,~$\boldsymbol{\Delta}^h_{i-1}$ should also be made up of three contributions.  \\

\noindent We saw at the very first stage of backpropagation how the upstream error signal~$\boldsymbol{\Delta}^h_i$ \hyperlink{split1}{splits} into two different branches.  The branch associated with the values feature maps~${\bf v}_h$ is directly connected to the input sequence~${\bf a}_{i-1}$ and therefore its error term~$\boldsymbol{\Delta}_{v_h}$, see~eq.\eqref{grad:deltaVh}, is the first contributor to the downstream signal~$\boldsymbol{\Delta}_{i-1}^h$. The other branch goes through a series of steps (all internal attributes of the attention head~$h$ not exposed elsewhere) following~$\boldsymbol{\Delta}^h_{i-1} \rightarrow \boldsymbol{\Delta}^h_\text{causal} \rightarrow \boldsymbol{\Delta}^h_\text{scaled} \rightarrow \boldsymbol{\Delta}^h_\text{raw}$, see eqs.\eqref{eq:DeltaCausal},~\eqref{eq:DeltaCausalScaled},~\eqref{eq:DeltaScaledRaw} before reaching another \hyperlink{split2}{split} due to the query/key product.  Since those feature maps~${\bf q}_h$ and~${\bf k}_h$ are themselves directly connected to the input sequence~${\bf a}_{i-1}$, the process ends here with their respective two contributions~$\boldsymbol{\Delta}_{q_h}$ and~$\boldsymbol{\Delta}_{k_h}$, see~eqs.\eqref{eq:DeltaKQ}, adding into the downstream error signal~$\boldsymbol{\Delta}^h_{i-1}$. \\

\noindent In summary, the downstream error signal for attention head~$h$ is given by
\begin{empheq}[box={\backPropBox[{\bf Error signal}: backward pass]}]{alignat=2}
\boldsymbol{\Delta}^h_{i-1} = \boldsymbol{\Delta}_{v_h} +\boldsymbol{\Delta}_{q_h} + \boldsymbol{\Delta}_{k_h} &\quad &\sim \mathbb{R}^{n_\mathcal{T} \times d} 
\label{eq:singleHeadCompleteSignal}
\end{empheq}

\noindent Notice how (as it should be) the dimensionality of the downstream error signal matches that of the input sequence~$\boldsymbol{\Delta}^h_{i-1} \sim {\bf a}_{i-1} \sim \mathbb{R}^{n_\mathcal{T}\times d}$ even though it is produced by a single attention head~$h$.  We will see in~Section~\ref{sec:multiHead} how multiple error signals coming from different attention heads are combined together into a complete~$\boldsymbol{\Delta}_{i-1} \sim \mathbb{R}^{n_\mathcal{T}\times d}$ of the same dimensionality. \\

\tcbset{colframe=black, colback=lightgray, boxrule=1pt, arc=0mm, breakable}
\begin{tcolorbox}

{\bf Some other noteworthy properties of self-attention}

\paragraph{Permutation equivariance in non-causal attention.} \hypertarget{permutationAtt}{Here}, we focus on non-causal attention where the mask~${\bf m}$ in~eq.\eqref{selfAttention:forward} is removed (restriction-free attention range with a matrix of ones~${\bf m} = {\bf J}_{n_\mathcal{T}}$).  Let us consider a permutation matrix~${\bf P}_\pi \sim \mathbb{R}^{n_\mathcal{T}\times n_\mathcal{T}}$ and apply it to the tokens (i.e. the rows) of the input sequence (i.e. from the left~\cite{wikiPerm}).  Passing this token-order permutated~$\pi(n_\mathcal{T})$  input sequence~${\bf P}_{\pi} \, {\bf a}_{i-1} \sim \mathbb{R}^{\pi(n_\mathcal{T})\times d}$ through a non-causal self-attention head~$h$, we get
\begin{flalign*}
\text{Att}_{\mathcal{P}_h} \big( {\bf P}_\pi \, {\bf a}_{i-1} \big) &= \text{softmax} \big[ {\bf P}_{\pi}  {\bf q}_h \left( {\bf P}_{\pi} {\bf k}_h \right)^t / \sqrt{d_\rho} \, \big] \, {\bf P}_{\pi}  {\bf v}_h && \\
&= \text{softmax} \big[ {\bf P}_{\pi} \big( {\bf q}_h {\bf k}_h^t / \sqrt{d_\rho} \big) {\bf P}_{\pi}^t \big] \, {\bf P}_{\pi}  {\bf v}_h \\
&\centerwithin{\downarrow}{\,\,=} \colorbox{Gray}{see the commutative properties  in eqs.\eqref{softmaxPermRows}-\eqref{softmaxPermCols}} \\
&= {\bf P}_{\pi} \,  \text{softmax} \big(  {\bf q}_h {\bf k}_h^t / \sqrt{d_\rho} \big)  {\bf P}_{\pi}^t  \, {\bf P}_{\pi}  {\bf v}_h \\
&\centerwithin{\downarrow}{\,\,=} \colorbox{Gray}{since~${\bf P}_{\pi}^t = {\bf P}_{\pi}^{-1}$ for permutation matrices~\cite{wikiPerm}.} \\
&= {\bf P}_{\pi}\,  \text{softmax} \big(  {\bf q}_h {\bf k}_h^t / \sqrt{d_\rho} \big) \, {\bf v}_h
\end{flalign*}

\noindent leading to $\mathcolorbox{Gray}{\text{Att}_{\mathcal{P}_h} \big( {\bf P}_{\pi} \, {\bf a}_{i-1} \big) = {\bf P}_{\pi} \,  \text{Att}_{\mathcal{P}_h} \big( {\bf a}_{i-1} \big)}$ demonstrating that non-causal self-attention is equivariant under permutation: Any permutation in the order of the input token feature maps is straightforwardly inherited by the output feature maps which end up permutated in exactly the same manner as the input was.  This symmetry ensures that the output feature map~${\bf a}_i^h(t=t^\star) \sim \mathbb{R}^{n_\mathcal{T}\times d_h}$ of a token~$t^\star \in [1, \cdots , t_{n_\mathcal{T}}]$ stays with the same feature values regardless of its position in the sequence.  Permutation equivariance can be understood as a structural constraint enforcing that each token preserves a distinct ``identity'' (where this ``identity'' is determined by a token's pairwise attention relationships with every other~$n_\mathcal{T}$ tokens in context) which is order-agnostic.  For some tasks (such as language) where order is crucial, positional encoding is added into the architecture along with shortcut connections to enforce token-order sensitivity. \\

\noindent Note that this equivariance property should be distinguished from permutation invariance which would require the output representation~${\bf a}_i^h$ to always be the same for all possible permutations~${\bf P}_{\pi} \, {\bf a}_{i-1}$ of the input effectively reducing~${\bf a}_i^h$ to a global ``pooling'' summary that eliminates individual token identities~\cite{deepSets}.  Finally, we also point out that permutation equivariance does not hold for causal attention heads where the mask implicitly injects positional information.  This raises the question of whether explicit positional encodings remain necessary in architectures employing causal masking~\cite{dissection, causalMaskPE}.

\paragraph{KV cache in autoregressive next-token generation.} \hypertarget{kvcacheNote}{Let} us consider a trained model designed for causal next-token sampling:  Given an input sequence~${\bf a}_{i-1} \sim \mathbb{N}^{n_\mathcal{T}}$ consisting of~$n_\mathcal{T}$ tokens, the forward pass returns a probability distribution over the vocabulary and picks a new token~$t_\text{next} \sim \mathbb{N}$.  This token is appended to the original input tokens to form a new sequence~$[ {\bf a}_{i-1} \oplus t_\text{next} ] \sim \mathbb{N}^{n_\mathcal{T}+1} $ now composed of~$n_\mathcal{T}+1$ tokens.  This updated sequence can then be fed back as a new input to the model to generate yet one more token.  Repeating this process multiple times allows the iterative generation of one token per forward pass resulting in an ever increasing length of input sequence (as each newly predicted token is appended back into a new input sequence for the model).   \\

\noindent As the \hyperlink{selfAttQuad}{computational complexity} of a self-attention head grows quadratically with the number of tokens, this process becomes computationally prohibitive for long sequences.  However, due to the autoregressive nature of next-token prediction, self-attention involves a lot of redundant computations across generation steps which can be eliminated via KV (Key~Value)~caching for optimized inference efficiency. \\

\noindent To see how KV caching works, let us assume that we have already evaluated the self-attention head for a sequence~${\bf a}_{i-1}$ of~$n_\mathcal{T}$ tokens and kept in memory the tokens' keys feature maps~${\bf k}_h^\text{cache} \leftarrow {\bf k}_h \sim \mathbb{R}^{n_\mathcal{T}\times d_\rho}$ as well as their values feature maps~${\bf v}_h^\text{cache} \leftarrow {\bf v}_h \sim \mathbb{R}^{n_\mathcal{T}\times d_h}$.  When a new token~$t_\text{next}$ is appended~$[{\bf a}_{i-1} \oplus t_\text{next}]$, we only need to evaluate three new feature maps~${\bf q}_h(t=t_\text{next}) \sim {\bf k}_h(t=t_\text{next}) \sim \mathbb{R}^{d_\rho}$ and ${\bf v}_h(t=t_\text{next}) \sim \mathbb{R}^{d_h}$ associated with this token only. Indeed, there is no need to recalculate the other feature maps for any of the other tokens since those features are functions of the individual tokens which are processed independently by fully connected layers (and therefore would end up with again the same features regardless of how many and/or which other tokens are present in the input sequence).  Moreover, because of causality in autoregressive generation, \colorbox{Gray}{we only need to compute the} \colorbox{Gray}{last row~$\sim \mathbb{R}^{n_\mathcal{T}+1}$ of the new attention weight matrix}~$\boldsymbol{\rho}_{([{\bf a}_{i-1} \oplus t_\text{next} ], \, h) } \sim \mathbb{R}^{(n_\mathcal{T}+1)\times (n_\mathcal{T}+1)} $.  The causal mask ensures that the last column (right-most) is always null except for the last row (bottom-most associated with the new token~$t_\text{next}$) which is the only one with a complete row of non-zero values, see~eq.\eqref{eq:mask}.  This means the presence of this new token~$t_\text{next}$ adds a new row to~$\boldsymbol{\rho}_{([{\bf a}_{i-1} \oplus t_\text{next} ], \, h) }$ but leaves all the other components completely unchanged. \\

\noindent To evaluate this last row of attention weights for~$t_\text{next}$, we carry out a vector-matrix product between the new query feature map~${\bf q}_h(t=t_\text{next})$ and the (transposed) keys feature maps
\begin{equation*}
\boldsymbol{\rho}_{([{\bf a}_{i-1} \oplus t_\text{next} ], \, h) }(t=t_\text{next}) = \text{softmax} \left( {\bf q}_h(t=t_\text{next}) \Big[ {\bf k}_h^\text{cache}  \oplus  {\bf k}_h (t=t_\text{next})  \Big]^t \right) \sim \mathbb{R}^{n_\mathcal{T}+1} 
\end{equation*}
where the new key feature map vector has been appended to the cached feature maps of the other tokens~$[ {\bf k}_h^\text{cache}  \oplus  {\bf k}_h (t=t_\text{next}) ] \sim \mathbb{R}^{(n_\mathcal{T}+1)\times d_\rho} $ (see above~eq.\eqref{eq:rawAttnDef} and consider only the last row of queries for a vector-matrix product). \\

\noindent The ouptut representation for~$t_\text{next}$ is then obtained as a linear combination of the cached and new values feature maps~$[ {\bf v}_h^\text{cache}  \oplus  {\bf v}_h (t=t_\text{next}) ] \sim \mathbb{R}^{(n_\mathcal{T}+1)\times d_h} $ weighted by the attention vector computed above using another vector-matrix product
\begin{equation*}
{\bf a}_i^h (t=t_\text{next}) = \boldsymbol{\rho}_{([{\bf a}_{i-1} \oplus t_\text{next} ], \, h) }(t=t_\text{next})   \big[ {\bf v}_h^\text{cache} \oplus  {\bf v}_h (t=t_\text{next})  \big] \sim \mathbb{R}^{(n_\mathcal{T}+1)\times d_h}
\end{equation*}

\noindent In most use-cases,~${\bf a}_i^h (t=t_\text{next})$ is, by itself, (after downstream processing by more fully-connected layers), enough to predict the probability distribution of token coming after~$t_\text{next}$ and the KV caching process is henceforth repeated iteratively to keep on generating more tokens. \\

\noindent Crucially, by expressing autoregressive next-token generation as a process that is limited to evaluating vector-matrix expressions of complexity~$\mathcal{O}(n_\mathcal{T})$, \colorbox{Gray}{KV~cache makes inference} \colorbox{Gray}{grow linearly with respect to sequence length instead of quadratically} as it would be with a na\"ive recalculation of all attention weights via matrix-matrix products of complexity~$\mathcal{O}(n_\mathcal{T}^2)$ (ignoring the complexity scaling on the dimensionality of the feature maps).  Obviously, this accelerated inference comes at the cost of significant memory requirements to cache the keys~${\bf k}_h^\text{cache} \sim \mathbb{R}^{n_\mathcal{T}\times d_\rho}$ and values feature maps~${\bf v}_h^\text{cache}$ whose size is unbounded as they grow linearly with the (potentially unknown in advance) number of tokens thereby creating \colorbox{Gray}{memory} \colorbox{Gray}{management complications} and restricted context windows~\cite{KVcacheMem}.  \\

{\it (Finally, as long as we define the rows of the attention matrix to represent the attention weights of the each token, we can \hyperlink{QKreverse}{swap} the notation from queries to keys and, in this case, we would instead talk about a~QV~cache...)}

\end{tcolorbox}

\section{Multi-headed attention layer}
\label{sec:multiHead}

\begin{table}
\hspace*{-2.66cm}
\captionsetup{singlelinecheck=off}
\begin{tabular}{|| l | y | x | z ||}
\hline
\multicolumn{4}{|c|}{\rule{0pt}{1.05\normalbaselineskip} {\bf Multi-headed self-attention}} \\[0.2em] \hline \hline
{\bf Layer} & \cellcolor{orange}{\bf Forward pass} & {\bf Shape} & \cellcolor{light-blue}{\bf Backward pass} \\
\hline
\rule{0pt}{1.1\normalbaselineskip}
Input data & ${\bf a}_{i-1} $ & $\mathbb{R}^{n_\mathcal{T}\times d}$ & $\boldsymbol{\Delta}_{i-1} = \sum_{h=1}^{n_h} \boldsymbol{\Delta}_{i-1}^h$ \\[0.3em] \hline
 \multicolumn{4}{|c|}{\rule{0pt}{0.5\normalbaselineskip} } \\ \hline
 \rule{0pt}{1.1\normalbaselineskip}
Head \#1 & ${\bf a}_i^{(h=1)} = \text{Att}_{(h=1)}({\bf a}_{i-1}) $ & $\mathbb{R}^{n_\mathcal{T} \times d_h}$ & $\boldsymbol{\Delta}_i^{(h=1)} \rightarrow \boldsymbol{\Delta}_{i-1}^{(h=1)} \sim \mathbb{R}^{n_\mathcal{T}\times d}$  \\[0.3em] \hline
\multicolumn{2}{|c|}{\rule{0pt}{1.05\normalbaselineskip} \cellcolor{Gray} $\vdots$ } & \multicolumn{1}{|c|}{\rule{0pt}{1.05\normalbaselineskip} \cellcolor{Gray} {$\vdots$}} &  \multicolumn{1}{|c|}{\rule{0pt}{1.05\normalbaselineskip} \cellcolor{Gray} $\vdots$   } \\[0.6em] \hline
\rule{0pt}{1.1\normalbaselineskip}
Head \#$n_h$ & ${\bf a}_i^{(h=n_h)} = \text{Att}_{(h=n_h)}({\bf a}_{i-1}) $ & $\mathbb{R}^{n_\mathcal{T} \times d_h}$ & $\boldsymbol{\Delta}_i^{(h=n_h)} \rightarrow \boldsymbol{\Delta}_{i-1}^{(h=n_h)} \sim \mathbb{R}^{n_\mathcal{T}\times d} $ \\[0.2em] \hline
\multicolumn{4}{|c|}{\rule{0pt}{0.5\normalbaselineskip} } \\ \hline
\multicolumn{2}{|c|}{\rule{0pt}{1.05\normalbaselineskip}  {\it $\downarrow$ }} & \multicolumn{1}{|c|}{\rule{0pt}{1.05\normalbaselineskip} {$\downarrow$\quad\quad$\uparrow$}} &  \multicolumn{1}{|c|}{\rule{0pt}{1.05\normalbaselineskip} \cellcolor{Gray}  {\it each head~$h$ is allocated its own~$\boldsymbol{\Delta}_i^h \sim \mathbb{R}^{n_\mathcal{T}\times d_h}$ sliced out of~$\boldsymbol{\Delta}_i$ }   } \\[0.3em] \hline
\rule{0pt}{1.2\normalbaselineskip}
Output data  & $ {\bf a}_i = \text{concat} \left[ {\bf a}_i^{(h=1)} , \cdots  \cdots, {\bf a}_i^{(h=n_h)}  \right]  $ & $\mathbb{R}^{n_\mathcal{T} \times d}$ & $\boldsymbol{\Delta}_i  $ \\[5pt] 
\hline
\end{tabular}
\end{table}

\noindent In Section~\ref{sec:singleHead}, we saw how a self-attention head~$h$ can be seen as a parametrized function~$\text{Att}_{\mathcal{P}_h}$ that transforms the~$d$-dimensional input feature maps~${\bf a}_{i-1} \sim \mathcal{R}^{n_\mathcal{T}\times d}$ of the~$n_\mathcal{T}$ tokens into new~$d_h$-dimensional representations~${\bf a}_i^h \sim \mathbb{R}^{n_\mathcal{T}\times d_h}$ (by treating the sequence itself as a collective unit).  Each attention head~$h$ is associated with its own set of parameters~$\mathcal{P}_h$. \\

\noindent In this Section, we focus on multi-headed attention in which, instead of a single attention head, we consider a layer composed of multiple independent attention heads.  In a manner somewhat similar to the different filters in the convolution layers of CNNs~\cite{deepPedestrians}, it may be argued that having multiple parallel heads in attention layers allows the network to learn different aspects of the data simultaneously; see~\cite{16heads} and citations for more discussion.  As far as practitioners are concerned, multi-headed attention has become an overwhelming standard.

\paragraph{Forward pass.}  Let us denote by~$n_h$ the number of attention heads in a multi-headed layer of self-attention which takes~${\bf a}_{i-1} \sim \mathbb{R}^{n_\mathcal{T}\times d}$ as its input.  In order to have an integer number of heads, we choose~$n_h = d / d_h \in \mathbb{N}$ enforcing that the dimensionality~$d_h$ of the tokens' feature vectors produced by the attention heads be an exact divisor of their input dimensionality~$d$.  \\

\noindent Since all attention heads are parametrized by their own set of parameters~$\mathcal{P}_1 \neq \cdots \neq \mathcal{P}_{n_h}$, each head operates independently of all the other ones and a multi-headed attention layer is defined as a list of functions
\begin{equation*}
\text{MultiAtt} = \left[ \text{Att}_{\mathcal{P}_1} , \cdots , \text{Att}_{\mathcal{P}_{n_h}}  \right]
\end{equation*}
Applying these functions to the same input sequence~${\bf a}_{i-1}$ produces~$n_h$ output representations
\begin{equation*}
\text{MultiAtt} ({\bf a}_{i-1})
 = \left[ \text{Att}_{\mathcal{P}_1} ({\bf a}_{i-1}), \cdots , \text{Att}_{\mathcal{P}_{n_h}} ({\bf a}_{i-1}) \right] 
 = \left[ {\bf a}_i^{(h=1)} \sim \mathbb{R}^{n_\mathcal{T}\times d_h} , \cdots , {\bf a}_i^{(h=n_h)} \sim \mathbb{R}^{n_\mathcal{T}\times d_h} \right]
\end{equation*}
where each head~$h$ contributes its own~${\bf a}_i^h = \text{Att}_{\mathcal{P}_h}({\bf a}_{i-1}) \sim \mathbb{R}^{n_\mathcal{T}\times d_h}$ like in~eq.\eqref{eq:attHfunction}.  Finally, one concatenates together the feature maps produced by all the heads (i.e. column-wise concatenation) to yield the consolidated output~${\bf a}_i \sim \mathbb{R}^{n_\mathcal{T}\times d}$ where we recover the expected~$d = n_h \times d_h$.  In summary
\begin{empheq}[box={\forwardBox[{\bf Multi-headed self-attention}: forward pass]}]{alignat=2}
{\bf a}_{i} &=  \text{concat} \left[ {\bf a}_i^{(h=1)}  , \cdots , {\bf a}_i^{(h=n_h)}  \right]
\label{multiHead:forward}
\end{empheq}

\noindent Because of the choice~$n_h = d/d_h \in  \mathbb{N}$, after passing through a multi-headed attention layer each token is represented by a new $d$-dimensional vector which is made up of~$n_h$ different~$d_h$-dimensional feature maps produced by all the self-attention heads.  This constraint on the values of the~($n_h, d_h, d$) tuple ensures that the dimensionality of the output tokens' feature maps is the same as that of the input feature maps, i.e.~${\bf a}_{i-1} \sim {\bf a}_i \sim \mathbb{R}^{n_\mathcal{T}\times d}$.  This way, one may easily compose multiple multi-headed attention layers together.  One benefit of stacking multiple attention layers is that, even though an individual self-attention head involves only pairwise token-to-token interactions, composing multiple such layers effectively introduces higher-level interactions which eventually span the entire sequence of length~$n_\mathcal{T}$.  This point was discussed in \hyperlink{compositionLayers}{detail} for a single head of self-attention and remains equally valid for multi-headed attention.

\paragraph{Backward pass.}  The first step consists in reversing the concatenation operation carried out in the last step of the forward pass by slicing out the upstream error signal~$\boldsymbol{\Delta}_i \sim \mathbb{R}^{n_\mathcal{T}\times d}$ column-wise into~$n_h$ sub-components
\begin{equation*}
\boldsymbol{\Delta}_i \sim \mathbb{R}^{n_\mathcal{T}\times d} \longrightarrow \left[ \boldsymbol{\Delta}_i^{(h=1)} \sim \mathbb{R}^{n_\mathcal{T}\times d_h} \, , \cdots , \,\boldsymbol{\Delta}_i^{(h=n_h)} \sim \mathbb{R}^{n_\mathcal{T}\times d_h}  \right]
\end{equation*}
where each~$\boldsymbol{\Delta}_i^h \sim \mathbb{R}^{n_\mathcal{T}\times d_h}$ is allocated to a specific head~$h$.  At this point, we can use simply~eq.\eqref{eq:singleHeadCompleteSignal} to propagate this error signal back through~$h$ to get the downstream error signal~$\boldsymbol{\Delta}_{i-1}^h \sim \mathbb{R}^{n_\mathcal{T}\times d}$ which, as required, recovers the dimensionality of the input sequence.  Doing the same to all~$n_h$ attention heads, leads the complete downstream error signal through a multi-headed attention layer as
\begin{empheq}[box={\backPropBox[{\bf Multi-headed self-attention}: backward pass]}]{alignat=2}
\boldsymbol{\Delta}_{i-1} = \sum_{h=1}^{n_h} \boldsymbol{\Delta}_{i-1}^h &\quad &\sim \mathbb{R}^{n_\mathcal{T} \times d} 
\label{eq:multiHeadCompleteSignal}
\end{empheq}

\section{Layer normalization}
\label{sec:ln}

\noindent Typically, neural network architectures designed for datasets with an inherent sequential nature favor layer normalization~\cite{layerNorm}~--~\texttt{LN} over batch normalization~\cite{bn}~--~\texttt{BN} for the purpose of training stabilization.  While the original motivation for layer normalization came from its observed empirical superiority in recurrent architectures, it remains preferred even in transformer-based models.  As layer normalization treats all tokens (referred to as samples in~\texttt{BN}) independently, it is able to gracefully handle variable-length sequences without being affected by cross-token/sample statistics.

\paragraph{Forward pass.} As a reminder, the input data~${\bf a}_{i-1} \sim \mathbb{R}^{n_\mathcal{T}\times d}$ represents the~$d$-dimensional feature vectors associated with each one of the~$n_\mathcal{T}$ tokens in a sequence.  We separate the forward pass into two distinct steps:
\begin{enumerate}
\item \uwave{Normalization.}  Considering a specific token~$t^\star$, the statistical distribution of its feature vector~${\bf a}_{i-1}(t=t^\star) = [ a_{i-1}^1(t^\star) , \cdots , a_{i-1}^d(t^\star)] \sim \mathbb{R}^{d}$ can be summarised by its first two moments
\begin{equation*}
\mu_{t^\star}  =  \dfrac{1}{d} \sum_{f=1}^d a_{i-1}^f(t=t^\star) \sim \mathbb{R} \quad \text{and} \quad \sigma_{t^\star} = \sqrt{\dfrac{1}{d} \sum_{f=1}^d \left( a_{i-1}^f(t=t^\star) - \mu_{t^\star}  \right)^2 }   \sim \mathbb{R}
\end{equation*}
Once the mean~$\mu_{t^\star}$ and standard deviation~$\sigma_{t^\star}$ have been evaluated, those summary statistics are used to produce a normalized feature vector~$\overline{\bf a}_{i-1}(t=t^\star)$ which is specific to this token via
\begin{equation*}
\overline{\bf a}_{i-1}(t=t^\star) = \dfrac{{\bf a}_{i-1}(t=t^\star) - \mu_{t^\star}}{\sigma_{t^\star}} \sim \mathbb{R}^{d}
\end{equation*}
where~$\mu_{t^\star}$ and~$\sigma_{t^\star}$ are both broadcast vector-wise such that~$\overline{\bf a}_{i-1}(t=t^\star)$ is well-defined and normalized with its own token-specific values.  Obviously, the same feature-wise normalization may be applied independently to all tokens yielding vectors~$(\boldsymbol{\mu}, \boldsymbol{\sigma})_\texttt{LN} \sim \mathbb{R}^{n_\mathcal{T}}$ of mean values and standard deviations which are used to normalize the feature vectors of each token from~${\bf a}_{i-1}$ to
\begin{equation}
\tcboxmath[colframe=black, colback=gray!20, drop lifted shadow, boxsep=0pt, left=6pt, right=6pt]{
\overline{\bf a}_{i-1} = \text{diag}\left( 1 / \boldsymbol{ \sigma}  \right) \left( {\bf a}_{i-1} - \widetilde{\boldsymbol{ \mu}} \right) \sim \mathbb{R}^{n_\mathcal{T}\times d}
} 
\label{eq:lnNorm}
\end{equation}
where the vector of mean values is column-wise broadcast~$\boldsymbol{ \mu} \sim \mathbb{R}^{n_\mathcal{T}} \rightarrow  \widetilde{\boldsymbol{ \mu}} \sim \mathbb{R}^{n_\mathcal{T}\times d}$ and the vector of standard deviations is lifted into a diagonal representation~$\text{diag} ( 1/ \boldsymbol{ \sigma} ) \sim \mathbb{R}^{n_\mathcal{T}\times n_\mathcal{T}}$ to reproduce the proper token normalization shown above. 

\item \uwave{Learnable affine transformation.}  Next we apply an affine transformation by introducing two vectors~$\{ {\bf w}_{i-1} \sim \mathbb{R}^d \, , {\bf b}_{i-1} \sim \mathbb{R}^d \}$.  Taking token~$t^\star$ as an example, we wish for its normalized feature vector~$\overline{\bf a}_{i-1}(t=t^\star)$ to be transformed into
\begin{equation*}
{\bf a}_i(t=t^\star) = \overline{\bf a}_{i-1}(t=t^\star) \circ {\bf w}_{i-1} + {\bf b}_{i-1} \sim \mathbb{R}^{d}
\end{equation*}
where the components of the weights~${\bf w}_{i-1}$ and biases~${\bf b}_{i-1}$ are learned during training.  Applying the same transformation to all tokens may be achieved by
\begin{equation}
\tcboxmath[colframe=black, colback=gray!20, drop lifted shadow, boxsep=0pt, left=6pt, right=6pt]{
{\bf a}_i = \overline{\bf a}_{i-1} \, \text{diag}\left( {\bf w}_{i-1} \right) + \widetilde{\bf b}_{i-1}
} 
\label{eq:affineLN}
\end{equation}
where the bias vector is broadcast row-wise~${\bf b}_{i-1} \sim \mathbb{R}^d \rightarrow  \widetilde{\bf b}_{i-1} \sim \mathbb{R}^{n_\mathcal{T} \times d}$. Lifting the components of~${\bf w}_{i-1} \sim \mathbb{R}^d$ into a diagonal~$\text{diag}({\bf w}_{i-1}) \sim \mathbb{R}^{d\times d}$ ensures that each feature~$f \in [1, \cdots , d]$ of the normalized~$\overline{\bf a}_{i-1}$ is associated with its designated weight value from~${\bf w}_{i-1}$.
\end{enumerate}

\noindent In summary, the forward pass of the layer normalization can be expressed as
\begin{empheq}[box={\forwardBox[{\bf Layer normalization}: forward pass]}]{alignat=2}
{\bf a}_{i} &= \overline{\bf a}_{i-1} \, \widetilde{\bf w}_{i-1} + \widetilde{\bf b}_{i-1} \quad \text{with} \quad \overline{\bf a}_{i-1} = \dfrac{{\bf a}_{i-1} - \widetilde{\boldsymbol{ \mu}} }{\widetilde{\boldsymbol{ \sigma}}}
\label{ln:forward}
\end{empheq}
where the broadcasting rules (using diagonal matrices) of the normalization step with~$(\boldsymbol{ \mu}, \boldsymbol{\sigma})_\texttt{LN}$ and those of the learnable affine transformation step with~$({\bf w}_{i-1}, {\bf b}_{i-1})$ can be understood by identification with eq.\eqref{eq:lnNorm} and~eq.\eqref{eq:affineLN}. \\

\noindent At this point, it is instructive to refer to~Section~9 of the reference paper~\cite{deepPedestrians} dedicated to batch normalization.  Indeed, although we have made the current~eq.\eqref{ln:forward} for layer normalization look identical to eq.(39) of the reference paper for batch normalization, there is a subtle but important difference in the way that the normalized feature vectors~$\overline{\bf a}_{i-1}$ are defined
\begin{itemize}
\item In the case of batch normalization, the mean and standard deviation used for the normalization step are evaluated across the different samples (i.e. tokens in the current context) leading to summary statistics vectors~$(\boldsymbol{ \mu} \, , \boldsymbol{ \sigma})_\texttt{BN} \sim \mathbb{R}^d$ that have the same dimensionality as the feature space (i.e. the number~$n_\mathcal{T}$ of samples/tokens is contracted out).  
\item On the contrary, in the case of layer normalization, these vectors are evaluated across the feature dimension so that each token has its own summary statistics leading to~$(\boldsymbol{ \mu} \, , \boldsymbol{ \sigma})_\texttt{LN} \sim \mathbb{R}^{n_\mathcal{T}}$ (i.e. the dimensionality~$d$ of the feature vectors is contracted out).
\end{itemize}
This difference in the way that the normalization vectors~$(\boldsymbol{ \mu} \, , \boldsymbol{ \sigma})$ are evaluated carries over to the broadcasting rules with the row-wise broadcast of~$\boldsymbol{ \mu}$ for~\texttt{BN} being replaced by column-wise broadcast for~\texttt{LN}.  Similarly the broadcasted division~$1/\boldsymbol{ \widetilde{\sigma}}$ is evaluated via marix multiplication from the right for~\texttt{BN} whereas it is from the left for~\texttt{LN}. Therefore, the crucial observation is that one can go from~\texttt{LN} to~\texttt{BN} and recover all these shape/statistics differences simply by applying the normalization part of~\texttt{BN} to the transpose of our current input data~${\bf a}_{i-1} \sim \mathbb{R}^{n_\mathcal{T}\times d}$ with
\begin{equation}
\texttt{LN}\left( {\bf a}_{i-1}  \right) \sim \mathbb{R}^{n_\mathcal{T}\times d} \,\, \cong \,\, \left[ \texttt{BN}\left( {\bf a}_{i-1}^t  \sim \mathbb{R}^{d\times n_\mathcal{T}} \right) \sim \mathbb{R}^{d\times n_\mathcal{T}} \right]^t 
\label{eq:LNBNmap}
\end{equation}
where we use the~$\cong$ symbol to reflect the fact that the number of parameters in the learnable affine transformation step is different since~\texttt{BN} needs to be applied to the transpose of~${\bf a}_{i-1}$ and that, generally,~$n_\mathcal{T}\neq d$. \\ 

\noindent In other words, the~\texttt{LN} and~\texttt{BN} layers are both composed of two steps~i)~a~\uwave{normalization} for which both layers are exact mirrors of each other \colorbox{Gray}{\bf up to a transpose operation} followed by~ii)~a mechanically identical \uwave{learnable affine transformation}.

\paragraph{Backward pass.}  Thanks to this ``transposed duality'' between~\texttt{LN} and~\texttt{BN}, we can immediately adapt the results of the backward pass derived in eqs.(41-43) of the reference paper~\cite{deepPedestrians} for batch normalization (there) to layer normalization (here) by applying the appropriate transpose operations. \\

\noindent In particular, since the second step~2) relating to the \uwave{learnable affine transformation} does not depend upon the details of how~$\overline{\bf a}_{i-1}$ is evaluated, the gradients of the loss with respect to the weights and biases~$\partial \mathcal{L}_{\text{seq}} / \{ {\bf w}_{i-1}, {\bf b}_{i-1} \} \sim \mathbb{R}^d$ remain unchanged for both~\texttt{BN} and~\texttt{LN} layers and we simply rename ``samples'' to ``tokens'' to better match the current context of sequence models.  \\

\noindent On the other hand, the backpropagation of the error signal from its upstream value~$\boldsymbol{\Delta}_{i} \sim \mathbb{R}^{n_\mathcal{T}\times d}$ to the downstream~$\boldsymbol{\Delta}_{i-1} \sim \mathbb{R}^{n_\mathcal{T}\times d}$ requires to handle the different \uwave{normalizations} of~$\overline{\bf a}_{i-1}$ of step~1)~which, as explained in~\eqref{eq:LNBNmap}, are related to each other via a simple transposition.  Carrying over this mapping from~\texttt{BN} to~\texttt{LN} is done by
\begin{enumerate}
\item copy/pasting the expression of~$\boldsymbol{\Delta}_{i-1}$ as it appears in the backward pass of~\texttt{LN} in~eq.(41) of the reference paper~\cite{deepPedestrians}
\item replacing both~$\overline{\bf a}_{i-1}$ and~$\boldsymbol{\Delta}_{i}$ by their transpose while ensuring consistent dimensionality of the matrix multiplications.  In other words~$ \big( \boldsymbol{\Delta}_i \widetilde{\bf w}_{i-1} \big)_\texttt{BN} \rightarrow  \big( \widetilde{\bf w}_{i-1} \boldsymbol{\Delta}_i^t \big)_\texttt{LN} \sim \mathbb{R}^{d \times n_\mathcal{T}} $
\item replacing sums over samples in~\texttt{BN} by sums over features for~\texttt{LN}
\item performing the outer transpose as shown in~\eqref{eq:LNBNmap} to recover the expected dimensionality of the downstream error signal~$\boldsymbol{\Delta}_{i-1} \sim \mathbb{R}^{n_\mathcal{T}\times d}$
\item bringing the broadcasted division~$1/\boldsymbol{ \widetilde{\sigma}}$ out of the outer transpose so this term continues to appear as applied from the left.  (Since the transpose of a diagonal matrix is equal to itself, there is no need for additional transpose symbols here).  
\item finally, modifying the scaling to~$1/d$ to correctly reflect the fact that normalization is carried out feature-wise in~\texttt{LN} as opposed to sample-wise in~\texttt{BN}.
\end{enumerate}

\noindent In summary, we have
\begin{empheq}[box={\backPropBox[{\bf Layer normalization}: backward pass]}]{alignat=2}
\boldsymbol{\Delta}_{i-1} &= \frac{1}{d \, \widetilde{\boldsymbol{ \sigma}}} \left( d \, \widetilde{\bf w}_{i-1} \boldsymbol{\Delta}_i^t -  \sum_\text{features} \widetilde{\bf w}_{i-1} \boldsymbol{\Delta}_i^t  - \overline{\bf a}_{i-1}^{\,t} \circ \sum_\text{features} \overline{\bf a}_{i-1}^{\,t} \circ \widetilde{\bf w}_{i-1} \boldsymbol{\Delta}_i^t   \right)^t \sim \mathbb{R}^{n_\mathcal{T} \times d} \label{ln::error} \\
\frac{\partial \mathcal{L}_{\text{seq}}}{\partial {\bf w}_{i-1}} &= \text{diag} \left( \overline{\bf a}_{i-1}^{\, t} \boldsymbol{\Delta}_i \right) \sim \mathbb{R}^{d} \label{ln::weight} \\
\frac{\partial \mathcal{L}_{\text{seq}}}{\partial {\bf b}_{i-1}} &= \sum_\text{tokens} \boldsymbol{\Delta}_i  \sim \mathbb{R}^{d} \label{ln::bias}
\end{empheq}

\section{LoRA Layer}

\paragraph{Forward pass.}  Typically, neural networks are composed of numerous fully-connected layers whose purpose is to modify the dimensionality of the feature maps.  Normally, going from an input feature map~${\bf a}_{i-1} \sim \mathbb{R}^{n\times f_{i-1}}$ to an output representation~${\bf a}_i \sim \mathbb{R}^{n\times f_i}$ would require~$(f_{i-1}\times f_i)$ parameters encoded into a weight matrix~${\bf w}_{i-1} \sim \mathbb{R}^{f_{i-1} \times f_i}$ (and maybe even another~$f_i$ parameters if one considers non-null biases~${\bf b}_{i-1} \sim \mathbb{R}^{f_i}$ in addition to the weight matrix).  In the context of this paper~$n \equiv n_\mathcal{T}$ refers to the number of tokens in the sequence whereas in the reference paper~\cite{deepPedestrians} it was referring to the number of samples in a mini-batch. (Regardless of the tokens/samples interpretation, all components are processed independently of each other so there is no distinction to be made as far as fully-connected layers are concerned.) \\ 

\noindent Dimensionality-wise, the same mapping from~$f_{i-1}$ to~$f_i$ may be achieved by decomposing the weight matrix into the product of two new matrices~${\bf d}_{i-1} \sim \mathbb{R}^{f_{i-1}\times r}$ (mapping from~$f_{i-1}$ ``down'' to~$r$) and~${\bf u}_{i-1} \sim \mathbb{R}^{r\times f_i}$ (mapping from~$r$ back ``up'' to~$f_i$).  The product~${\bf d}_{i-1} \, {\bf u}_{i-1} \sim \mathbb{R}^{f_{i-1}\times f_i}$ that composes these two mappings is of the same dimensionality as that of the original weight matrix~${\bf w}_{i-1}$ in fully-connected layers.  The trick is to choose a rank~$r$ such that~$r \ll \text{min}(f_{i-1}, f_i)$.  In this case, the number of parameters associated with this low-rank decomposition~${\bf d}_{i-1} \,{\bf u}_{i-1}$ is therefore
\begin{equation*}
r \times \big( f_{i-1} + f_i \big) \ll \big( f_{i-1} \times f_i \big)
\end{equation*}

\noindent Normally, LoRA layers would not be used as a substitute to linear layers but more as companions for parameter-efficient fine-tuning~\cite{lora}.  In practice, this means that the full linear layers are first trained to produce a large pre-trained model.  Then, during fine-tuning, those weights are kept frozen and LoRA layers are introduced to receive gradient updates specific to the fine-tuning task.  Since the LoRA layers and the original dense linear layers both have the same dimensionalities, the data representations are simply added together at inference time. \\

\noindent In summary, the parameters associated with this LoRA layer are
\begin{equation*}
\mathcal{P}_{i-1} \begin{cases} 
\hspace{0.1cm} {\bf d}_{i-1} \sim \mathbb{R}^{f_{i-1}\times r}  \\
\hspace{0.1cm} {\bf u}_{i-1} \sim \mathbb{R}^{r\times f_i} 
\end{cases}
\end{equation*}
and the forward pass can be summarized as
\begin{empheq}[box={\forwardBox[{\bf LoRA}: forward pass]}]{alignat=2}
{\bf a}_i &= \alpha \, {\bf a}_{i-1} \, {\bf d}_{i-1} \, {\bf u}_{i-1}
\label{lora:forward}
\end{empheq}
where~$\alpha \sim \mathbb{R}$ controls the relative importance of LoRA layers during backpropagation (somewhat analogously to a layer-specific learning rate) and is (usually) chosen such that~$\alpha = r$.

\paragraph{Backward pass.}  The backward pass is evaluated via the usual recursive expression and here it is useful to leverage the \hyperlink{fourMatFrobenius}{cyclic property} of Frobenius products to expand
\begin{align*}
\boldsymbol{\Delta}_i \cdot \text{d}{\bf a}_{i} &= \boldsymbol{\Delta}_i \cdot \text{d} \big( \alpha \, {\bf a}_{i-1} \, {\bf d}_{i-1} \, {\bf u}_{i-1} \big) \\
&= \alpha \, \boldsymbol{\Delta}_i \cdot \text{d} \big( {\bf a}_{i-1} \, {\bf d}_{i-1} \, {\bf u}_{i-1} \big) \\
&= \alpha \, \boldsymbol{\Delta}_i \cdot \big[ \big( \text{d} {\bf a}_{i-1} \big)  \, {\bf d}_{i-1} \, {\bf u}_{i-1} + {\bf a}_{i-1} \big( \text{d} {\bf d}_{i-1} \big) {\bf u}_{i-1} + {\bf a}_{i-1} \, {\bf d}_{i-1} \big( \text{d} {\bf u}_{i-1} \big) \big] \\
&= \underbrace{ \mathcolorbox{shadecolor}{\alpha \big[ \boldsymbol{\Delta}_i \big( {\bf d}_{i-1} {\bf u}_{i-1} \big)^t \big]}}_{\textstyle
    \begin{gathered}
      \boldsymbol{\Delta}_{i-1}
    \end{gathered} } \, \cdot \, \text{d} {\bf a}_{i-1} + \underbrace{ \mathcolorbox{shadecolor}{ \alpha \big( {\bf a}_{i-1}^t \boldsymbol{\Delta}_i {\bf u}_{i-1}^t \big) } }_{\textstyle
    \begin{gathered}
      \frac{\partial \mathcal{L}_{\text{seq}}}{\partial {\bf d}_{i-1}}
    \end{gathered} } \cdot \, \text{d} {\bf d}_{i-1} + \underbrace{ \mathcolorbox{shadecolor}{\alpha \big[ \big( {\bf a}_{i-1} {\bf d}_{i-1} \big)^t \boldsymbol{\Delta}_i \big] } }_{\textstyle
    \begin{gathered}
     \frac{\partial \mathcal{L}_{\text{seq}}}{\partial {\bf u}_{i-1}}
    \end{gathered} } \cdot \, \text{d} {\bf u}_{i-1}
\end{align*}

\noindent In summary, the backward pass through a LoRA layer is given by:
\begin{empheq}[box={\backPropBox[{\bf LoRA}: backward pass]}]{alignat=2}
\boldsymbol{\Delta}_{i-1} &= \alpha \, \boldsymbol{\Delta}_i \big( {\bf d}_{i-1} \, {\bf u}_{i-1} \big)^t  &\quad &\sim \mathbb{R}^{n\times f_{i-1}}  \\
\frac{\partial \mathcal{L}_{\text{seq}}}{\partial {\bf d}_{i-1}} &= \alpha \, {\bf a}_{i-1}^t \boldsymbol{\Delta}_i {\bf u}_{i-1}^t &\quad  &\sim \mathbb{R}^{f_{i-1}\times r} \\
\frac{\partial \mathcal{L}_{\text{seq}}}{\partial {\bf u}_{i-1}} &= \alpha \, \big( {\bf a}_{i-1} {\bf d}_{i-1} \big)^t \boldsymbol{\Delta}_i &\quad  &\sim \mathbb{R}^{r \times f_i} 
\end{empheq}

\noindent We can see that~$\alpha$ acts as a multiplicative scaling factor to influence the gradient updates in a way similar to learning rate scaling (although acting specifically on LoRA layers only).

\section{Conclusion: A minimalistic transformer-based architecture}
\label{sec:TransformerBlock}

\paragraph{Minimalistic architecture.} Released by OpenAI in 2019, GPT-2 may still be used as a reference to illustrate transformer-based networks.  In this note we reproduce a smaller version of this architecture as specified in~Table~\ref{table:minimalTransformer}.  Complete expressions for all parameter gradients are provided in~Table~\ref{table:minimalTransformer_grads}. \\

\noindent After a tokenizer has already processed the input sequence, the resulting input tokens~${\bf a}_0 \sim \mathbb{N}^{n_\mathcal{T}}$ are transformed into token and position embeddings~${\bf a}_1^\text{tok} \sim {\bf a}_1^\text{pos} \sim \mathbb{R}^{n_\mathcal{T}\times d}$ of the same dimensionality via weight matrices~${\bf w}_\text{tok} \sim \mathbb{R}^{n_\text{vocab}\times d}$ and~${\bf w}_\text{pos} \sim \mathbb{R}^{n_\text{context}\times d}$.  In keeping with the ``pedestrian'' spirit of this note, we follow the ``small'' version of~GPT-2 with
\begin{equation*}
\mathcolorbox{Gray}{\Big( d = 768 \,\,\, ; \, n_\text{context} = 1,024 \,\,\, ; \, n_\text{vocab} = 50,257 \Big)}
\end{equation*} 
Other versions of~GPT-2 differ only in the values of these parameters without any modification to the architecture itself.  Both embedding representations are added to each other~${\bf a}_1 = {\bf a}_1^\text{tok} + {\bf a}_1^\text{pos}$ and serve as input to the \colorbox{brilliantlavender}{transformer block}.  Generically, a transformer block is composed of two functional sublayers each wrapped in a~\big(LayerNorm~$\triangleright$~Sublayer~$\triangleright$~Skip/Add \big) pattern.  Those sublayers consist of 
\begin{itemize}
\item \uline{``Self-attention''} sublayer~$\equiv$~\big(MHA~$\triangleright$~FC$_\text{attProj}$\big).  For the sake of clarity we separate the pure~MHA part described in Section~\ref{sec:multiHead} from its final output projection~FC$_\text{attProj}$. (Standard implementations of self-attention typically keep both steps as a single integrated layer.)
\item \uline{``Expand-and-contract''} sublayer~$\equiv$~\big(FC$_\text{expand}$~$\triangleright$~$g$~$\triangleright$~FC$_\text{contract}$\big).  The first fully-connected layer expands the dimensionality of the feature maps from~$d$ to~$4d$ and the second one contracts it back to~$d$ after having gone through a non-linear activation function~$g$ such as~ReLU, GELU$\dots$
\end{itemize}

\noindent Denoting by~$\triangleright$ the ``left-to-right'' (forward) function composition operator, the architecture of a complete transformer block is summarized visually in the diagram below with~${\bf a}_1 \sim \mathbb{R}^{n_\mathcal{T}\times d}$ as the input data and~${\bf a}_{10} \sim \mathbb{R}^{n_\mathcal{T}\times d}$ as the output data representation after processing by the transformer block.

\begin{adjustwidth}{-2.2cm}{0cm}
\begin{center}
\begin{tikzpicture}[baseline=(current bounding box.center), node distance=0.3cm, background rectangle/.style={fill=brilliantlavender},  show background rectangle, rounded corners=12pt]

\usetikzlibrary{positioning, calc}

\node[fill=green-yellow, inner sep=5pt] (a1)  {${\bf a}_1$};
\node (ln1) [right=of a1] {LayerNorm};
\node (mha) [right=of ln1] {\Big( MHA};
\node (fc)  [right=of mha] {FC$_\text{attProj}$ \Big)};
\node (add1) [right=of fc] {Skip/Add};
\node (ln2) [right=of add1] {LayerNorm};
\node (fc1) [right=of ln2] {\Big( FC$_\text{expand}$ };
\node (act) [right=of fc1] {$g$};
\node (fc2) [right=of act] {FC$_\text{contract}$ \Big)};
\node (add2) [right=of fc2] {Skip/Add};
\node[fill=green-yellow, inner sep=5pt] (a10) [right=of add2] {${\bf a}_{10}$};

\draw[draw=none] (a1) -- node[midway] {$\triangleright$} (ln1);
\draw[draw=none] (ln1) -- node[midway] {$\triangleright$} (mha);
\draw[draw=none] (mha) -- node[midway] {$\triangleright$} (fc);
\draw[draw=none] (fc) -- node[midway] {$\triangleright$} (add1);
\draw[draw=none] (add1) -- node[midway] {$\triangleright$} (ln2);
\draw[draw=none] (ln2) -- node[midway] {$\triangleright$} (fc1);
\draw[draw=none] (fc1) -- node[midway] {$\triangleright$} (act);
\draw[draw=none] (act) -- node[midway] {$\triangleright$} (fc2);
\draw[draw=none] (fc2) -- node[midway] {$\triangleright$} (add2);
\draw[draw=none] (add2) -- node[midway] {$\triangleright$} (a10);

\coordinate (mid1) at ($(a1.north) + (0,0.5)$);
\coordinate (mid2) at ($(add1.north) + (0,0.5)$);
\coordinate (mid3) at ($(add1.north) + (0,0)$);

\coordinate (mid4) at ($(add1.south) - (0,0.5)$);
\coordinate (mid5) at ($(add2.south) - (0,0.5)$);
\coordinate (mid6) at ($(add2.south) - (0,0)$);

\draw[-, dashed] (a1.north) -- (mid1);
\draw[-, dashed] (mid1) -- node[below] {$\triangleright$} (mid2); 
\draw[-, dashed] (mid2) -- (mid3);

\draw[-, dashed] (add1.south) -- (mid4);             
\draw[-, dashed] (mid4) -- node[above] {$\triangleright$} (mid5); 
\draw[-, dashed] (mid5) -- (mid6);

\end{tikzpicture}
\end{center}
\end{adjustwidth} 

\noindent \\

\noindent The step-by-step breakdown from~${\bf a}_1$ to~${\bf a}_{10}$ is presented in~Table~\ref{table:minimalTransformer} where the layers belonging to the transformer block are \colorbox{brilliantlavender}{color highlighted}.  This construction of transformer blocks as two functional sublayers may also be visualized as
\begin{align*}
{\bf a}_5 &= {\bf a}_1 + \text{FC}_\text{attProj} \big[ \text{MHA} \left( \text{LayerNorm} \, \mathcolorbox{green-yellow}{{\bf a}_1}  \right) \big] \\
\mathcolorbox{green-yellow}{{\bf a}_{10}} &= {\bf a}_5 + \text{FC}_\text{contract} \Big( g \Big[ \text{FC}_\text{expand} \big( \text{LayerNorm} \,  {\bf a}_5 \big) \Big] \Big)
\end{align*}

\noindent Instead of feeding the ouptut~${\bf a}_{10}$ of the transformer block back as an input into another transformer block (with~$n_\text{blocks} = 12$ back-to-back blocks in ``small''~GPT-2), we pass~${\bf a}_{10}$ directly into a final set of layer normalization and fully-connected layer~$\text{FC}_\text{logits}$ to produce the ``logits''~${\bf a} \sim \mathbb{R}^{n_\mathcal{T} \times n_\text{vocab}}$ which are ultimately converted, via a Softmax function, into~$n_\mathcal{T}$ probability distributions~${\bf y}_\text{pred} \sim \mathbb{R}^{n_\mathcal{T} \times n_\text{vocab}}$ over the vocabulary for all tokens in the sequence.  These last steps are summarized as
\begin{equation*}
{\bf y}_\text{pred} = {\bf a}_{10} \,\,\triangleright \,\, \text{LayerNorm} \,\,\triangleright \,\, \text{FC}_\text{logits} \,\, \triangleright \,\, \text{Softmax} 
\end{equation*}
The complete architecture of our minimalistic GPT-2 version with a single~$n_\text{blocks} = 1$ transformer block is presented in~Table~\ref{table:minimalTransformer}.

\paragraph{How many parameters does the model have?}  Overall, the number of parameters in a transformer block is given by
\begin{equation*}
n_\text{params}^\text{(block)} = \bigg[ \underbrace{ \mathcolorbox{Gray}{ (2 \times d) }}_{\textstyle
\begin{gathered}
\text{LN(1)}
\end{gathered} } + \underbrace{ \mathcolorbox{Gray}{ (3 + 1) \times \big[ (d \times d) + d \big] }}_{\textstyle
\begin{gathered}
\text{MHA} + \text{FC}_\text{attProj}
\end{gathered} }  + \underbrace{ \mathcolorbox{Gray}{ (2 \times d) }}_{\textstyle
\begin{gathered}
\text{LN(2)}
\end{gathered} } + \underbrace{ \mathcolorbox{Gray}{ \big[ (d \times 4d)+ 4d \big] }}_{\textstyle
\begin{gathered}
\text{FC}_\text{expand}
\end{gathered} } + \underbrace{ \mathcolorbox{Gray}{ \big[ (4d \times d)+ d \big] }}_{\textstyle
\begin{gathered}
\text{FC}_\text{contract}
\end{gathered} } \bigg] 
\end{equation*}

\noindent Using the standard~GPT-2 values, each transformer block therefore contains~$n_\text{params}^\text{(block)} = 7,087,872$ parameters.  Adding on the parameters associated with token/position embeddings, final layer normalization and fully-connected layer (to produce the logits), we end up with a total number of parameters given by
\begin{equation*}
n_\text{params} = \bigg[ \underbrace{ \mathcolorbox{Gray}{ (n_\text{vobab} \times d) }}_{\textstyle
\begin{gathered}
\text{token emdedding}
\end{gathered} }  + \underbrace{ \mathcolorbox{Gray}{ (n_\text{context} \times d ) }}_{\textstyle
\begin{gathered}
\text{position emdedding}
\end{gathered} } + \underbrace{ \mathcolorbox{Gray}{ (n_\text{blocks} \times n_\text{params}^\text{(block)}) }}_{\textstyle
\begin{gathered}
\text{transformer blocks}
\end{gathered} } + \underbrace{ \mathcolorbox{Gray}{ (2 \times d) }}_{\textstyle
\begin{gathered}
\text{LN}\text{(final)}
\end{gathered} } + \underbrace{ \mathcolorbox{Gray}{  \big[ (d \times n_\text{vocab}) + n_\text{vocab} \big]  }}_{\textstyle
\begin{gathered}
\text{FC}_\text{logits}
\end{gathered} } \bigg] 
\end{equation*}
where~$n_\text{blocks}$ denotes the number of transformer blocks.  \\

\noindent In the minimalistic network specified in Table~\ref{table:minimalTransformer}, we have a single transformer block~$n_\text{blocks} = 1$ for a total of~$n_\text{params} = 85,120,849$ parameters.  With~$n_\text{blocks} = 12$, the complete~GPT-2 network has a total of~$n_\text{params}^\text{(gpt2)} = 163,087,441$ parameters.  \\

\noindent Note that this is only about twice the number of parameters compared to our minimalistic network even though there are~12 tranformer blocks instead of a single one. \\

\noindent In practice, it is common to tie the weights of the token embedding layer~${\bf w}_\text{tok} \sim \mathbb{R}^{n_\text{vocab}\times d} $ together with those of the final fully-connected layer~$\text{FC}_\text{logits} \sim \mathbb{R}^{d \times n_\text{vocab}} $ since those have the same dimensionality (up to a transpose) and account for a large number of parameters~$n_\text{vocab} \times d \approx 39,000,000$.  In this ``weight-tying'' scenario, one simply ignores the biases from~$\text{FC}_\text{logits}$ and, instead of learning two independent weight matrices, the model learns only one matrix.  This optimization reduces the number of parameters from~$\approx 163,000,000$ down to~$\approx 124,000,000$ leading not only to~$\approx 24\%$ savings in parameter count but may also act as a mild regularizer that enforces consistency between input and output representations.

\paragraph{With LoRA: How many parameters now?}  As an illustration of LoRA fine-tuning, let us replace the last fully-connected layer in Table~\ref{table:minimalTransformer} by a LoRA layer.  In this case, the \colorbox{orange}{forward pass} is given by
\begin{equation*}
\mathcolorbox{green-yellow}{{\bf a} \equiv {\bf a}_{12} = \texttt{FC}_\text{frozen}({\bf a}_{11}) + \texttt{LoRA}({\bf a}_{11}) = \texttt{FC}_\text{frozen}({\bf a}_{11}) + \alpha \, {\bf a}_{11} \, {\bf d}_{11} \, {\bf u}_{11}}
\end{equation*}
where~$\texttt{FC}_\text{frozen}$ indicates that the weights of the fully-connected layer are frozen and will not be updated during the \colorbox{light-blue}{backward pass}
\begin{equation*}
\mathcolorbox{shadecolor}{\boldsymbol{\Delta}_{11} = \alpha \, \boldsymbol{\Delta}_{12} \big( {\bf d}_{11} \, {\bf u}_{11} \big)^t  \quad ; \quad \frac{\partial \mathcal{L}_{\text{seq}}}{\partial {\bf d}_{11}} = \alpha \, {\bf a}_{11}^t \boldsymbol{\Delta}_{12} {\bf u}_{11}^t \quad ; \quad \frac{\partial \mathcal{L}_{\text{seq}}}{\partial {\bf u}_{11}} = \alpha \, \big( {\bf a}_{11} {\bf d}_{11} \big)^t \boldsymbol{\Delta}_{12}}
\end{equation*}

\noindent Instead of having a fully-connected layer with~$(d \times n_\text{vocab}) + n_\text{vocab} = 38,647,633$ parameters that need to be optimized in the backward pass, using the LoRA layer reduces the number of trainable parameters down to~$r \times (d + n_\text{vocab}) = 816,400$ which is about~$\approx 2\%$ of the original amount (using a standard rank of~$r=16$).

\begin{equation*}
n_\text{params}^\text{(lora)} = n_\text{blocks} \times \bigg[  \underbrace{ \mathcolorbox{Gray}{ (3 + 1) \times \big[ r \times (d + d) \big] }}_{\textstyle
\begin{gathered}
\text{MHA}_\text{(LoRA)} + \text{LoRA}_\text{attProj}
\end{gathered} } + \underbrace{ \mathcolorbox{Gray}{ \big[ r \times (d + 4d) \big] }}_{\textstyle
\begin{gathered}
\text{LoRA}_\text{expand}
\end{gathered} } + \underbrace{ \mathcolorbox{Gray}{ \big[ r \times (4d + d) \big] }}_{\textstyle
\begin{gathered}
\text{LoRA}_\text{contract}
\end{gathered} } \bigg] + \underbrace{ \mathcolorbox{Gray}{ \big[ r \times (d + n_\text{vocab}) \big] }}_{\textstyle
\begin{gathered}
\text{LoRA}_\text{logits}
\end{gathered} }
\end{equation*}

\noindent With~$n_\text{blocks} = 12$, we get~$n_\text{params}^\text{(lora)} = 3,470,608$ to be compared with~$n_\text{params}^\text{(gpt2)} = 163,087,441$ for the complete GPT-2 (without weight-tying) which is consistent with a~$\approx 98\%$ reduction in number of parameters.

\begin{table}
\hspace*{-2.69cm}
\captionsetup{singlelinecheck=off}
\begin{tabular}{|| l | y | x | z ||}
\hline
\rule{0pt}{1.001\normalbaselineskip}
{\bf Layer} & \cellcolor{orange}{\bf Forward pass} & {\bf Shape} & \cellcolor{light-blue}{\bf Backward pass} \\
\hline
\hline
\rule{0pt}{1.1\normalbaselineskip}
Input data & ${\bf a}_0$ & $\mathbb{N}^{n_\mathcal{T}}$ &  \cellcolor{Gray} Sequence of~$n_\mathcal{T}$ tokens encoded as integers~$t \sim \mathbb{N}$ \\
\hline \hline \rule{0pt}{1.1\normalbaselineskip}
\hspace{-0.25cm} Token embedding & ${\bf a}_{1}^\text{tok} = \text{ohe}({\bf a}_0) \, {\bf w}_\text{tok}  $ & $ \mathbb{R}^{n_\mathcal{T}\times d}  $ &  \cellcolor{Gray} \\[0.6em]
Position embedding & ${\bf a}_{1}^\text{pos} = \text{ohe}(1\!:\!n_{\mathcal{T}}) \, {\bf w}_\text{pos} $ & $ \mathbb{R}^{n_\mathcal{T}\times d}  $ & \cellcolor{Gray} \\[0.6em] \hline \rule{0pt}{1.1\normalbaselineskip}
Input embedding & ${\bf a}_{1} = {\bf a}_{1}^\text{tok} + {\bf a}_{1}^\text{pos}$  & $ \mathbb{R}^{n_\mathcal{T}\times d}  $ & $\boldsymbol{\Delta}_1 =  \boldsymbol{\Delta}_5 + \frac{1}{\vphantom{^X} d \, \widetilde{\boldsymbol{ \sigma_1}}} \bigg[ \bigg( d \, \widetilde{\bf w}_1 \boldsymbol{\Delta}_2^t -  \sum \widetilde{\bf w}_1 \boldsymbol{\Delta}_2^t - \overline{\bf a}_1^{\,t} \circ \sum \overline{\bf a}_1^{\,t} \circ \widetilde{\bf w}_1 \boldsymbol{\Delta}_2^t \bigg)^t \,\, \bigg] $ \\[0.6em]
\hline \hline \rule{0pt}{1.1\normalbaselineskip}
\cellcolor{brilliantlavender} \hspace{-0.4cm} $\star$ 1\textsuperscript{st} LayerNorm & ${\bf a}_2 = \overline{\bf a}_1 \, \text{diag}\left( {\bf w}_1 \right) + \widetilde{\bf b}_1 $ & $\mathbb{R}^{n_\mathcal{T}\times d}  $  & $ \boldsymbol{\Delta}_2 = \sum_h \left( \boldsymbol{\rho}_{({\bf a}_{2}, \, h) }^t \boldsymbol{\Delta}^h_3 {\bf w}^t_{v_h} + \boldsymbol{\Delta}^h_\text{3\,(raw)} \, {\bf k}_h {\bf w}_{q_h}^t + (\boldsymbol{\Delta}^h_\text{3\,(raw)})^t \, {\bf q}_h \, {\bf w}_{k_h}^t \right) $ \\[0.6em]
\hline \rule{0pt}{1.1\normalbaselineskip}
\cellcolor{brilliantlavender} & & \cellcolor{white} & \cellcolor{white} $ \boldsymbol{\Delta}_3$ \text{split into different heads} $\rightarrow\left[ \boldsymbol{\Delta}_3^{(h=1)}  \, , \cdots , \,\boldsymbol{\Delta}_3^{(h=n_h)} \right] $ \\
\cellcolor{brilliantlavender}  &  & \multirow{-2}*{\cellcolor{white} $\mathbb{R}^{n_\mathcal{T}\times d_h}$} & \cellcolor{white} $ [ ({\bf w}_{v}, {\bf w}_{q}, {\bf w}_{k} )_h \, , \, (\boldsymbol{\rho}_{{\bf a}_{2} }, {\bf k}, {\bf q} )_h ]$ $\sim$ internal to ${\bf a}_3
$ in  \colorbox{green-yellow}{\texttt{MHA} forward} \\[0.3em]
\cline{3-4}
\multirow{-3}*{\cellcolor{brilliantlavender} \hspace{-0.3cm} $\star$ Multi-headed attention} & \multirow{-3}*{${\bf a}_3 = \texttt{MHA} \, {\bf a}_2  $} & $\mathbb{R}^{n_\mathcal{T} \times d}$ & $ \boldsymbol{\Delta}_3 = \boldsymbol{\Delta}_{5} \, {\bf w}_{3}^t   $ \\
\hline \rule{0pt}{1.1\normalbaselineskip}
\cellcolor{brilliantlavender} \hspace{-0.4cm} $\star$ FC$_\text{attProj}$ & ${\bf a}_4 = {\bf a}_3 {\bf w}_3 + \widetilde{{\bf b}}_3 $ & $\mathbb{R}^{n_\mathcal{T}\times d}  $  & $ \boldsymbol{\Delta}_5 $ \\[0.6em]
\hline \rule{0pt}{1.1\normalbaselineskip}
\cellcolor{brilliantlavender} & & & $\boldsymbol{\Delta}_5 = \boldsymbol{\Delta}_{10} + \frac{1}{\vphantom{^X} d \, \widetilde{\boldsymbol{ \sigma_5}}} \bigg[ \bigg( d \, \widetilde{\bf w}_5 \boldsymbol{\Delta}_6^t -  \sum \widetilde{\bf w}_5 \boldsymbol{\Delta}_6^t - \overline{\bf a}_5^{\,t} \circ \sum \overline{\bf a}_5^{\,t} \circ \widetilde{\bf w}_5 \boldsymbol{\Delta}_6^t \bigg)^t \,\, \bigg] $  \\
\cellcolor{brilliantlavender} \multirow{-2}*{\hspace{-0.3cm} $\star$  1\textsuperscript{st} Skip/Add} & \multirow{-2}*{${\bf a}_5 = {\bf a}_1 + {\bf a}_4 $} & \multirow{-2}*{$\mathbb{R}^{n_\mathcal{T}\times d}  $} & $\boldsymbol{\Delta}_5 \rightarrow$ ``dispatched'' to ${\bf a}_1$ and ${\bf a}_4$ \\[0.2em]
\hline \hline \rule{0pt}{1.1\normalbaselineskip}
\cellcolor{brilliantlavender} \hspace{-0.4cm} $\star$ 2\textsuperscript{nd} LayerNorm & ${\bf a}_6 = \overline{\bf a}_5 \, \text{diag}\left( {\bf w}_5 \right) + \widetilde{\bf b}_5 $ & $\mathbb{R}^{n_\mathcal{T}\times d}  $  & $ \boldsymbol{\Delta}_6 = \boldsymbol{\Delta}_{7} \, {\bf w}_{6}^t  $ \\[0.3em]
\hline  \rule{0pt}
{1.1\normalbaselineskip}
\cellcolor{brilliantlavender} \hspace{-0.4cm} $\star$ FC$_\text{expand}$ & ${\bf a}_7 = {\bf a}_6 {\bf w}_6 + \widetilde{{\bf b}}_6 $ & $\mathbb{R}^{n_\mathcal{T}\times 4d} $  & $\boldsymbol{\Delta}_7 = \boldsymbol{\Delta}_8 \circ g^\prime({\bf a}_7)$ \\[0.3em]
\hline \rule{0pt}
{1.1\normalbaselineskip}
\cellcolor{brilliantlavender} \hspace{-0.4cm} $\star$ Activation & ${\bf a}_8 = g({\bf a}_7) $  & $\mathbb{R}^{n_\mathcal{T}\times 4d} $ & $\boldsymbol{\Delta}_8 = \boldsymbol{\Delta}_{10} \, {\bf w}_{8}^t$ \\[0.3em]
\hline \rule{0pt}
{1.1\normalbaselineskip}
\cellcolor{brilliantlavender} \hspace{-0.4cm} $\star$ FC$_\text{contract}$ & ${\bf a}_9 = {\bf a}_8 {\bf w}_8 + \widetilde{{\bf b}}_8 $ & $\mathbb{R}^{n_\mathcal{T}\times d} $ & $\boldsymbol{\Delta}_{10}$ \\[0.3em]
\hline \rule{0pt}
{1.1\normalbaselineskip}
\cellcolor{brilliantlavender} & & & \hspace{-0.25cm} $\boldsymbol{\Delta}_{10} = \frac{1}{\vphantom{^X} d \, \widetilde{\boldsymbol{ \sigma_{10}}}} \bigg[ \bigg( d \, \widetilde{\bf w}_{10} \boldsymbol{\Delta}_{11}^t -  \sum \widetilde{\bf w}_{10}\boldsymbol{\Delta}_{11}^t - \overline{\bf a}_{10}^{\,t} \circ \sum \overline{\bf a}_{10}^{\,t} \circ \widetilde{\bf w}_{10} \boldsymbol{\Delta}_{11}^t \bigg)^t \, \bigg] \! $  \\ 
\cellcolor{brilliantlavender} \multirow{-2}*{\hspace{-0.3cm} $\star$ 2\textsuperscript{nd} Skip/Add} & \multirow{-2}*{${\bf a}_{10} = {\bf a}_5 + {\bf a}_9 $} & \multirow{-2}*{$\mathbb{R}^{n_\mathcal{T}\times d}  $} & $\boldsymbol{\Delta}_{10} \rightarrow$ ``dispatched'' to ${\bf a}_5$ and ${\bf a}_9$ \\[0.2em]
\hline \hline \rule{0pt}{1.1\normalbaselineskip}
LayerNorm(final) & \hspace{-0.26cm} ${\bf a}_{11} \!=\! \overline{\bf a}_{10} \text{diag}\left( {\bf w}_{10} \right) \! + \!\widetilde{\bf b}_{10} \!$ & $\mathbb{R}^{n_\mathcal{T}\times d} $ & $\boldsymbol{\Delta}_{11} = \boldsymbol{\Delta}_{12} \, {\bf w}_{11}^t$ \\[0.3em]
\hline \hline \rule{0pt}
{1.1\normalbaselineskip}
FC$_\text{logits}$ & \hspace{-0.24cm} ${\bf a} \equiv {\bf a}_{12} = {\bf a}_{11} {\bf w}_{11} + \widetilde{{\bf b}}_{11}$ & $\mathbb{R}^{n_\mathcal{T}\times n_\text{vocab}}  $  & $ \boldsymbol{\Delta}_{12} = {\bf y}_\text{pred} - {\bf y}_\text{gt} $ \\[0.3em]
\hline \hline \rule{0pt}
{1.1\normalbaselineskip}
Softmax & ${\bf y}_\text{pred} = \text{softmax} \, {\bf a} $ & $ \mathbb{R}^{n_\mathcal{T}\times n_\text{vocab}} $ & \cellcolor{Gray} probability distributions over $n_\text{vocab}$ classes for all~$n_\mathcal{T}$ tokens \\[0.3em]
\hline
\end{tabular}
\caption{{\bf Minimalistic transformer architecture for next-token prediction.}  Other than the fact that this network has only a single \colorbox{brilliantlavender}{transformer block} (as opposed to~$n_\text{blocks}=12$ back-to-back in GPT-2), its structure is conceptually identical in all other aspects to the GPT-like family of models. The fully connected layer immediately after the multi-headed attention layer allows to mix information across different heads.  {\it (Instead of a simple concatenation as is done in our bare-bone~\texttt{MHA}, deep learning frameworks usually incorporate this fully-connected layer directly as part of their multi-headed attention APIs as an ``output projection'' but we keep them separate here for the sake of clarity.  In case that the attention heads are not returning~$d_h = d / n_h$ dimensional feature maps, this output projection can also be used to bring the dimensionality of the feature maps back to~$d$.)}  Thanks to the softmax function,~${\bf y}_\text{pred}$ represent the next-token predicted probability distributions.  For example, consider a specific token~$t^\star$, its prediction vector~${\bf y}_\text{pred}(t^\star) \sim \mathbb{R}^{n_\text{vocab}}$ is such that~$\sum {\bf y}_\text{pred}(t^\star) = 1$.  Different sampling strategies may be applied to select tokens from these probability distributions.
}
\label{table:minimalTransformer}
\end{table}

\begin{table}
\hspace*{-1.5cm}
\captionsetup{singlelinecheck=off}
\begin{tabular}{|| l | y | y | x | x | z ||}
\hline
\multicolumn{3}{|c|}{\cellcolor{orange}{{\bf Parameters}}} & \multicolumn{2}{c|}{\cellcolor{Gray}{{\bf Dimensionality}}} & \cellcolor{light-blue}{\bf Loss derivative} \\
\hline
\hline
\rule{0pt}{1.1\normalbaselineskip}
\hspace{-0.25cm} Token embedding & & ${\bf w}_\text{tok}$ & $\mathbb{R}^{n_\text{vocab} \times d}$ & $38,597,376$ & $\partial \mathcal{L}_{\text{batch}} / \partial {\bf w}_\text{tok} = \text{ohe}({\bf a}_{0})^t \, \boldsymbol{\Delta}_{1} $  \\[0.3em]
Position embedding & \multirow{-2}*{$\mathbfcal{P}_\text{emb}$} & ${\bf w}_\text{pos}$ & $\mathbb{R}^{n_\text{context} \times d}$ & $786,432$ & $\partial \mathcal{L}_{\text{batch}} / \partial {\bf w}_\text{pos} = \text{ohe}(1\!:\!n_{\mathcal{T}})^t \, \boldsymbol{\Delta}_{1} $ \\[0.3em]
\hline \hline
\rule{0pt}{1.1\normalbaselineskip}
\cellcolor{brilliantlavender} & & ${\bf w}_{1}$ & $\mathbb{R}^{d}$ & $ 768 $ & $\partial \mathcal{L}_{\text{seq}} / \partial {\bf w}_{1} = \text{diag} \left( \overline{\bf a}_{1}^{\, t} \boldsymbol{\Delta}_2 \right) $  \\[0.3em]
\multirow{-2}*{\cellcolor{brilliantlavender} 1\textsuperscript{st} LayerNorm} & \multirow{-2}*{$\mathbfcal{P}_{1}$} & ${\bf b}_{1}$ & $\mathbb{R}^{d} $ & $ 768$ & $\partial \mathcal{L}_{\text{seq}} / \partial {\bf b}_{1} = \sum_\text{tokens} \boldsymbol{\Delta}_2  $ \\[0.3em]
\hline \hline
\rule{0pt}{1.1\normalbaselineskip}
\rule{0pt}{1.1\normalbaselineskip}
\cellcolor{brilliantlavender} & & $ {\bf w}_{q_h} $ & $ \mathbb{R}^{d\times d_\rho} $ & $ 589,824 / n_h $ & $\partial \mathcal{L}_{\text{seq}} / \partial {\bf w}_{q_h} = {\bf a}_{2}^t \, \boldsymbol{\Delta}^h_\text{3(raw)} \, {\bf k}_h $  \\[0.3em]
\multirow{-2}*{\cellcolor{brilliantlavender} Multi-headed attention} & \multirow{-2}*{$ \mathbfcal{P}_{Q_h} $} & $ {\bf b}_{q_h} $ & $\mathbb{R}^{d_\rho} $ & $ 768 / n_h $ & $\partial \mathcal{L}_{\text{seq}} / \partial {\bf b}_{q_h} = \sum_\text{tokens} \boldsymbol{\Delta}^h_\text{3(raw)} \, {\bf k}_h $ \\[0.3em]
\rule{0pt}{1.1\normalbaselineskip}
\rule{0pt}{1.1\normalbaselineskip}
\cellcolor{brilliantlavender} & & ${\bf w}_{k_h}$ & $\mathbb{R}^{d\times d_\rho}$ & $ 589,824 / n_h $ & $\partial \mathcal{L}_{\text{seq}} / \partial {\bf w}_{k_h} =  \, (\boldsymbol{\Delta}^h_\text{3(raw)} \, {\bf a}_{2} )^t \, {\bf q}_h  $  \\[0.3em]
\multirow{-2}*{\cellcolor{brilliantlavender} {\it (  Queries, Keys, Values ) } } & \multirow{-2}*{$ \mathbfcal{P}_{K_h} $} & ${\bf b}_{k_h}$ & $\mathbb{R}^{d_\rho} $ & $ 768 /n_h$ & $\partial \mathcal{L}_{\text{seq}} / \partial {\bf b}_{k_h} \equiv {\bf 0} $ \\[0.3em]
\rule{0pt}{1.1\normalbaselineskip}
\rule{0pt}{1.1\normalbaselineskip}
\cellcolor{brilliantlavender} & & ${\bf w}_{v_h}$ & $\mathbb{R}^{d\times d_h}$ & $ 589,824 / n_h $ & $\partial \mathcal{L}_{\text{batch}} / \partial {\bf w}_{v_h} = \big( \boldsymbol{\rho}_{({\bf a}_{2}, \, h) } \, {\bf a}_{2} \big)^t \boldsymbol{\Delta}^h_3 $  \\[0.3em]
\multirow{-2}*{\cellcolor{brilliantlavender} for all attention heads $h \in [\, 1, \cdots , n_h ]$} & \multirow{-2}*{$ \mathbfcal{P}_{V_h} $} & $ {\bf b}_{v_h} $ & $\mathbb{R}^{d_h} $ & $ 768 / n_h $ & $\partial \mathcal{L}_{\text{batch}} / \partial {\bf b}_{v_h} = \sum_\text{tokens} \boldsymbol{\Delta}_3^h $ \\[0.3em]
\hline \hline
\rule{0pt}{1.1\normalbaselineskip}
\cellcolor{brilliantlavender} & & ${\bf w}_{3}$ & $\mathbb{R}^{d \times d}$ & $ 589,824 $ & $\partial \mathcal{L}_{\text{seq}} / \partial {\bf w}_{3} = {\bf a}_3^t \,\boldsymbol{\Delta}_{5} $  \\[0.3em]
 \multirow{-2}*{\cellcolor{brilliantlavender}  FC$_\text{attProj}$}  & \multirow{-2}*{$\mathbfcal{P}_{3} $} & ${\bf b}_{3}$ & $\mathbb{R}^{d} $ & $ 768 $ & $\partial \mathcal{L}_{\text{seq}} / \partial {\bf b}_{3} = \sum_\text{tokens} \boldsymbol{\Delta}_{5} $ \\[0.3em]
\hline \hline
\rule{0pt}{1.1\normalbaselineskip}
\cellcolor{brilliantlavender} & & ${\bf w}_{5}$ & $\mathbb{R}^{d}$ & $ 768 $ & $\partial \mathcal{L}_{\text{seq}} / \partial {\bf w}_{5} = \text{diag} \left( \overline{\bf a}_{5}^{\, t} \boldsymbol{\Delta}_6 \right)  $  \\[0.3em]
\multirow{-2}*{\cellcolor{brilliantlavender} 2\textsuperscript{nd} LayerNorm} & \multirow{-2}*{$\mathbfcal{P}_{5}$} & ${\bf b}_{5}$ & $\mathbb{R}^{d} $ & $ 768$ & $\partial \mathcal{L}_{\text{seq}} / \partial {\bf b}_{5} = \sum_\text{tokens} \boldsymbol{\Delta}_6 $ \\[0.3em]
\hline \hline
\rule{0pt}{1.1\normalbaselineskip}
\cellcolor{brilliantlavender} & & ${\bf w}_{6}$ & $\mathbb{R}^{d \times 4d}$ & $ 2,359,296 $ & $\partial \mathcal{L}_{\text{seq}} / \partial {\bf w}_{6} = {\bf a}_6^t \,\boldsymbol{\Delta}_{7} $  \\[0.3em]
 \multirow{-2}*{\cellcolor{brilliantlavender}  FC$_\text{expand}$}  & \multirow{-2}*{$\mathbfcal{P}_{6} $} & ${\bf b}_{6}$ & $\mathbb{R}^{4d} $ & $ 3072 $ & $\partial \mathcal{L}_{\text{seq}} / \partial {\bf b}_{6} = \sum_\text{tokens} \boldsymbol{\Delta}_{7} $ \\[0.3em]
\hline \hline
\rule{0pt}{1.1\normalbaselineskip}
\cellcolor{brilliantlavender} & & ${\bf w}_{8}$ & $\mathbb{R}^{4d \times d}$ & $ 2,359,296 $ & $\partial \mathcal{L}_{\text{seq}} / \partial {\bf w}_{8} = {\bf a}_8^t \,\boldsymbol{\Delta}_{10} $  \\[0.3em]
 \multirow{-2}*{\cellcolor{brilliantlavender}  FC$_\text{contract}$}  & \multirow{-2}*{$\mathbfcal{P}_{8} $} & ${\bf b}_{8}$ & $\mathbb{R}^{d} $ & $ 768 $ & $\partial \mathcal{L}_{\text{seq}} / \partial {\bf b}_{8} = \sum_\text{tokens} \boldsymbol{\Delta}_{10} $ \\[0.3em]
\hline \hline
\rule{0pt}{1.1\normalbaselineskip}
\multirow{2}*{LayerNorm(final)} & & ${\bf w}_{10}$ & $\mathbb{R}^{d}$ & $ 768$ & $\partial \mathcal{L}_{\text{seq}} / \partial {\bf w}_{10} =  \text{diag} \left( \overline{\bf a}_{10}^{\, t} \boldsymbol{\Delta}_{11} \right) $  \\[0.3em]
& \multirow{-2}*{$\mathbfcal{P}_{10} $} & ${\bf b}_{10}$ & $\mathbb{R}^{d} $ & $ 768 $ & $\partial \mathcal{L}_{\text{seq}} / \partial {\bf b}_{10} = \sum_\text{tokens} \boldsymbol{\Delta}_{11} $ \\[0.3em]
\hline \hline
\rule{0pt}{1.1\normalbaselineskip}
\multirow{2}*{FC$_\text{logits}$} & & ${\bf w}_{11}$ & $\mathbb{R}^{d \times n_\text{vocab}}$ & $ 38,597,376  $ & $\partial \mathcal{L}_{\text{seq}} / \partial {\bf w}_{11} = {\bf a}_{11}^t \,\boldsymbol{\Delta}_{12} $  \\[0.3em]
& \multirow{-2}*{$\mathbfcal{P}_{11}^\text{(fc)}$} & ${\bf b}_{11}$ & $\mathbb{R}^{n_\text{vocab}} $ & $ 50,257 $ & $\partial \mathcal{L}_{\text{seq}} / \partial {\bf b}_{11} = \sum_\text{tokens} \boldsymbol{\Delta}_{12} $ \\[0.3em]
\hline
\rule{0pt}{1.1\normalbaselineskip}
\multirow{2}*{LoRA$_\text{logits}$} & & ${\bf d}_{11}$ & $\mathbb{R}^{d\times r}$ & $12,288$ & $\partial \mathcal{L}_{\text{seq}} / \partial {\bf d}_{11} = \alpha \, {\bf a}_{11}^t \, \boldsymbol{\Delta}_{12} \, {\bf u}^t_{11} $  \\[0.3em]
& \multirow{-2}*{$\mathbfcal{P}_{11}^\text{(lora)}$} & ${\bf u}_{11}$ & $\mathbb{R}^{r\times n_\text{vocab}} $ & $804,112$ & $\partial \mathcal{L}_{\text{seq}} / \partial {\bf u}_{11} = \alpha \big( {\bf a}_{11} \, {\bf d}_{11} \big)^t \boldsymbol{\Delta}_{12} $ \\[0.3em]
\hline
\end{tabular}
\caption[capParam]{%
{\bf Parameter gradients should be read from bottom to top following the order in which they are updated during backpropagation.} As per common practice, the dimensionality~$d_\rho$ of the queries/keys feature maps in~$\mathbfcal{P}_{Q_h/K_h}$ is set to match~$d_\rho \equiv d_h$ the dimensionality of the values output feature maps in~$\mathbfcal{P}_{V_h}$ and~$n_h = d / d_h \in \mathbb{N}$ denotes the number of head in the multi-headed attention layer.}
\label{table:minimalTransformer_grads}
\end{table}

\newpage

\fcolorbox{black}{pink}{%
\minipage[t]{\dimexpr0.9\linewidth-2\fboxsep-2\fboxrule\relax}

{\it ``The Queen propped her up against a tree, and said kindly,  You may rest a little now. \\

Alice looked round her in great surprise. Why, I do believe we've been under this tree the whole time! Everything's just as it was! \\

Of course it is, said the Queen, what would you have it? \\

Well, in our country, said Alice, still panting a little, you'd generally get to somewhere else~\textemdash~if you ran very fast for a long time, as we've been doing. \\

A slow sort of country! said the Queen. Now, here, you see, it takes all the running you can do, to keep in the same place. \\

If you want to get somewhere else, you must run at least twice as fast as that!''} \\

(Lewis Carroll, Through the Looking-Glass, 1871)

\noindent \\

\hrule

\noindent \\

{\it ``If you're thinking without writing, you only think you're thinking.''} \\

(Leslie Lamport)

\endminipage}

\noindent \\ \\

\newpage

\begin{appendices}

\section{Matrices: some more potpourri...}

\paragraph{Linear mixing.}  For the sake of generality, let us consider a mirror version of~eq.\eqref{eq:finalDesiredAout} where the components~$\rho_{\alpha \beta}$ of the weight matrix~$\boldsymbol{\rho} \sim \mathbb{R}^{n\times n}$ are not limited by causality and where we simplify the notation by denoting with~${\bf a} \sim \mathbb{R}^{n\times d}$ the stack of~$d$-dimensional vectors representing~$n$ tokens. \\

\noindent Let us denote by~$\tilde{\bf a}$ the weighted average of~${\bf a}$ such that
\begin{equation*}
\tilde{\bf a} = \boldsymbol{\rho} \, {\bf a} \quad \Longrightarrow \quad \left(
\begin{matrix}
\tilde{a}_{11} & \horzbar & \tilde{a}_{1d}  \\
 & \vdots &  \\
\tilde{a}_{n1} & \horzbar & \tilde{a}_{nd}
\end{matrix} \right) = \left(
\begin{matrix}
\rho_{11} & \cdots & \rho_{1n}  \\
\vdots & \ddots & \vdots \\
\rho_{n1} & \cdots & \rho_{nn}
\end{matrix} \right) \left(
\begin{matrix}
a_{11} & \horzbar & a_{1d}  \\
 & \vdots &  \\
a_{n1} & \horzbar & a_{nd}
\end{matrix} \right)
\end{equation*}
and focus on a specific token~$\tilde{\bf a}(t=t^\star) = \begin{bmatrix}
   \tilde{a}_{t^\star 1} & \cdots & \tilde{a}_{t^\star d}
\end{bmatrix} \sim \mathbb{R}^d$ of~$\tilde{\bf a} \sim \mathbb{R}^{n\times d}$.  The components of~$\tilde{\bf a}(t=t^\star)$ are given by
\begin{align}
\begin{bmatrix}
\tilde{a}_{t^\star 1} & \cdots & \tilde{a}_{t^\star d}
\end{bmatrix} &= 
\begin{bmatrix}
\rho_{t^\star 1} & \cdots & \rho_{t^\star n}
\end{bmatrix}  \left(
\begin{matrix}
a_{11} & \cdots & a_{1d}  \\
\vdots & \ddots &  \vdots \\
a_{n1} & \cdots & a_{nd}
\end{matrix} \right) \nonumber \\
&= \begin{bmatrix}
\big( \rho_{t^\star 1} a_{11} + \cdots + \rho_{t^\star n} a_{n1} \big) & \cdots & \big( \rho_{t^\star 1} a_{1d} + \cdots + \rho_{t^\star n} a_{nd} \big)
\end{bmatrix} \nonumber \\
&= \rho_{t^\star 1} \begin{bmatrix}
a_{11} & \cdots & a_{1d}
\end{bmatrix} + \cdots + \rho_{t^\star n} \begin{bmatrix}
a_{n1} & \cdots & a_{nd}
\end{bmatrix} \nonumber \\
&\centerwithin{\downarrow}{\,\,=} \colorbox{light-blue}{in vectorized notation} \nonumber \\
\tilde{\bf a}(t=t^\star) &= \rho_{t^\star 1} \, {\bf a}(t=1) + \cdots + \rho_{t^\star n} \, {\bf a}(t=n) \sim \mathbb{R}^d
\label{eq:weighedSumIndividual}
\end{align}

\noindent  This shows that~$\tilde{\bf a}(t=t^\star)$ is a linear combination of all the token feature vectors (i.e. all rows) of~${\bf a}$.  Applying this to all tokens, we get the expected linear mixing
\begin{equation*} \tilde{\bf a} = 
\left( \begin{matrix}
\horzbar \,\, \tilde{\bf a}(t=1) \sim \mathbb{R}^d \,\, \horzbar \\
\vdots \\
\horzbar \,\, \tilde{\bf a}(t=n) \sim \mathbb{R}^d \,\, \horzbar
\end{matrix} \, \right) =
\left( \begin{matrix}
\rho_{11} \, {\bf a}(t=1) + \cdots + \rho_{1n} \, {\bf a}(t=n) \\
\vdots \\
\rho_{n1} \, {\bf a}(t=1) + \cdots + \rho_{nn} \, {\bf a}(t=n)
\end{matrix} \, \right)
\end{equation*}

\noindent In the special case where the weight components~$\rho_{\alpha \beta}$ are causal with~$\rho_{\alpha \beta} = 0$ if~$\beta > \alpha$, we recover the expected system of equations presented in~Section~\ref{sec:singleHead} in the main part of the text
\begin{equation}
\text{Eqs.}\eqref{eq:weighted1}–\eqref{eq:weighted4} \equiv
\begin{cases} \,
\tilde{\bf a}(t=1) = \rho_{11} \, {\bf a}(t=1) \\
\tilde{\bf a}(t=2) = \rho_{21} \, {\bf a}(t=1) + \rho_{22} \, {\bf a}(t=2) \\
\tilde{\bf a}(t=3) = \rho_{31} \, {\bf a}(t=1) + \rho_{32} \, {\bf a}(t=2) + \rho_{33} \, {\bf a}(t=3) \\
\phantom{\tilde{\bf a}(t=1)\,\,\,} \vdots \\
\tilde{\bf a}(t=n) = \rho_{n1} \, {\bf a}(t=1) + \rho_{n2} \, {\bf a}(t=2) + \rho_{n3} \, {\bf a}(t=3) + \cdots + \rho_{nn} \, {\bf a}(t=n)
\end{cases}
\label{eq:rowLinearMix}
\end{equation}
simply by identifying~$\tilde{\bf a} \equiv {\bf a}_i^h$,~${\bf a} \equiv {\bf v}_h$ and~$n\equiv n_\mathcal{T}$.

\paragraph{Cycles of Frobenius products.} \hypertarget{fourMatFrobenius}{Generalizing} the Frobenius product identity shown in~Eq.~(52) of reference~\cite{deepPedestrians}, we consider 4 matrices~${\bf A}, {\bf B}, {\bf C}$ and~${\bf D}$ subject to dimensionality constraints such that~${\bf A} \sim {\bf B} {\bf C} {\bf D} \sim \mathbb{R}^{n\times f}$.  By definition of the Frobenius product and simple rewriting of the expressions, we have
\begin{align*}
{\bf A} \cdot ({\bf B} {\bf C} {\bf D}) &= \text{Tr} \big( {\bf A}^t \, {\bf B} {\bf C} {\bf D}  \big) \\
&= \text{Tr} \big[ \big( ( {\bf A}^t \, {\bf B} {\bf C})^t  \big)^t \, {\bf D}  \big] \\
&= \text{Tr} \big[ \big( {\bf C}^t {\bf B}^t {\bf A} \big)^t \, {\bf D}  \big] \\
&= {\bf C}^t {\bf B}^t {\bf A} \cdot {\bf D} \\
&= \big[ \big( {\bf B} {\bf C} \big)^t {\bf A} \big] \cdot {\bf D}
\end{align*}

\noindent Other useful identities may be obtained by using the invariance of the trace under circular shifts and the same manipulations as above
\begin{equation*}
{\bf A} \cdot ({\bf B} {\bf C} {\bf D}) = \text{Tr} \big( {\bf A}^t \, {\bf B} {\bf C} {\bf D}  \big) = \begin{cases} 
\hspace{0.1cm}  \text{Tr} \big( {\bf D} {\bf A}^t \, {\bf B} {\bf C} \big) = \big( {\bf B}^t {\bf A} {\bf D}^t \big) \cdot {\bf C}  \\
\hspace{0.1cm}  \text{Tr} \big( {\bf C} {\bf D} {\bf A}^t \, {\bf B} \big) = \big[ {\bf A} \big( {\bf C} {\bf D} \big)^t \, \big] \cdot {\bf B}
\end{cases}
\end{equation*}

\noindent Another useful identity which can be derived from the same cyclic property and transpose invariance of the trace operator
\begin{equation}
{\bf A} \cdot {\bf B}^t = {\bf A}^t \cdot {\bf B}
\label{eq:FrobeniusTranspose}
\end{equation}

\paragraph{More broadcasting semantics.}  Let us start by considering two matrices~${\bf A} \sim {\bf B} \sim \mathbb{R}^{n\times n}$ to mirror part of the expression for~$\boldsymbol{\Delta}^h_{\text{causal}}$ in eq.~\eqref{eq:DeltaCausal} (where~${\bf A} =  \boldsymbol{\Delta}^h_i {\bf v}_h^t$ and~${\bf B} = \boldsymbol{\rho}_{({\bf a}_{i-1}, \, h)}$).  Their feature dot-product~${\bf A}\ominus {\bf B} \sim \mathbb{R}^n$ results in a column vector which is broadcast column-wise as
\begin{equation}
\widetilde{ {\bf A} \ominus {\bf B} } = \left(
\begin{matrix}
{\bf a}_1 \cdot {\bf b}_1 & \cdots & {\bf a}_1 \cdot {\bf b}_1  \\
\vdots & \ddots & \vdots \\
{\bf a}_n \cdot {\bf b}_n & \cdots & {\bf a}_n \cdot {\bf b}_n 
\end{matrix} \right)
\label{eq:BroadcastFtDot}
\end{equation}

\noindent See~eq.(48) of the reference paper~\cite{deepPedestrians} for a reminder of the~$\ominus$ operator.  The same broadcast can be expressed in a more convenient vectorized manner for practical implementations with
\begin{align*}
\widetilde{ {\bf A} \ominus {\bf B} }
&= \left(
\begin{matrix}
(a_{11} b_{11} + a_{1n} b_{1n}) & \cdots & (a_{11} b_{11} + a_{1n} b_{1n})  \\
\vdots & \ddots & \vdots \\
(a_{n1} b_{n1} + a_{nn} b_{nn}) & \cdots & (a_{n1} b_{n1} + a_{nn} b_{nn})
\end{matrix} \right) \\
&= \left[ \left(
\begin{matrix}
a_{11} & \cdots & a_{1n}  \\
\vdots & \ddots & \vdots \\
a_{n1} & \cdots & a_{nn}
\end{matrix} \right) \circ \left(
\begin{matrix}
b_{11} & \cdots & b_{1n}  \\
\vdots & \ddots & \vdots \\
b_{n1} & \cdots & b_{nn}
\end{matrix} \right) \right] \left(
\begin{matrix}
1 & \cdots & 1  \\
\vdots & \ddots & \vdots \\
1 & \cdots & 1
\end{matrix} \right)  \\
&= ({\bf A} \circ {\bf B}) \, {\bf J}_{n,n}
\end{align*}

\noindent Next, let us consider three square matrices~${\bf A} \sim {\bf B} \sim {\bf C} \sim \mathbb{R}^{n\times n}$ and simplify~$\big( {\bf A} \circ {\bf B}  \big) \cdot \big( \widetilde{{\bf B} \ominus {\bf C}}  \big)$ consisting of the Frobenius product between~$({\bf A} \circ {\bf B}) \sim \mathbb{R}^{n\times n}$ and the column-wise broadcast of the feature vector~$\mathbb{R}^n$ given by the feature dot-product~${\bf B} \ominus {\bf C} \sim \mathbb{R}^n$.  Since the result of the feature dot-product is a column vector (one dot-product per row), the broadcast has to be column-wise.  Note that in~\cite{deepPedestrians}, the sum over tokens is denoted as a sum over ``samples''.  In this paper, we use the terminology ``token'' as the unit instead of sample since we prefer to think of a sample as a sequence of tokens.  Regardless, tokens have their own feature maps just like samples do in~\cite{deepPedestrians}.  Therefore, the symbols~$\sum_\text{tokens}$ and~$\sum_\text{samples}$ are functionally equivalent to each other and one should just consider them as a vertical sum along columns (i.e. that cuts across rows).  With this in mind, we can now evaluate the desired expression
{\allowdisplaybreaks
\begin{align*}
\big( {\bf A} \circ {\bf B}  \big) \cdot \big( \widetilde{{\bf B} \ominus {\bf C}}  \big) &= \sum_{\text{tokens}} \big( {\bf A} \circ {\bf B}  \big) \ominus \big( \widetilde{{\bf B} \ominus {\bf C}}  \big) \\
&= \sum_{\text{tokens}} \left[ \left(
\begin{matrix}
 a_{11} & \cdots & a_{1n}  \\
\vdots & \ddots & \vdots \\
a_{n1} & \cdots &  a_{nn}
\end{matrix} \right) \left(
\begin{matrix}
b_{11} & \cdots & b_{1n}  \\
\vdots & \ddots & \vdots \\
b_{n1} & \cdots &  b_{nn}
\end{matrix} \right) \ominus \widetilde{ \left( \left(
\begin{matrix}
b_{11} & \cdots & b_{1n}  \\
\vdots & \ddots & \vdots \\
b_{n1} & \cdots &  b_{nn}
\end{matrix} \right) \ominus \left(
\begin{matrix}
c_{11} & \cdots & c_{1n}  \\
\vdots & \ddots & \vdots \\
c_{n1} & \cdots &  c_{nn}
\end{matrix} \right) \right) } \right] \\
&= \sum_{\text{tokens}} \left[ \left(
\begin{matrix}
a_{11} b_{11}& \cdots & a_{1n} b_{1n}  \\
\vdots & \vdots & \vdots \\
a_{n1} b_{n1} & \cdots & a_{nn} b_{nn} 
\end{matrix} \right) \ominus \widetilde{ \left(
\begin{matrix}
{\bf b}_1 \cdot {\bf c}_1   \\
\vdots \\
{\bf b}_n \cdot {\bf c}_n 
\end{matrix} \right) } \right] \\
&\centerwithin{\downarrow}{\,\,=} \colorbox{light-blue}{using the same column-wise broadcast as in eq.\eqref{eq:BroadcastFtDot}} \\
&= \sum_{\text{tokens}} \left[ \left(
\begin{matrix}
a_{11} b_{11}& \cdots & a_{1n} b_{1n}  \\
\vdots & \ddots & \vdots \\
a_{n1} b_{n1} & \cdots & a_{nn} b_{nn} 
\end{matrix} \right) \ominus \left(
\begin{matrix}
{\bf b}_1 \cdot {\bf c}_1  & \horzbar & {\bf b}_1 \cdot {\bf c}_1  \\[0.4em]
\cdots & \cdots & \cdots \\
{\bf b}_n \cdot {\bf c}_n & \horzbar & {\bf b}_n \cdot {\bf c}_n
\end{matrix} \right) \right] \\
&= \sum_{\text{tokens}} \left(
\begin{matrix}
\big( a_{11} b_{11} + \cdots + a_{1n} b_{1n}  \big) \, {\bf b}_1 \cdot {\bf c}_1  \\
\vdots \\
\big( a_{n1} b_{n1} + \cdots + a_{nn} b_{nn}  \big) \, {\bf b}_n \cdot {\bf c}_n 
\end{matrix} \right) \\
&= \sum_{\text{tokens}} \left(
\begin{matrix}
\big( {\bf a}_1 \cdot {\bf b}_1  \big) \, {\bf b}_1 \cdot {\bf c}_1  \\
\vdots \\
\big(  {\bf a}_n \cdot {\bf b}_n \big) \, {\bf b}_n \cdot {\bf c}_n 
\end{matrix} \right) \\
&= \sum_{\text{tokens}} \left(
\begin{matrix}
\big( {\bf a}_1 \cdot {\bf b}_1  \big)\,  b_{11} & \cdots &  \big( {\bf a}_1 \cdot {\bf b}_1  \big) \, b_{1n}  \\
\vdots & \ddots & \vdots \\
\big(  {\bf a}_n \cdot {\bf b}_n \big) \, b_{n1} & \cdots & \big( {\bf a}_1 \cdot {\bf b}_1  \big)\,  b_{nn}
\end{matrix} \right) \ominus \left(
\begin{matrix}
{\bf c}_1  \\
\vdots \\
{\bf c}_n 
\end{matrix} \right) \\
&= \sum_{\text{tokens}} \left[ \left(
\begin{matrix}
{\bf a}_1 \cdot {\bf b}_1  & \cdots & {\bf a}_1 \cdot {\bf b}_1   \\
\vdots & \ddots & \vdots \\
{\bf a}_n \cdot {\bf b}_n  & \cdots & {\bf a}_1 \cdot {\bf b}_1 
\end{matrix} \right) \circ \left(
\begin{matrix}
 b_{11} & \cdots &  b_{1n}  \\
\vdots & \ddots & \vdots \\
b_{n1} & \cdots & b_{nn}
\end{matrix} \right) \right] \ominus \left(
\begin{matrix}
c_{11} & \cdots &  c_{1n}  \\
\vdots & \ddots & \vdots \\
c_{n1} & \cdots &  c_{nn}
\end{matrix} \right)   \\
&= \sum_{\text{tokens}} \big[ \big( \widetilde{{\bf A}\ominus {\bf B}} \big) \circ {\bf B}  \big] \ominus {\bf C} \\
&= \big[ \big( \widetilde{{\bf A}\ominus {\bf B}} \big) \circ {\bf B}  \big] \cdot {\bf C}
\end{align*}
}

\noindent Summarizing the result, we have
\begin{equation}
\big( {\bf A} \circ {\bf B}  \big) \cdot \big( \widetilde{{\bf B} \ominus {\bf C}}  \big) = \big[ \big( \widetilde{{\bf A}\ominus {\bf B}} \big) \circ {\bf B}  \big] \cdot {\bf C}
\label{eq:potPourri}
\end{equation}

\paragraph{Sum over rows over matrix product.}  Let us consider two matrices~${\bf A} \sim \mathbb{R}^{n\times n}$ and~${\bf B}\sim \mathbb{R}^{n \times f}$ to mirror the situation of~eqs.\eqref{eq:biasesValuesSimplified} and~\eqref{eq:biasZeroGrad}. We are interested in simplifying the row-wise sum (i.e. vertical) of their matrix product~$\sum_\text{rows} {\bf A} \, {\bf B}$ resulting in a vector~$\sim \mathbb{R}^f$ (one component per column). Indexing rows by~$i$ and columns by~$j$, the row-wise sum associated with column~$j$ of the product~${\bf A}\,{\bf B}$ is given by
\begin{equation*}
\sum_{i=1}^n \left( {\bf A}\,{\bf B}  \right)_{ij}
= \sum_{i=1}^n \left( \sum_{k=1}^n a_{ik} b_{kj}  \right) 
=  \sum_{k=1}^n \sum_{i=1}^n  a_{ik} b_{kj}
=  \sum_{k=1}^n \left( \sum_{i=1}^n  a_{ik} \right) b_{kj}
=  \sum_{k=1}^n {a}_k^{\updownarrow} \, b_{kj}
\end{equation*}
where we defined~${a}_k^{\updownarrow} = \sum_{i=1}^n a_{ik}$ as the row-wise (vertical) sum of the $k^\text{th}$ column of~${\bf A}$. Collecting together the~$n$ different column sums~${a}_k^{\updownarrow}$, we define a vector
\begin{equation*}
{\bf a}^{\updownarrow} = \big[ {a}_1^{\updownarrow} = \sum_{i=1}^n a_{i1} \, , \cdots , {a}_n^{\updownarrow} = \sum_{i=1}^n a_{in} \big] \sim \mathbb{R}^n
\end{equation*}
with which the all the components of the desired expression can now be evaluated as the vector-matrix product
\begin{equation}
\sum_\text{rows} {\bf A} \, {\bf B} = {\bf a}^{\updownarrow} \, {\bf B} \sim \mathbb{R}^f 
\label{eq:sumRows}
\end{equation}

\section{Some properties of the softmax function}

\paragraph{Shift invariance.}  Let us consider a square matrix~${\bf G} \sim \mathbb{R}^{n\times n}$ (analogous to an attention weight matrix), a feature map matrix~${\bf H} \sim \mathbb{R}^{n \times f}$ and a feature bias vector~${\bf b} \sim \mathbb{R}^f$.  In this case, we can reproduce the pattern of~eq.\eqref{eq:softMaxInvariance} that we wish to expand as
\begin{equation}
\text{softmax} ( \underbrace{ \mathcolorbox{Gray}{ \left( {\bf a}_{i-1} {\bf w}_{q_h} + \widetilde{\bf b}_{q_h} \right) \left( {\bf a}_{i-1} {\bf w}_{k_h} \right)^t } }_{\textstyle
\begin{gathered}
{\bf G}
\end{gathered} } + \underbrace{ \mathcolorbox{Gray}{ \left( {\bf a}_{i-1} {\bf w}_{q_h} + {\bf b}_{q_h} \right)  } }_{\textstyle
\begin{gathered}
{\bf H}
\end{gathered} } \, \underbrace{ \mathcolorbox{Gray}{ \widetilde{{\bf b}_{k_h}^t}} }_{\textstyle
\begin{gathered}
\widetilde{\bf b^t}
\end{gathered} }
 ) \quad \Longrightarrow \quad \text{softmax} \left( {\bf G} + {\bf H} \, \widetilde{\bf b^t} \right) 
\label{eq:reproducesoftMaxInvariance}
\end{equation}

\noindent The first step consists in broadcasting the vector~${\bf b} \sim \mathbb{R}^f$ row-wise into a matrix~$\widetilde{\bf b} \sim \mathbb{R}^{n\times f}$ with dimensions compatible with~${\bf H} \sim \mathbb{R}^{n\times f}$.
\begin{equation*}
{\bf b} = \left[ b_1 , \cdots , b_f \right] \sim \mathbb{R}^f \quad  \Longrightarrow \,\,
\begin{cases} 
\widetilde{\bf b} &\leftarrow \left(
\begin{matrix}
\horzbar \,{\bf b} \sim \mathbb{R}^f \, \horzbar \\
\vdots \\
\horzbar \, {\bf b} \sim \mathbb{R}^f \, \horzbar
\end{matrix} \right) = \left(
\begin{matrix}
b_1 & \horzbar & b_f \\
\cdots & \cdots & \cdots \\
\cdots & \cdots & \cdots \\
b_1 & \horzbar & b_f
\end{matrix} \right)  \sim \mathbb{R}^{n \times f}  \\[0.6em]
\widetilde{\bf b^t} &\leftarrow \left( \begin{array}{ccccc}
\vertbar &   & \vertbar \\
{\bf b}  &  \ldots & {\bf b}   \\
\vertbar &     & \vertbar 
\end{array} \right) = \left(
\begin{matrix}
b_1 & \vdots & \vdots & b_1 \\
\vertbar & \vdots & \vdots & \vertbar \\
b_f & \vdots & \vdots & b_f
\end{matrix} \right) \sim \mathbb{R}^{f \times n}
\end{cases}
\end{equation*}

\noindent This way, the matrix product~${\bf H} \, \widetilde{\bf b^t} \sim \mathbb{R}^{n \times n}$ produces a square matrix which can be added to~${\bf G}$.  This is written explicitly as
\begin{align*}
\text{softmax} \left(  {\bf G} + {\bf H} \, \widetilde{\bf b^t} \right) &= \text{softmax} \left[ \left(
\begin{matrix}
g_{11} & \cdots & g_{1n} \\[0.4em]
\vdots & \ddots & \vdots \\[0.4em]
g_{n1} & \cdots & g_{nn}
\end{matrix} \right) + \left(
\begin{matrix}
h_{11} & \cdots & h_{1f} \\[0.4em]
\vdots & \ddots & \vdots \\[0.4em]
h_{n1} & \cdots & h_{nf}
\end{matrix} \right) \left(
\begin{matrix}
b_1 & \vdots & \vdots & b_1 \\
\vertbar & \vdots & \vdots & \vertbar \\
b_f & \vdots & \vdots & b_f
\end{matrix} \right) \, \right] \\
&= \text{softmax} \left[ \left(
\begin{matrix}
g_{11} & \cdots & g_{1n} \\[0.4em]
\vdots & \ddots & \vdots \\[0.4em]
g_{n1} & \cdots & g_{nn}
\end{matrix} \right) + \left(
\begin{matrix}
{\bf h}_1 \cdot {\bf b} & \horzbar & {\bf h}_1 \cdot {\bf b} \\
\cdots & \cdots & \cdots \\
\cdots & \cdots & \cdots \\
{\bf h}_n \cdot {\bf b} & \horzbar & {\bf h}_n \cdot {\bf b}
\end{matrix} \right) \right] \\
&\centerwithin{\downarrow}{\,\,=} \colorbox{light-blue}{the product ${\bf H} \, \widetilde{\bf b^t}  $ creates a shift-matrix where all rows are constant} \\
&\centerwithin{}{\,\,=} \colorbox{light-blue}{define $\lambda_i \equiv {\bf v}_i \cdot {\bf b} \sim  \mathbb{R}$ as a short-hand notation for the row-specific constant} \\
&= \text{softmax} \left(
\begin{matrix}
g_{11} + \lambda_1 & \cdots & g_{1n} + \lambda_1 \\
\cdots & \ddots & \cdots \\
g_{n1} + \lambda_n & \cdots & g_{nn} + \lambda_n
\end{matrix} \right) \\
&\centerwithin{\downarrow}{\,\,=} \colorbox{light-blue}{showing that the components of each row~$i$ are all shifted by a constant value~$\lambda_i$} \\
&= \left(
\begin{matrix}
\text{softmax} \left( g_{11} + \lambda_1 , \, \cdots \, , g_{1n} + \lambda_1 \right) \\
\vdots  \\
\text{softmax} \left( g_{n1} + \lambda_n , \, \cdots \, , g_{nn} + \lambda_n \right)
\end{matrix} \right)
\end{align*}

\noindent Using one specific row~$i$ as an example, the \colorbox{light-blue}{shift-invariance property of the softmax normalization} can be understood as
\begin{align*}
\text{softmax} \left( g_{i1} + \lambda_i , \, \cdots \, , g_{in} + \lambda_i \right) &= \left[ \dfrac{e^{(g_{i1} + \lambda_i)}}{  e^{ (g_{i1} + \lambda_i)} + \cdots + e^{(g_{in} + \lambda_i)} } , \cdots , \dfrac{ e^{(g_{in} + \lambda_i)}} {   e^{ (g_{i1} + \lambda_i)} + \cdots + e^{(g_{in} + \lambda_i)}  } \right] \\
&= \left[ \dfrac{e^{g_{i1}} \, e^{\lambda_i} }{ \left( e^{ g_{i1}} + \cdots + e^{g_{in}} \right) \, e^{\lambda_i} } , \cdots , \dfrac{ e^{g_{in}} \, e^{\lambda_i} }{  \left( e^{ g_{i1}} + \cdots + e^{g_{in}} \right) \, e^{\lambda_i}  } \right] \\
&= \left[ \dfrac{e^{g_{i1}} }{  e^{ g_{i1}} + \cdots + e^{g_{in}}  } , \cdots , \dfrac{ e^{g_{in}} } {  e^{ g_{i1}} + \cdots + e^{g_{in}}  } \right] \\
&= \text{softmax} \left( g_{i1} , \, \cdots \, , g_{in} \right)
\end{align*}
where the dependence on~$\lambda_i$ cancels out.  Applying this shift-invariance to all rows leads us to the result
\begin{equation}
\text{softmax} \left(  {\bf G} + {\bf H} \, \widetilde{\bf b}^t \right) = \text{softmax} \, {\bf G}
\label{eq:softMaxId}
\end{equation}
showing that the dependence on~$\widetilde{\bf b^t}$ completely drops out.  Notice that this was possible due to the transpose of the {\bf broadcast} of~${\bf b}$ which created a shift-matrix~${\bf H} \, \widetilde{\bf b^t}$. \\

\noindent Going back to the original notation, one should note that~${\bf G}$ contains the parameters~$\{ {\bf w}_{q_h}, {\bf b}_{q_h} , {\bf w}_{k_h} \}$.  So it is really just the bias parameter~${\bf b}_{k_h}$ that is redundant in evaluating~eq.\eqref{eq:reproducesoftMaxInvariance}, copy of~\eqref{eq:softMaxInvariance}.

\paragraph{Permutations.}  We investigate the behavior of the softmax function for
\begin{itemize}
\item \uwave{row-wise} permutations.  Let us consider a matrix~${\bf A} \sim \mathbb{R}^{n\times d}$ with pre-multiplication by the permutation matrix~$\mathcal{P}_1 \sim \mathbb{R}^{n\times n}$ so that~$\mathcal{P}_1 \, {\bf A}$ is a row-permutated version of~${\bf A}$~\cite{wikiPerm}.  What is the effect of~$\mathcal{P}_1$ on~$\text{softmax} \big( \mathcal{P}_1 \, {\bf A} \big)$?  Since the softmax function is applied independently to each row and does not mix data across the rows, there is no difference between~i) first permutating the rows and next applying softmax to the permutated rows, i.e.~$\text{softmax} \big( \mathcal{P}_1 \, {\bf A}  \big)$ or~ii)~ first applying the softmax to each row and next permutating the rows, i.e.~$ \mathcal{P}_1 \,\text{softmax} \, {\bf A}$.  Therefore we have
\begin{equation}
\refstepcounter{equation}
\setcounter{subEquation}{0}
\refstepcounter{subEquation}
\tag{\thesubEquation} \label{softmaxPermRows} \\
\text{softmax} \big( \mathcal{P}_1 \, {\bf A}  \big) = \mathcal{P}_1 \,\text{softmax} \, {\bf A}
\end{equation}
\item \uwave{column-wise} permutations.  Let us consider another matrix of transposed dimensions~${\bf B} \sim \mathbb{R}^{d\times n}$ with post-multiplication by another permutation matrix~$\mathcal{P}_2 \sim \mathbb{R}^{n\times n}$ so that~${\bf B}\,\mathcal{P}_2$ is a column-permutated version of~${\bf B}$~\cite{wikiPerm}. What is the effect of~$\mathcal{P}_2$ on~$\text{softmax} \big( {\bf B} \, \mathcal{P}_2 \big)$?  Just like before, the softmax function does not mix computations across different rows.  This means that, for a given row, the normalization of the denominator is constant across all the columns.  Since here we are considering only a re-ordering of the columns, it does not matter whether the exponential of the numerator is applied before or after the re-ordering is performed.  Therefore, we have
\begin{equation}
\text{softmax} \big( {\bf B} \, \mathcal{P}_2 \big) = \text{softmax} ({\bf B}) \, \mathcal{P}_2
\refstepcounter{subEquation}
\tag{\thesubEquation}
\label{softmaxPermCols}
\end{equation}
\end{itemize} 

\noindent In the main part of the text, we are considering a special case where~${\bf B} = {\bf A}^t$ and the same permutation matrix with~$\mathcal{P}_2 = \mathcal{P}_1^t$.

\end{appendices}

\newpage

\hypertarget{contents}{}
\tableofcontents

\end{document}